\documentclass[twoside,11pt]{article}

\usepackage{amsmath}
\usepackage{bbm}
\usepackage{booktabs, array}
\usepackage{enumitem}

\usepackage[usenames,dvipsnames]{xcolor}
\definecolor{shadecolor}{gray}{0.9}

\definecolor{mydarkblue}{rgb}{0,0.08,0.45}
\usepackage[colorlinks,citecolor=mydarkblue,urlcolor=mydarkblue,linkcolor=mydarkblue,filecolor=mydarkblue]{hyperref}
\usepackage{datetime2}

\usepackage[parfill]{parskip}[=v1]
\usepackage[nameinlink]{cleveref}
\creflabelformat{equation}{#2\textup{#1}#3}  

\usepackage{subcaption} 
\usepackage{wrapfig}
\usepackage{tabularx}

\usepackage[nohyperref]{jmlr2e}

\usepackage{subfiles}

\usepackage{amsmath,amsfonts,bm}

\def\eqref#1{equation~\ref{#1}}

\def\1{\bm{1}}

\def\rva{{\mathbf{a}}}
\def\rvb{{\mathbf{b}}}

\def\rvu{{\mathbf{i}}}

\def\rvs{{\mathbf{s}}}

\def\rvu{{\mathbf{u}}}
\def\rvv{{\mathbf{v}}}
\def\rvw{{\mathbf{w}}}
\def\rvx{{\mathbf{x}}}
\def\rvy{{\mathbf{y}}}
\def\rvz{{\mathbf{z}}}

\def\ervu{{\textnormal{u}}}

\def\ervx{{\textnormal{x}}}

\def\ervz{{\textnormal{z}}}

\def\rmA{{\mathbf{A}}}

\def\rmD{{\mathbf{D}}}

\def\rmH{{\mathbf{H}}}
\def\rmI{{\mathbf{I}}}

\def\rmL{{\mathbf{L}}}

\def\rmP{{\mathbf{P}}}
\def\rmQ{{\mathbf{Q}}}
\def\rmR{{\mathbf{R}}}
\def\rmS{{\mathbf{S}}}

\def\rmU{{\mathbf{U}}}
\def\rmV{{\mathbf{V}}}
\def\rmW{{\mathbf{W}}}
\def\rmX{{\mathbf{X}}}

\def\vmu{{\bm{\mu}}}
\def\vsigma{{\bm{\sigma}}}
\def\vepsilon{{\bm{\epsilon}}}
\def\valpha{{\bm{\alpha}}}
\def\vtheta{{\bm{\theta}}}
\def\vbeta{{\bm{\beta}}}

\def\veta{{\bm{\eta}}}

\def\vphi{{\bm{\phi}}}
\def\vpsi{{\bm{\psi}}}

\def\vh{{\bm{h}}}

\def\vs{{\bm{s}}}

\def\vu{{\bm{u}}}

\DeclareMathAlphabet{\mathsfit}{\encodingdefault}{\sfdefault}{m}{sl}
\SetMathAlphabet{\mathsfit}{bold}{\encodingdefault}{\sfdefault}{bx}{n}

\def\sR{{\mathbb{R}}}
\def\sS{{\mathbb{S}}}

\def\sZ{{\mathbb{Z}}}

\newcommand{\R}{\mathbb{R}}

\DeclareMathOperator*{\argmax}{arg\,max}

\newcommand{\br}[1]{\mathopen{}\left(#1\right)\mathclose{}}
\newcommand{\abs}[1]{\left|#1\right|}
\newcommand{\norm}[1]{\left\|#1\right\|}
\newcommand{\set}[1]{\left\{#1\right\}}
\newcommand{\kl}[2]{D_{\mathrm{KL}}\left[\,#1\,\|\,#2\,\right]}
\newcommand{\avg}[2]{{\mathop{\mathbb{E}}}_{#2}\left[\,#1\,\right]}
\newcommand{\integral}[2]{\int #1 \,d#2}
\newcommand{\integralab}[4]{\int_{#3}^{#4} #1 \,d#2}
\newcommand{\integralx}[3]{\int_{#3} #1 \,d#2}
\newcommand{\bigo}[1]{\mathcal{O}\br{#1}}
\newcommand{\trace}[1]{\mathrm{Tr}\!\left\{#1\right\}}

\newcommand{\createDensityWithSub}[2]{{#1}_{\mathrm{#2}}}
\newcommand{\pz}{\createDensityWithSub{p}{z}}
\newcommand{\pzprime}{\createDensityWithSub{p}{z'}}
\newcommand{\px}{\createDensityWithSub{p}{x}}
\newcommand{\ptildex}{\createDensityWithSub{\widetilde{p}}{x}}
\newcommand{\pu}{\createDensityWithSub{p}{u}}
\newcommand{\puprime}{\createDensityWithSub{p}{u'}}

\newcommand{\qu}{\createDensityWithSub{q}{u}}

\newcommand{\prob}{\mathrm{Pr}}
\newcommand{\g}{\,|\,}

\newcommand{\setU}{\mathcal{U}}
\newcommand{\setX}{\mathcal{X}}
\newcommand{\setI}{\mathcal{I}}
\newcommand{\setZ}{\mathcal{Z}}

\usepackage{ragged2e}

\newcommand{\ourtitle}{Normalizing Flows for Probabilistic Modeling and Inference}
\usepackage{lastpage}
\jmlrheading{22}{2021}{1-\pageref{LastPage}}{12/19}{3/21}{19-1028}{George Papamakarios, Eric Nalisnick, Danilo Jimenez Rezende, Shakir Mohamed, Balaji Lakshminarayanan}
\ShortHeadings{\ourtitle}{Papamakarios, Nalisnick, Rezende, Mohamed and Lakshminarayanan}
\firstpageno{1}

\begin{document}

\title{\ourtitle}

\author{\name George Papamakarios\thanks{Authors contributed equally.} \email gpapamak@google.com  \\   
\name Eric Nalisnick\footnotemark[1] \email enalisnick@google.com  \\ 
\name Danilo Jimenez Rezende \email danilor@google.com \\ 
\name Shakir Mohamed \email shakir@google.com \\ 
\name Balaji Lakshminarayanan \email balajiln@google.com \\
\addr DeepMind 
} 
\editor{Ryan P.~Adams}

\maketitle

\begin{abstract}
Normalizing flows provide a general mechanism for defining expressive probability distributions, only requiring the specification of a (usually simple) base distribution and a series of bijective transformations. There has been much recent work on normalizing flows, ranging from improving their expressive power to expanding their application. We believe the field has now matured and is in need of a unified perspective. In this review, we attempt to provide such a perspective by describing flows through the lens of probabilistic modeling and inference. We place special emphasis on the fundamental principles of flow design, and discuss foundational topics such as expressive power and computational trade-offs. We also broaden the conceptual framing of flows by relating them to more general probability transformations. Lastly, we summarize the use of flows for tasks such as generative modeling, approximate inference, and supervised learning.  
\end{abstract}

  

\section{Introduction}
\label{sec:intro}

The search for well-specified probabilistic models---models that correctly describe the processes that produce data---is one of the enduring ideals of the statistical sciences. Yet, in only the simplest of settings are we able to achieve this goal. A central need in all of statistics and machine learning is then to develop the tools and theories that allow ever-richer probabilistic descriptions to be made, and consequently, that make it possible to develop better-specified models.

This paper reviews one tool we have to address this need: building probability distributions as normalizing flows. Normalizing flows operate by pushing a simple density through a series of transformations to produce a richer, potentially more multi-modal distribution---like a fluid flowing through a set of tubes.  As we will see, repeated application of even simple transformations to a unimodal initial density leads to models of exquisite complexity.  This flexibility means that flows are ripe for use in the key statistical tasks of modeling, inference, and simulation. 

Normalizing flows are an increasingly active area of machine learning research. Yet there is an absence of a unifying lens with which to understand the latest advancements and their relationships to previous work. The thesis of \citet{papamakarios2019phd_thesis} and the survey by \citet{kobyzev2020flowreview} have made steps in establishing this broader understanding. Our review complements these existing papers. In particular, our treatment of flows is more comprehensive than \citet{papamakarios2019phd_thesis}'s but shares some organizing principles. \citet{kobyzev2020flowreview}'s article is commendable in its coverage and synthesis of the literature, discussing both finite and infinitesimal flows (as we do) and curating the latest results in density estimation. Our review is more tutorial in nature and provides in-depth discussion of several areas that \citet{kobyzev2020flowreview} label as open problems (such as extensions to discrete variables and Riemannian manifolds).

Our exploration of normalizing flows attempts to illuminate enduring principles that will guide their construction and application for the foreseeable future.  Specifically, our review begins by establishing the formal and conceptual structure of normalizing flows in \Cref{sec:flows}. Flow construction is then discussed in detail, both for finite (\Cref{sec:constructing_discrete}) and infinitesimal (\Cref{sec:constructing_continuous}) variants. A more general perspective is then presented in \Cref{sec:generalizations}, which in turn allows for extensions to structured domains and geometries. Lastly, we discuss commonly encountered applications in \Cref{sec:applications}.

\paragraph{Notation} We use bold symbols to indicate vectors (lowercase) and matrices (uppercase), otherwise variables are scalars. We indicate probabilities by $\prob(\cdot)$ and probability densities by $p(\cdot)$.  We will also use $p(\cdot)$ to refer to the distribution with that density function. We often add a subscript to probability densities---e.g.~$\px(\rvx)$---to emphasize which random variable they refer to. The notation $p(\rvx; \vtheta)$ represents the distribution of random variables $\rvx$ with distributional parameters $\vtheta$. The symbol $\nabla_{\vtheta}$ represents the gradient operator that collects all partial derivatives of a function with respect to parameters in the set $\vtheta$, that is $\nabla_{\vtheta}f = [\frac{\partial f}{\partial \theta_1}, \ldots, \frac{\partial f}{\partial \theta_K}]$ for $K$-dimensional parameters. The Jacobian of a function \mbox{$f:\R^D\rightarrow\R^D$} is denoted by $J_f(\cdot)$. Finally, we represent the sampling or simulation of variates $\rvx$ from a distribution $p(\rvx)$ using the notation $\rvx \sim p(\rvx)$.

\section{Normalizing Flows}
\label{sec:flows}

We begin by outlining basic definitions and properties of normalizing flows. We establish the expressive power of flow-based models, explain how to use flows in practice, and provide some historical background. This section doesn't assume prior familiarity with normalizing flows, and can serve as an introduction to the field.

\subsection{Definition and Basics}
\label{sec:flows:definition}

Normalizing flows provide a general way of constructing flexible probability distributions over continuous random variables.
Let $\rvx$ be a $D$-dimensional real vector, and suppose we would like to define a joint distribution over $\rvx$. The main idea of flow-based modeling is to express $\rvx$ as a transformation $T$ of a real vector $\rvu$ sampled from $\pu(\rvu)$:
\begin{equation}
    \rvx = T(\rvu)\quad\text{where}\quad\rvu\sim \pu(\rvu).
    \label{eq:flow_sampling}
\end{equation}
We refer to $\pu(\rvu)$ as the \emph{base distribution} of the flow-based model.\footnote{Some papers refer to $\pu(\rvu)$ as the `prior' and to $\rvu$ as the `latent variable'. We believe that this terminology is not as well-suited for normalizing flows as it is for latent-variable models. Upon observing $\rvx$, the corresponding $\rvu = T^{-1}(\rvx)$ is uniquely determined and thus no longer `latent'.} The transformation $T$ and the base distribution $\pu(\rvu)$ can have parameters of their own (denote them as $\vphi$ and $\vpsi$ respectively); this induces a family of distributions over $\rvx$ parameterized by $\set{\vphi, \vpsi}$.

The defining property of flow-based models is that the transformation $T$ must be \emph{invertible} and both $T$ and $T^{-1}$ must be \emph{differentiable}. Such transformations are known as \textit{diffeomorphisms} and require that $\rvu$ be $D$-dimensional as well \citep{milnor1997topology}. Under these conditions, the density of $\rvx$ is well-defined and can be obtained by a change of variables \citep{rudin2006real, bogachev2007measure}:
\begin{equation}
    \px(\rvx) = \pu(\rvu)\,\abs{\det J_T(\rvu)}^{-1}
    \quad\text{where}\quad
    \rvu = T^{-1}(\rvx).
    \label{eq:flow_evaluating_density_T}
\end{equation}
Equivalently, we can also write $\px(\rvx)$ in terms of the Jacobian of $T^{-1}$:
\begin{equation}
    \px(\rvx) = \pu\br{T^{-1}\br{\rvx}}\,\abs{\det J_{T^{-1}}\br{\rvx}}.
    \label{eq:flow_evaluating_density_Tinv}
\end{equation}  The Jacobian $J_T(\rvu)$ is the $D\times D$ matrix of all partial derivatives of $T$ given by:
\begin{equation}
     J_T(\rvu) =
   \begin{bmatrix}
    \frac{\partial T_1}{\partial \ervu_{1}} & \cdots & \frac{\partial T_1}{\partial \ervu_{D}} \\
    \vdots & \ddots & \vdots\\
    \frac{\partial T_D}{\partial \ervu_{1}} & \cdots & \frac{\partial T_D}{\partial \ervu_{D}} \\
\end{bmatrix}.
\end{equation}
In practice, we often construct a flow-based model by implementing $T$ (or $T^{-1}$) with a neural network and taking $\pu(\rvu)$ to be a simple density such as a multivariate normal.  In Sections \ref{sec:constructing_discrete} and \ref{sec:constructing_continuous} we will discuss in detail how to implement $T$ (or $T^{-1}$).

Intuitively, we can think of the transformation $T$ as warping the space $\sR^D$ in order to mold the density $\pu(\rvu)$ into $\px(\rvx)$. The absolute Jacobian determinant $\abs{\det J_T(\rvu)}$ quantifies the relative change of volume of a small neighbourhood around $\rvu$ due to $T$. Roughly speaking, take $\mathrm{\mathbf{d}}\rvu$ to be an (infinitesimally) small neighbourhood around $\rvu$ and $\mathrm{\mathbf{d}}\rvx$ to be the small neighbourhood around $\rvx$ that $\mathrm{\mathbf{d}}\rvu$ maps to.  We then have that $\abs{\det J_T(\rvu)} \approx \text{Vol}(\mathrm{\mathbf{d}}\rvx) / \text{Vol}(\mathrm{\mathbf{d}}\rvu)$, the volume of $\mathrm{\mathbf{d}}\rvx$ divided by the volume of $\mathrm{\mathbf{d}}\rvu$.  The probability mass in $\mathrm{\mathbf{d}}\rvx$ must  equal the probability mass in $\mathrm{\mathbf{d}}\rvu$.  So, if $\mathrm{\mathbf{d}}\rvu$ is expanded, then the density at $\rvx$ is smaller than the density at $\rvu$.  If $\mathrm{\mathbf{d}}\rvu$ is contracted, then the density at $\rvx$ is larger.

An important property of invertible and differentiable transformations is that they are \emph{composable}.  Given two such transformations $T_1$ and $T_2$, their composition $T_2 \circ T_1$ is also invertible and differentiable. Its inverse and Jacobian determinant are given by:
\begin{align}
    \br{T_2 \circ T_1}^{-1} &= T_1^{-1} \circ T_2^{-1}\\
    \det J_{T_2 \circ T_1}(\rvu) &= \det J_{T_2}\br{T_1(\rvu)}\cdot\det J_{T_1}(\rvu).
\end{align}
In consequence, we can build complex transformations by composing multiple instances of simpler transformations, without compromising the requirements of invertibility and differentiability, and hence without losing the ability to calculate the density $\px(\rvx)$.

\begin{figure}[t]
    \centering
    \includegraphics[width=\textwidth]{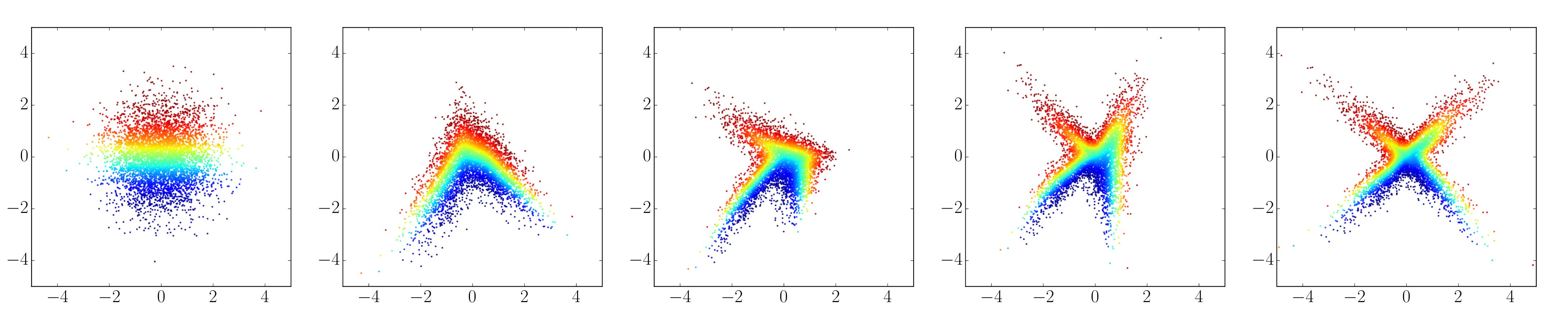}
    \caption{Example of a $4$-step flow transforming samples from a standard-normal base density to a cross-shaped target density. \label{fig:flow_samples_demo}} 
\end{figure}

In practice, it is common to chain together multiple transformations $T_1, \ldots, T_K$ to obtain $T = T_K \circ \cdots \circ T_1$, where each $T_k$ transforms $\rvz_{k-1}$ into $\rvz_{k}$, assuming $\rvz_{0} = \rvu$ and $\rvz_{K} = \rvx$. Hence, the term `flow' refers to the trajectory that a collection of samples from $\pu(\rvu)$ follow as they are gradually transformed by the sequence of transformations $T_1, \ldots, T_K$. The term `normalizing' refers to the fact that the inverse flow through $T_K^{-1}, \ldots, T_1^{-1}$ takes a collection of samples from $\px(\rvx)$ and transforms them (in a sense, `normalizes' them) into a collection of samples from a prescribed density $\pu(\rvu)$ (which is often taken to be a multivariate normal). \Cref{fig:flow_samples_demo} illustrates a flow ($K=4$) transforming a standard-normal base distribution to a cross-shaped target density.

In terms of functionality, a flow-based model provides two operations: sampling from the model via \Cref{eq:flow_sampling}, and evaluating the model's density via \Cref{eq:flow_evaluating_density_Tinv}. These  operations have different computational requirements.  Sampling from the model requires the ability to sample from $\pu(\rvu)$ and to compute the forward transformation $T$.  Evaluating the model's density requires computing the inverse transformation $T^{-1}$ and its Jacobian determinant, and evaluating the density $\pu(\rvu)$. The application will dictate which of these operations need to be implemented and how efficient they need to be. We  discuss the computational trade-offs associated with various implementation choices in Sections \ref{sec:constructing_discrete} and \ref{sec:constructing_continuous}.

\subsection{Expressive Power of Flow-Based Models}
\label{sec:flows:expressivity}

Before discussing the particulars of flows, a question of foremost importance is: how expressive are flow-based models? Can they represent \emph{any} distribution $\px(\rvx)$, even if the base distribution is restricted to be simple?  We show that this universal representation is possible under reasonable conditions on $\px(\rvx)$.  Specifically, we will show that for any pair of well-behaved distributions $\px(\rvx)$ (the target) and $\pu(\rvu)$ (the base), there exists a diffeomorphism that can turn $\pu(\rvu)$ into $\px(\rvx)$. The argument is constructive, and is based on a similar proof of the existence of non-linear ICA by \citet{hyvarinen1999nonlinearica}; a more formal treatment is provided by e.g.~\citet{Bogachev:2005:triangular}.

Suppose that $\px(\rvx) > 0$ for all $\rvx\in\sR^D$, and assume that all conditional probabilities \mbox{$\prob(\ervx_i' \le \ervx_i\g \rvx_{<i})$}---with $\ervx_i'$ being the random variable this probability refers to---are differentiable with respect to $(\ervx_i, \rvx_{< i})$. Using the chain rule of probability, we can decompose $\px(\rvx)$ into a product of conditional densities as follows:
\begin{equation}
    \px(\rvx) = \prod_{i=1}^D \px(\ervx_i\g \rvx_{<i}).
\end{equation}
Since $\px(\rvx)$ is non-zero everywhere, $\px(\ervx_i\g \rvx_{<i})>0$ for all $i$ and $\rvx$. Next, define the transformation $F: \rvx \mapsto \rvz\in(0, 1)^D $ whose $i$-th element is given by the \emph{cumulative distribution function} of the $i$-th conditional:
\begin{equation}
    \ervz_i = F_i(\ervx_i, \rvx_{<i}) =  \integralab{\px(\ervx_i'\g \rvx_{<i})}{\ervx_i'}{-\infty}{\ervx_i} = \prob(\ervx_i' \le \ervx_i\g \rvx_{<i}).
\end{equation}
Since each $F_i$ is differentiable with respect to its inputs, $F$ is differentiable with respect to $\rvx$. Moreover, each $F_i(\cdot, \rvx_{<i}):\R\rightarrow(0,1)$ is invertible, since its derivative $\frac{\partial F_i}{\partial\ervx_i} =  \px(\ervx_i\g \rvx_{<i})$ is positive everywhere. Because $\ervz_i$ doesn't depend on $\ervx_j$ for $i<j$, that implies we can invert $F$, with its inverse $F^{-1}$ given element-by-element as follows:
\begin{equation}
    \ervx_i = \br{F_i(\cdot, \rvx_{<i})}^{-1} (\ervz_i)
    \quad\text{for }i=1, \ldots, D.
\end{equation}
The Jacobian of $F$ is lower triangular since $\frac{\partial F_i}{\partial\ervx_j} = 0$ for $i<j$. Hence, the Jacobian determinant of $F$ is equal to the product of its diagonal elements:
\begin{equation}
    \det J_F(\rvx) = \prod_{i=1}^D \frac{\partial F_i}{\partial\ervx_i} = \prod_{i=1}^D \px(\ervx_i\g \rvx_{<i}) = \px(\rvx).
\end{equation}
Since $\px(\rvx) >0$, the Jacobian determinant is non-zero everywhere. Therefore, the inverse of $J_F(\rvx)$ exists, and is equal to the Jacobian of  $F^{-1}$, so $F$ is a diffeomorphism. Using a change of variables, we can calculate the density of $\rvz$ as follows:
\begin{equation}
    \pz(\rvz) = \px(\rvx)\,\abs{\det J_F(\rvx)}^{-1} = \px(\rvx)\,\abs{\px(\rvx)}^{-1} = 1,
\end{equation}
which implies $\rvz$ is distributed uniformly in the open unit cube $(0, 1)^D$.

The above argument shows that a flow-based model can express any distribution $\px(\rvx)$ (satisfying the conditions stated above) even if we restrict the base distribution to be uniform in $(0, 1)^D$. We can extend this statement to any base distribution $\pu(\rvu)$ (satisfying the same conditions as $\px(\rvx)$) by first transforming $\rvu$ to a uniform $\rvz\in(0, 1)^D$ as an intermediate step. In particular, given any $\pu(\rvu)$ that satisfies the above conditions, define $G$ to be the following transformation:
\begin{equation}
    \ervz_i = G_i(\ervu_i, \rvu_{<i}) =  \integralab{\pu(\ervu_i'\g \rvu_{<i})}{\ervu_i'}{-\infty}{\ervu_i} = \prob(\ervu_i' \le \ervu_i\g \rvu_{<i}).
\end{equation}
By the same argument as above, $G$ is a diffeomorphism, and $\rvz$ is uniformly distributed in $(0, 1)^D$. Thus, a flow with transformation $T = F^{-1}\circ G$ can turn $\pu(\rvu)$ into $\px(\rvx)$.

\subsection{Using Flows for Modeling and Inference}

Similarly to fitting any probabilistic model, fitting a flow-based model $\px(\rvx; \vtheta)$ to a target distribution $\px^{*}(\rvx)$ can be done by minimizing some divergence or discrepancy between them. This minimization is performed with respect to the model's parameters $\vtheta = \{ \vphi, \vpsi \}$, where $\vphi$ are the parameters of $T$ and $\vpsi$ are the parameters of $\pu(\rvu)$. In the following sections, we discuss a number of divergences for fitting flow-based models, with a particular focus on the Kullback--Leibler (KL) divergence as it is one of the most popular choices.

\subsubsection{Forward KL Divergence and Maximum Likelihood Estimation} \label{sec:mle}

The forward KL divergence between the target distribution $\px^{*}(\rvx)$ and the flow-based model $\px(\rvx; \vtheta)$  can be written as follows:
\begin{equation}\begin{split}
    \mathcal{L}(\vtheta) &= \kl{\px^{*}(\rvx)}{\px(\rvx; \vtheta)} \\
    &= -\avg{\log \px(\rvx; \vtheta) }{\px^{*}(\rvx)} + \mathrm{const}. \\
    &= -\avg{\log \pu\br{T^{-1}(\rvx; \vphi); \vpsi } + \log \abs{\det J_{T^{-1}}(\rvx; \vphi) }}{\px^{*}(\rvx)} + \mathrm{const}.
\end{split}\end{equation}
The forward KL divergence is well-suited for situations in which we have samples from the target distribution (or the ability to generate them), but we cannot necessarily evaluate the target density $\px^{*}(\rvx)$. Assuming we have a set of samples $\{ \rvx_{n} \}_{n=1}^{N}$ from $\px^{*}(\rvx)$, we can estimate the expectation over $\px^{*}(\rvx)$ by Monte Carlo as follows:
\begin{equation}
    \mathcal{L}(\vtheta)
    \approx - \frac{1}{N} \sum_{n=1}^{N} \log \pu(T^{-1}\br{\rvx_{n}; \vphi); \vpsi} + \log  \left| \det J_{T^{-1}}(\rvx_{n}; \vphi)  \right| + \mathrm{const}.
\end{equation} 
Minimizing the above Monte Carlo approximation of the KL divergence is equivalent to fitting the flow-based model to the samples $\{ \rvx_{n} \}_{n=1}^{N}$ by maximum likelihood estimation.

In practice, we often optimize the parameters $\vtheta$ iteratively with stochastic gradient-based methods. We can obtain an unbiased estimate of the gradient of the KL divergence with respect to the parameters as follows:
\begin{align}
    &\nabla_{\vphi}\mathcal{L}(\vtheta) \approx
    -\frac{1}{N} \sum_{n=1}^{N} \nabla_{\vphi}\log \pu\br{T^{-1}(\rvx_{n}; \vphi); \vpsi}   + \nabla_{\vphi}\log  \left| \det J_{T^{-1}}(\rvx_{n}; \vphi)  \right|\\
    &\nabla_{\vpsi}\mathcal{L}(\vtheta) \approx
    -\frac{1}{N} \sum_{n=1}^{N} \nabla_{\vpsi}\log \pu\br{T^{-1}(\rvx_{n}; \vphi); \vpsi}.
\end{align}
The update with respect to $\vpsi$ may also be done in closed form if $\pu(\rvu; \vpsi)$ admits closed-form maximum likelihood estimates, as is the case for example with Gaussian distributions.

In order to fit a flow-based model via maximum likelihood, we need to compute $T^{-1}$, its Jacobian determinant and the density $\pu(\rvu; \vpsi)$---as well as differentiate through all three, if using gradient-based optimization. That means we can train a flow model with maximum likelihood even if we are not able to compute $T$ or sample from $\pu(\rvu; \vpsi)$.  Yet these operations will be needed if we want to sample from the model after it is fitted.

\subsubsection{Reverse KL Divergence}

Alternatively, we may fit the flow-based model by minimizing the reverse KL divergence, which can be written as follows:
\begin{equation}\begin{split}
    \mathcal{L}(\vtheta) &= \kl{\px(\rvx; \vtheta)}{\px^{*}(\rvx)} \\
    &= \avg{\log \px(\rvx; \vtheta)  - \log \px^{*}(\rvx)}{\px(\rvx; \vtheta) } \\
    &= \avg{\log \pu\br{\rvu; \vpsi} - \log\abs{\det J_{T}(\rvu; \vphi)} - \log \px^{*}\br{T(\rvu; \vphi)}}{\pu(\rvu; \vpsi) }.
\end{split}\end{equation}
We made use of a change of variable in order to express the expectation with respect to $\rvu$. The reverse KL divergence is suitable when we have the ability to evaluate the target density $\px^{*}(\rvx)$ but not necessarily sample from it. In fact, we can minimize $\mathcal{L}(\vtheta)$ even if we can only evaluate $\px^{*}(\rvx)$ up to a multiplicative normalizing constant $C$, since in that case $\log C$ will be an additive constant in the above expression for $ \mathcal{L}(\vtheta)$. We may therefore assume that $\px^{*}(\rvx) = \ptildex(\rvx) / C$, where $\ptildex(\rvx)$ is tractable but $C = \integral{\ptildex(\rvx)}{\rvx}$ is not, and rewrite the reverse KL divergence as:
\begin{equation}
    \mathcal{L}(\vtheta) =
    \avg{\log \pu\br{\rvu; \vpsi} - \log\abs{\det J_{T}(\rvu; \vphi)} - \log \ptildex\br{T(\rvu; \vphi)}}{\pu(\rvu; \vpsi) } + \mathrm{const}.
\end{equation}

In practice, we can minimize $\mathcal{L}(\vtheta)$ iteratively with stochastic gradient-based methods. Since we reparameterized the expectation to be with respect to the base distribution $\pu(\rvu; \vpsi)$, we can easily obtain an unbiased estimate of the gradient of $\mathcal{L}(\vtheta)$ with respect to $\vphi$ by Monte Carlo. In particular, let $\set{\rvu_n}_{n=1}^{N}$ be a set of samples from $\pu(\rvu; \vpsi)$; the gradient of $\mathcal{L}(\vtheta)$ with respect to $\vphi$ can be estimated as follows:
\begin{equation}
    \nabla_{\vphi}\mathcal{L}(\vtheta) \approx
    -\frac{1}{N} \sum_{n=1}^{N} \nabla_{\vphi}\log  \left| \det J_{T}(\rvu_{n}; \vphi)  \right|
    + \nabla_{\vphi}\log \ptildex\br{T(\rvu_n; \vphi)}.
\end{equation}
Similarly, we can estimate the gradient with respect to $\vpsi$ by reparameterizing $\rvu$ as:
\begin{equation}
    \rvu = T'(\rvu'; \vpsi)\quad\text{where}\quad\rvu'\sim \puprime(\rvu'),
\end{equation}
and then writing the expectation with respect to $\puprime(\rvu')$. However, since we can equivalently absorb the reparameterization $T'$ into $T$ and replace the base distribution with $\puprime(\rvu')$, we can assume without loss of generality that the parameters $\vpsi$ are fixed and only optimize with respect to $\vphi$.

In order to minimize the reverse KL divergence as described above, we need to be able to sample from the base distribution $\pu(\rvu; \vpsi)$ as well as compute and differentiate through the transformation $T$ and its Jacobian determinant. That means that we can fit a flow-based model by minimizing the reverse KL divergence even if we cannot evaluate the base density or compute the inverse transformation $T^{-1}$. However, we will need these operations if we would like to evaluate the density of the trained model.

The reverse KL divergence is often used for \emph{variational inference} \citep{wainwright2008graphical, blei2017variational}, a form of approximate Bayesian inference.  In this case, the target is the posterior, making $\ptildex(\rvx)$ the product between a likelihood function and a prior density.  Examples of work using flows in variational inference are given by \citet{rezende2015variational, van2018sylvester, iaf, tomczak2016householder, louizos2017multiplicative}.  We cover this topic in more detail in \Cref{sec:applications:variational_inference}.   

Another application of the reverse KL divergence is in the context of \emph{model distillation}: a flow model is trained to replace a target model $\px^{*}\br{\rvx}$ whose density can be evaluated but that is otherwise inconvenient. An example of model distillation with flows is given by \citet{oord2017parallel}. In their case, samples cannot be efficiently drawn from the target model and so they distill it into a flow that supports fast sampling.

\subsubsection{Relationship Between Forward and Reverse KL Divergence}

An alternative perspective of a flow-based model is to think of the target $\px^*(\rvx)$ as the \emph{base distribution} and the inverse flow as inducing a distribution $\pu^*(\rvu; \vphi)$.  Intuitively, $\pu^*(\rvu; \vphi)$ is the distribution that the training data would follow when passed through $T^{-1}$. Since the target and the base distributions uniquely determine each other given the transformation between them, $\pu^*(\rvu; \vphi) = \pu(\rvu; \vpsi)$ if and only if $\px^*(\rvx) = \px(\rvx;\vtheta)$. Therefore, fitting the model $\px(\rvx;\vtheta)$ to the target $\px^*(\rvx)$ can be equivalently thought of as fitting the induced distribution $\pu^*(\rvu; \vphi)$ to the base $\pu(\rvu; \vpsi)$.

We may now ask: how does fitting $\px(\rvx;\vtheta)$ to the target relate to fitting $\pu^*(\rvu; \vphi)$ to the base? Using a change of variables (see \Cref{app:kl_duality} for details), we can show the following equality \citep{papamakarios2017masked}:
\begin{align}
    \kl{\px^*(\rvx)}{\px(\rvx;\vtheta)} = \kl{\pu^*(\rvu; \vphi)}{\pu(\rvu; \vpsi)}.
\end{align}
The above equality means that fitting the model to the target using the forward KL divergence (maximum likelihood) is equivalent to fitting the induced distribution $\pu^*(\rvu; \vphi)$ to the base $\pu(\rvu; \vpsi)$ under the reverse KL divergence. In \Cref{app:kl_duality}, we also show that:
\begin{align}
    \kl{\px(\rvx;\vtheta)}{\px^*(\rvx)} = \kl{\pu(\rvu; \vpsi)}{\pu^*(\rvu; \vphi)},
\end{align}
which means that fitting the model to the target via the reverse KL divergence is equivalent to fitting $\pu^*(\rvu; \vphi)$ to the base via the forward KL divergence (maximum likelihood).

\subsubsection{Alternative Divergences}\label{sec:alternative_divergences}

Learning the parameters of flow models is not restricted to the use of the KL divergence. Many alternative measures of difference between distributions are available. These alternatives are often grouped into two general families, the \textit{$f$-divergences} that use density ratios to compare distributions, and the \textit{integral probability metrics} (IPMs) that use differences for comparison:
\begin{align}
   & f\textrm{-divergence} &  D_f\left[\,\px^*(\rvx) \,\|\, \px(\rvx;\vtheta)\,\right]  &= \avg{f\left(\frac{\px^*(\rvx)}{\px(\rvx;\vtheta)} \right)}{\px(\rvx;\vtheta)} \label{eq:fdiv}\\
& \textrm{IPM} & \delta_{s}\left[\,\px^*(\rvx) \,\|\, \px(\rvx;\vtheta)\,\right]  &= \avg{s(\rvx)}{\px^*(\rvx)} - \avg{s(\rvx)}{\px(\rvx;\vtheta)}. \label{eq:ipm}
\end{align}
For the $f$-divergences, the function $f$ is convex; when this function is $f(r) = r\log r$
we recover the KL divergence. For IPMs, the function $s$ can be chosen from a set of test statistics, or can be a witness function chosen adversarially.

The same considerations that applied to the KL divergence previously apply here and inform the choice of divergence: can we simulate from the model $\px(\rvx; \vtheta)$, do we know the true distribution up to a multiplicative constant, do we have access to the transform or its inverse? When considering these divergences, we see a connection between flow-based models, whose design principles use composition and change of variables, to the more general class of implicit probabilistic models
\citep{diggle1984monte, mohamed2016learning}. 
If we choose the generator of a generative adversarial network as a normalizing flow, we can train the flow parameters using adversarial training \citep{grover2018flow, danihelka2017comparison}, with Wasserstein losses \citep{arjovsky2017wasserstein}, using maximum mean discrepancy \citep{binkowski2018demystifying}, or other approaches.

\subsection{Brief Historical Overview}

\textit{Whitening transformations} \citep{johnson1966, friedman1987exploratory}---so-called as they transform data into white noise---are the clearest intellectual predecessor to the use of normalizing flows within machine learning.  \citet{chen2001gaussianization} were perhaps the first to use whitening as a density estimation technique rather than for feature pre-processing, calling the method \textit{Gaussianization}. 
\citet{tabak2010density} approached Gaussianization from the perspective of diffusion processes, making connections to statistical mechanics---specifically, using Liouville's equation to characterize the flow. 
In a follow-up paper, \citet{tabak2013family} introduce what can be thought of as the modern conception of normalizing flows: introducing the very term \textit{normalizing flow} and defining the flow generally as a composition of $K$ simple maps. As we will discuss in \Cref{sec:constructing_discrete}, this definition via composition is essential for enabling flows to be expressive while preserving computational and analytical tractability. 

The idea of composition saw its recent emergence in machine learning starting with \citet{rippel2013high}, who were perhaps the first to recognize that parameterizing flows with deep neural networks could result in quite general and expressive distribution classes. \citet{dinh2014nice} then introduced a scalable and computationally efficient architecture, demonstrating further improvements to image modeling and inference. \citet{rezende2015variational} used the idea and language from \citet{tabak2013family} to apply normalizing flows in the setting of variational inference. Following this, as the papers that are reviewed here will show, normalizing flows now constitute a broad literature in their own right, with much work expanding the applicability and scalability of these initial efforts.

Because of the connection to change of variables, we can also connect the current use of normalizing flows in machine learning with its pre-history in many other settings. The change of measure has been studied intensely in the development of statistical mechanics---a famous example being the aforementioned \textit{Liouville's theorem} (1838). \textit{Copulas} \citep{sklar1959fonctions, gal2013copulas} can be viewed as rudimentary flows, where each dimension is transformed independently using the empirically-estimated marginal cumulative distribution function.  Optimal transport and the Wasserstein metric \citep{villani2008optimal} can also be formulated in terms of transformations of measures (`transport of measures')---also known as the \textit{Monge problem}.
In particular, \textit{triangular maps} (a concept deeply related to autoregressive flows) can be shown to be a limiting solution to a class of Monge--Kantorovich problems \citep{carlier2010knothe}.  This class of triangular maps itself has a long history, with \citet{rosenblatt1952remarks} studying their properties for transforming multivariate distributions uniformly over the hypercube.  Optimal transport could be a tutorial unto itself and therefore we mostly sidestep this framework, instead choosing to think in terms of the change of variables.

\section{Constructing Flows Part I: Finite Compositions}
\label{sec:constructing_discrete}

\begin{figure}[t]
    \centering
    \includegraphics[width=\textwidth]{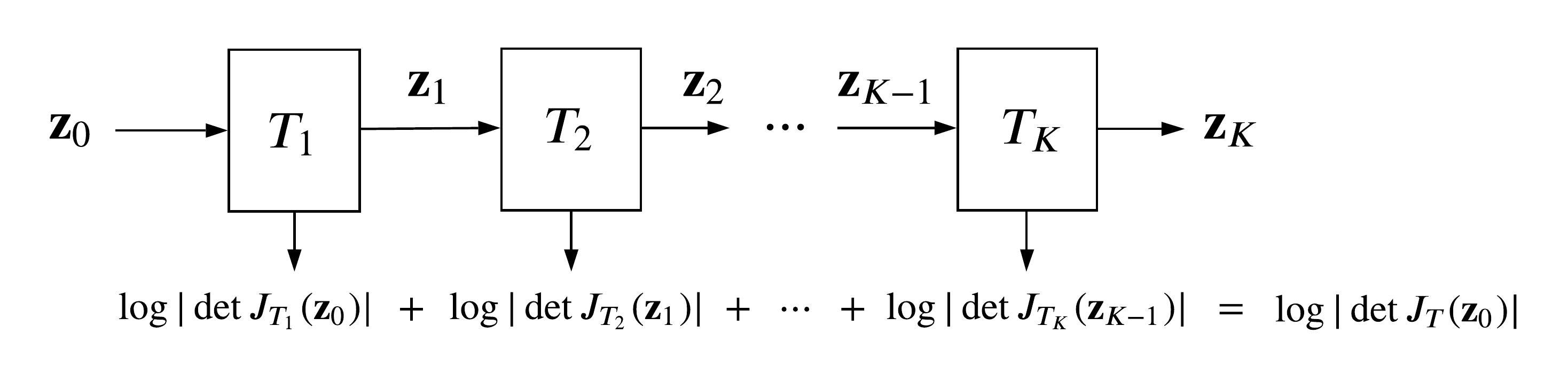}
    \caption{Illustration of a flow composed of $K$ transformations. \label{fig:flow_composition}} 
\end{figure}

Having described some high-level properties and uses of flows, we transition into describing, categorizing, and unifying the various ways to construct a flow. As discussed in \Cref{sec:flows:definition}, normalizing flows are \emph{composable}; that is, we can construct a flow with transformation $T$ by composing a finite number of simple transformations $T_k$ as follows:
\begin{equation}
    T = T_{K} \circ \cdots \circ T_{1}.
\end{equation}
The idea is to use simple transformations as building blocks---each having a tractable inverse and Jacobian determinant---to define a complex transformation with more expressive power than any of its constituent components. Importantly, the flow's forward and inverse evaluation and Jacobian-determinant computation can be localized to the sub-flows.  As illustrated in \Cref{fig:flow_composition}, assuming $\rvz_0=\rvu$ and $\rvz_K=\rvx$, the forward evaluation is:
\begin{align}
    \rvz_{k} = T_k(\rvz_{k-1})\quad\text{ for }k=1,\ldots,K,
\end{align}
the inverse evaluation is:
\begin{align}
    &\rvz_{k-1} = T_k^{-1}(\rvz_{k})\quad\text{ for }k=K,\ldots,1,
\end{align}
and the Jacobian-determinant computation (in the log domain) is:
\begin{equation}
\log \left| \det J_{T}(\rvz_0)  \right| = \log \abs{\prod_{k=1}^{K} \det J_{T_{k}}(\rvz_{k-1})} = \sum_{k=1}^{K} \log \left| \det J_{T_{k}}(\rvz_{k-1})  \right|.
\end{equation}
Increasing the `depth' (i.e.\ number of composed sub-flows) of the transformation crucially results in only $\bigo{K}$ growth in the computational complexity---a pleasantly practical cost to pay for the increased expressivity.

\begin{table}[t]
    \renewcommand{\arraystretch}{1.5}
    \aboverulesep = 2mm
    \begin{tabularx}{\textwidth}{p{4cm}p{5cm}X}
        \toprule
        Autoregressive flows &
        \begin{minipage}[t]{0.3\textwidth}
            Transformer type:
            {
            \begin{itemize}
                \item[--] Affine
                \item[--] Combination-based
                \item[--] Integration-based
                \item[--] Spline-based
            \end{itemize}}
        \end{minipage} &
        \begin{minipage}[t]{0.3\textwidth}
            Conditioner type:
            \begin{itemize}
                \item[--] Recurrent
                \item[--] Masked
                \item[--] Coupling layer
            \end{itemize}
        \end{minipage}\\
        \midrule
        Linear flows &
        Permutations & \\
        & \begin{minipage}[t]{0.3\textwidth}
            Decomposition-based:
            \begin{itemize}
                \item[--] PLU
                \item[--] QR
            \end{itemize}
        \end{minipage} &\\
        & \begin{minipage}[t]{0.3\textwidth}
            Orthogonal:
            \begin{itemize}
                \item[--] Exponential map
                \item[--] Cayley map
                \item[--] Householder
            \end{itemize}
        \end{minipage} &\\
        \midrule
        Residual flows &
        Contractive residual &\\
        & \begin{minipage}[t]{0.5\textwidth}
            Based on matrix determinant lemma:
            \begin{itemize}
                \item[--] Planar
                \item[--] Sylvester
                \item[--] Radial
            \end{itemize}
        \end{minipage} &\\
        \bottomrule
    \end{tabularx}
    \caption{Overview of methods for constructing flows based on finite compositions.}
    \label{table:flows_overview}
\end{table}

In practice we implement either $T_k$ or $T_k^{-1}$ using a model (such as a neural network) with parameters $\vphi_k$, which we will denote as $f_{\vphi_k}$. That is, we can take the model $f_{\vphi_k}$ to implement either $T_k$, in which case it will take in $\rvz_{k-1}$ and output $\rvz_k$, or  $T_k^{-1}$, in which case it will take in $\rvz_{k}$ and output $\rvz_{k-1}$. In either case, we must ensure that the model is invertible and has a tractable Jacobian determinant. In the rest of this section, we will describe several approaches for constructing $f_{\vphi_k}$ so that these requirements are satisfied. An overview of all the methods discussed in this section  is shown in \Cref{table:flows_overview}.

Ensuring that $f_{\vphi_k}$ is invertible and explicitly calculating its inverse are \emph{not} synonymous. In many implementations, even though the inverse of $f_{\vphi_k}$ is guaranteed to exist, it can be expensive or even intractable to compute exactly. As discussed in \Cref{sec:flows}, the forward transformation $T$ is used when sampling, and the inverse transformation $T^{-1}$ is used when evaluating densities. If the inverse of $f_{\vphi_k}$ is not efficient, either density evaluation or sampling will be inefficient or even intractable, depending on whether $f_{\vphi_k}$ implements $T_k$ or $T_k^{-1}$.  Whether $f_{\vphi_k}$ should be designed to have an efficient inverse and whether it should be taken to implement $T_k$ or $T_k^{-1}$ are decisions that ought to be based on intended usage.

We should also clarify what we mean by `tractable Jacobian determinant'. We can always compute the Jacobian matrix of a differentiable function with $D$ inputs and $D$ outputs, using $D$ passes of either forward-mode or reverse-mode automatic differentiation. Then, we can explicitly calculate the determinant of that Jacobian. However, this computation has a time cost of $\bigo{D^3}$, which can be intractable for large $D$. For most applications of flow-based models, the Jacobian-determinant computation should be at most $\bigo{D}$. Hence, in the following sections, we will describe functional forms that allow the Jacobian determinant to be computed in linear time with respect to the input dimensionality.

To simplify notation from here on, we will drop the dependence of the model parameters on $k$ and denote the model by $f_{\vphi}$. Also, we will denote the model's input by $\rvz$ and its output by $\rvz'$, regardless of whether the model implements $T_k$ or $T_k^{-1}$.

\subsection{Autoregressive Flows}\label{sec:autoregressive_flows}

\begin{figure}[t]
    \centering
    \begin{subfigure}[b]{0.49\textwidth}
        \includegraphics[width=\textwidth]{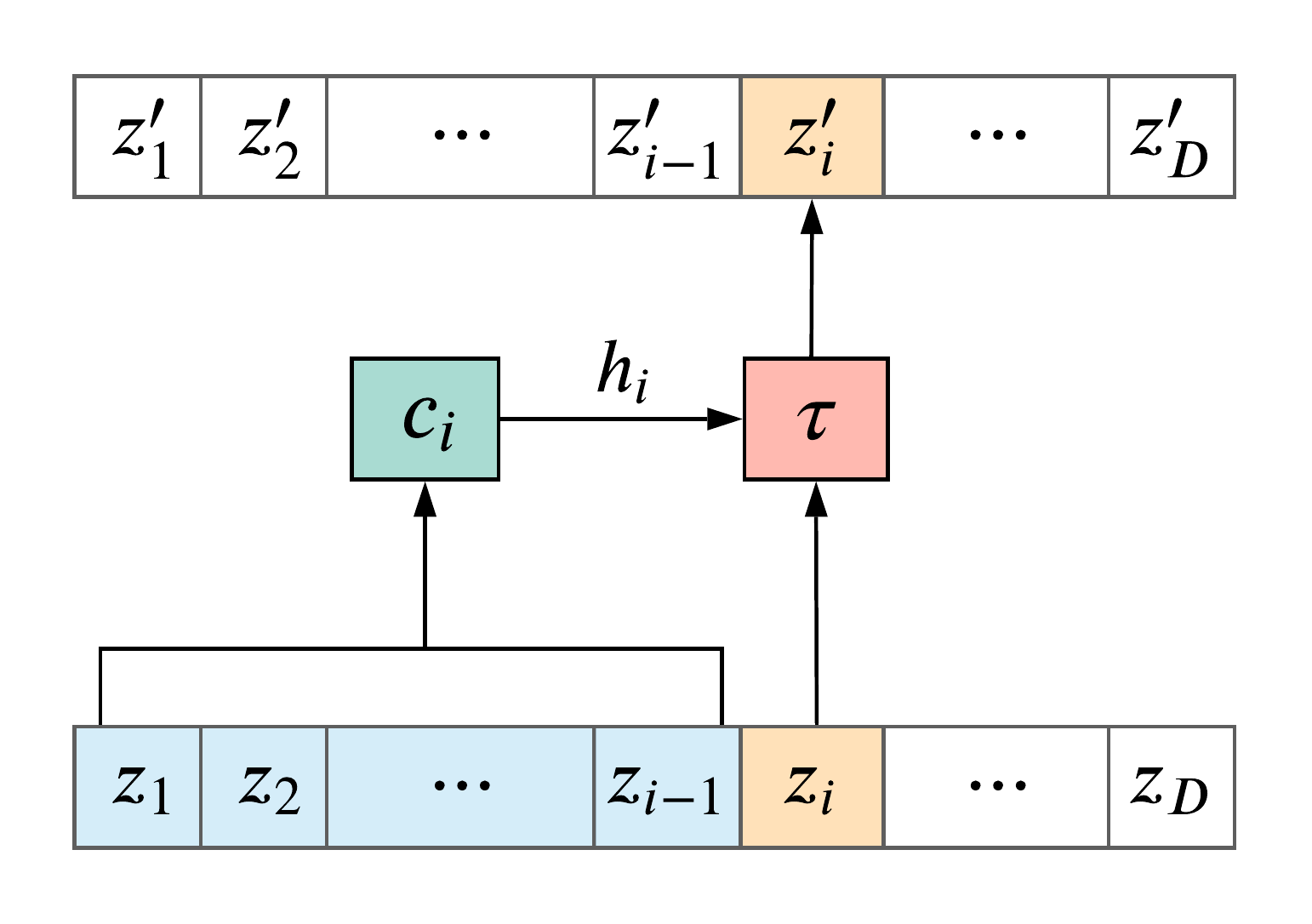}
        \caption{Forward\label{fig:autoregressive_flow:forward}}
    \end{subfigure}
    \begin{subfigure}[b]{0.49\textwidth}
        \includegraphics[width=\textwidth]{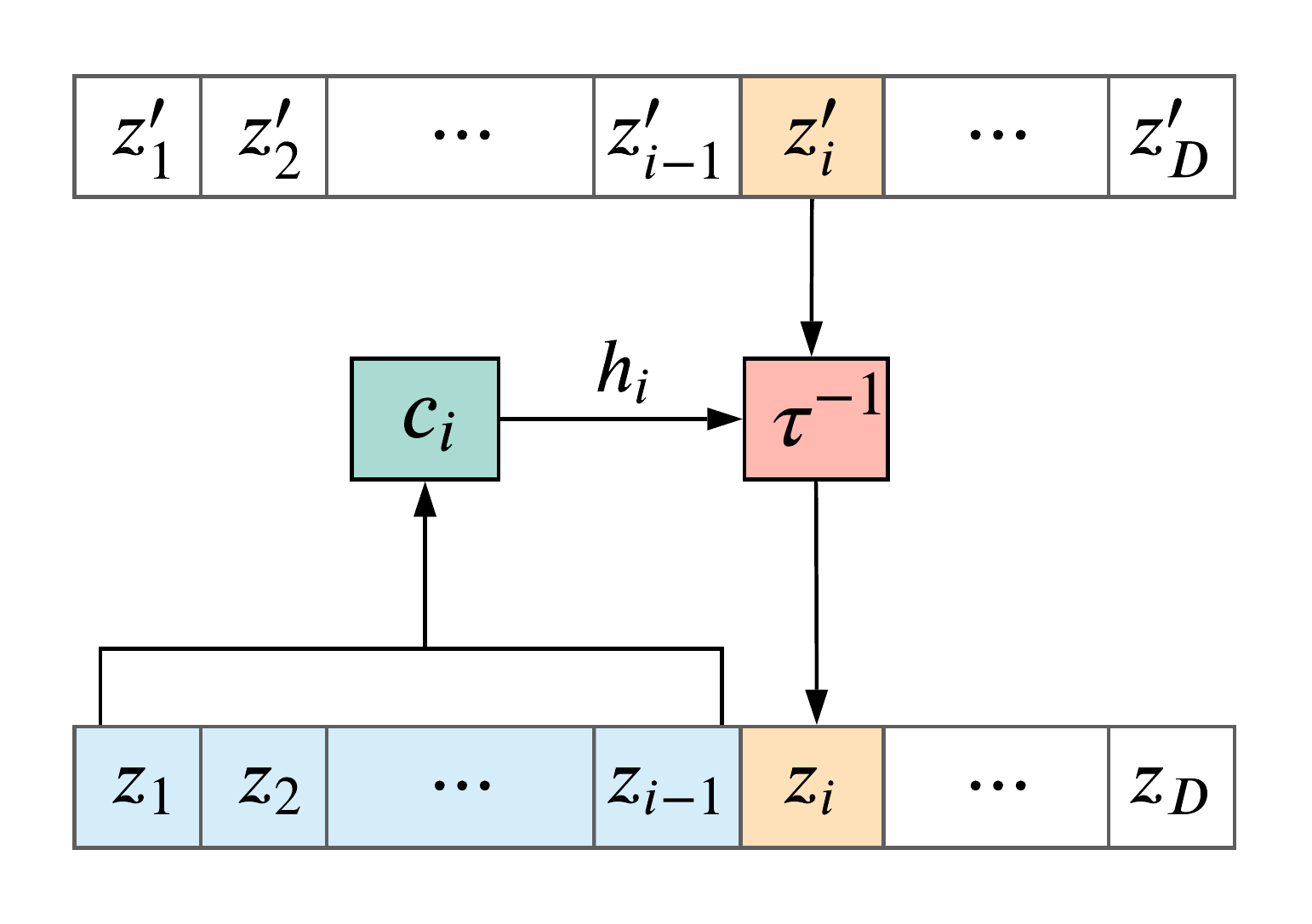}
        \caption{Inverse\label{fig:autoregressive_flow:inverse}}
    \end{subfigure}
    \caption{Illustration of the $i$-th step of an autoregressive flow. \label{fig:autoregressive_flow}} 
\end{figure}

Autoregressive flows were one of the first classes of flows developed and remain among the most popular.
In \Cref{sec:flows:expressivity} we saw that, under mild conditions, we can transform any distribution $\px(\rvx)$ into a uniform distribution in $(0, 1)^D$ using maps with a triangular Jacobian. 
Autoregressive flows are a direct implementation of this construction, specifying $f_{\vphi}$ to have the following form \citep[as described by e.g.][]{huang2018neural, jaini2019sum}:
\begin{equation}\label{eq:ar_flow_forward}
\ervz_{i}' = \tau(\ervz_{i}; \vh_i)
\quad\text{where}\quad
\vh_i = c_{i}(\rvz_{<i}),
\end{equation}
where $\tau$ is termed the \textit{transformer} and $c_i$ the $i$-th \textit{conditioner}. This is illustrated in \Cref{fig:autoregressive_flow:forward}. The transformer is a strictly monotonic function of $\ervz_i$ (and therefore invertible), is parameterized by $\vh_i$, and specifies how the flow acts on $\ervz_i$ in order to output $\ervz_i'$.
The conditioner determines the parameters of the transformer, and in turn, can modify the transformer's behavior.  The conditioner does \emph{not} need to be a bijection. Its one constraint is that the $i$-th conditioner can take as input only the variables with dimension indices less that $i$. The parameters $\vphi$ of $f_\vphi$ are typically the parameters of the conditioner (not shown above for notational simplicity), but sometimes the transformer has its own parameters too (in addition to $\vh_i$).

It is easy to check that the above construction is invertible for any choice of $\tau$ and $c_i$ as long as the transformer is invertible. Given $\rvz'$, we can compute $\rvz$ iteratively as follows:
\begin{equation}\label{eq:ar_flow_inverse}
    \ervz_i = \tau^{-1}(\ervz_i'; \vh_i)
    \quad\text{where}\quad
    \vh_i = c_{i}(\rvz_{<i}).
\end{equation}
This is illustrated in \Cref{fig:autoregressive_flow:inverse}. In the forward computation, each $\vh_i$ and therefore each $\ervz_i'$ can be computed independently in any order or in parallel. In the inverse computation however, all $\rvz_{<i}$ need to have been computed \emph{before} $\ervz_{i}$, so that $\rvz_{<i}$ is available to the conditioner for computing $\vh_i$.

It is also easy to show that the Jacobian of the above transformation is \emph{triangular}, and thus the Jacobian determinant is tractable. Since each $\ervz_i'$ doesn't depend on $\rvz_{> i}$, the partial derivative of $\ervz_i'$ with respect to $\ervz_j$ is zero whenever $j>i$. Hence, the Jacobian of $f_{\vphi}$ can be written in the following form:
\begin{equation}\label{eq:ar_flow_jacobian}
 J_{f_{\vphi}}(\rvz)
= 
\begin{bmatrix}
    \frac{\partial \tau}{\partial \ervz_{1}}(\ervz_1; \vh_1) & & \mathbf{0} \\
    & \ddots & \\
    \rmL(\rvz) & & \frac{\partial \tau}{\partial \ervz_{D}}(\ervz_D; \vh_D) \\
\end{bmatrix}.
\end{equation}
The Jacobian is a lower-triangular matrix whose diagonal elements are the derivatives of the transformer for each of the $D$ elements of $\rvz$. Since the determinant of any triangular matrix is equal to the product of its diagonal elements, the log-absolute-determinant of $J_{f_{\vphi}}(\rvz)$ can be calculated in $\bigo{D}$ time as follows:
\begin{equation}
    \log\abs{\det{J_{f_{\vphi}}(\rvz)}} = 
    \log \abs{\prod_{i=1}^D \frac{\partial \tau}{\partial \ervz_{i}}(\ervz_i; \vh_i)} = 
    \sum_{i=1}^D \log \abs{\frac{\partial \tau}{\partial \ervz_{i}}(\ervz_i; \vh_i)}.
\end{equation}
The lower-triangular part of the Jacobian--- denoted here by $\rmL(\rvz)$---is irrelevant.  The derivatives of the transformer can be computed either analytically or via automatic differentiation, depending on the implementation. 

Autoregressive flows are \emph{universal approximators} (under the conditions discussed in \Cref{sec:flows:expressivity}) provided the transformer and the conditioner are flexible enough to represent any function arbitrarily well.  This follows directly from the fact that the universal transformation from \Cref{sec:flows:expressivity}, which is based on the cumulative distribution functions of the conditionals, is indeed an autoregressive flow.  Yet, this is just a statement of representational power and makes no guarantees about the flow's behavior in practice.

An alternative, but mathematically equivalent, formulation of autoregressive flows is to have the conditioner $c_i$ take in $\rvz_{<i}'$ instead of $\rvz_{<i}$. This is equivalent to swapping $\tau$ with $\tau^{-1}$ and $\rvz$ with $\rvz'$ in the formulation presented above. Both formulations are common in the literature; here we use the convention that $c_i$ takes in $\rvz_{<i}$ without loss of generality. The computational differences between the two alternatives are discussed in more detail by e.g.~\citet{iaf, papamakarios2017masked}. 

Implementing an autoregressive flow boils down to (a) implementing the transformer $\tau$ and (b) implementing the conditioner $c_i$. These are independent choices: any type of transformer can be paired up with any type of conditioner, yielding the various combinations that have appeared in the literature. In the following paragraphs, we will list a number of transformer implementations (\Cref{sec:ar_flows:transformer}) and an number of conditioner implementations (\Cref{sec:ar_flows:conditioner}). We will discuss their pros and cons, and point out the choices that have been used by the specific models in the literature.

\subsubsection{Implementing the Transformer}
\label{sec:ar_flows:transformer}

\paragraph{Affine transformers}  One of the simplest possible choices for the transformer---and one of the first used---is the class of affine functions:
\begin{equation}\label{eq:affine_ar_flow}
    \tau(\ervz_i; \vh_i) = \alpha_i\ervz_i + \beta_i
    \quad\text{where}\quad
    \vh_i = \set{\alpha_i, \beta_i}.
\end{equation}
The above can be thought of as a \emph{location-scale transformation}, where $\alpha_i$ controls the scale and $\beta_i$ controls the location. Invertibility is guaranteed if $\alpha_i\neq 0$, and this can be easily achieved by e.g.~taking $\alpha_i=\exp \tilde{\alpha}_i$, where $\tilde{\alpha}_i$ is an unconstrained parameter (in which case $\vh_i = \set{\tilde{\alpha}_i, \beta_i}$). The derivative of the transformer with respect to $\ervz_i$ is equal to $\alpha_i$; hence the log absolute Jacobian determinant is:
\begin{equation}
    \log\abs{\det{J_{f_{\vphi}}(\rvz)}} = \sum_{i=1}^D\log\abs{\alpha_i} = \sum_{i=1}^D\tilde{\alpha}_i.
\end{equation}
Autoregressive flows with affine transformers are attractive because of their simplicity and analytical tractability, but their expressivity is  limited. To illustrate why, suppose $\rvz$ follows a Gaussian distribution; then, each $\ervz_i'$ conditioned on $\rvz_{<i}'$ will also follow a Gaussian distribution. In other words, a single affine autoregressive transformation of a multivariate Gaussian results in a distribution whose conditionals $\pzprime(\ervz_i'\g\rvz_{<i}')$ will necessarily be Gaussian. Nonetheless, expressive flows can still be obtained by stacking multiple affine autoregressive layers, but it's unknown whether affine autoregressive flows with multiple layers are universal approximators or not. Affine transformers are popular in the literature, having been used in models such as NICE \citep{dinh2014nice}, Real NVP \citep{dinh2016density}, IAF \citep{iaf}, MAF \citep{papamakarios2017masked}, and Glow \citep{kingma2018glow}.

\paragraph{Combination-based transformers} Non-affine transformers can be constructed from simple components based on the observation that \textit{conic combinations} as well as \textit{compositions} of monotonic functions are also monotonic. Given monotonic functions $\tau_1, \ldots, \tau_K$ of a real variable $\ervz$, the following functions are also monotonic:
\begin{itemize}
\item Conic combination: $\tau(\ervz) = \sum_{k=1}^{K} w_k \tau_k(\ervz)$, where $w_k > 0$ for all $k$.
\item Composition: $\tau(\ervz) = \tau_{K}\circ\cdots\circ\tau_{1}(\ervz)$.
\end{itemize}
For example, a non-affine transformer can be constructed using a conic combination of monotonically increasing activation functions $\sigma(\cdot)$ (such as logistic sigmoid, tanh, leaky ReLU \citep{maas2013rectifier}, and others):
\begin{equation}\label{eq:non_affine_neural_transformer}
    \tau(\ervz_i; \vh_i) = w_{i0} + \sum_{k=1}^K w_{ik} \sigma(\alpha_{ik} \ervz_i + \beta_{ik})
    \quad\text{where}\quad
    \vh_{i} = \set{w_{i0}, \ldots, w_{iK}, \alpha_{ik}, \beta_{ik}},
\end{equation}
provided $\alpha_{ik}>0$ and $w_{ik} > 0$ for all $k\ge1$. Clearly, the above construction corresponds to a monotonic \textit{single-layer perceptron}. By repeatedly combining and composing monotonic activation functions, we can construct a \textit{multi-layer perceptron} that is monotonic, provided that all its weights are strictly positive.

Transformers such as the above can represent any monotonic function arbitrarily well, which follows directly from the universal-approximation capabilities of multi-layer perceptrons \citep[see e.g.][for details]{huang2018neural}. The derivatives of the transformer needed for the computation of the Jacobian determinant are in principle analytically obtainable, but more commonly they are computed via backpropagation. A drawback of combination-based transformers is that in general they cannot be inverted analytically, and can be inverted only iteratively  e.g.~using bisection search \citep{rburden81:numerical}. Variants of combination-based transformers have been used in models such as NAF \citep{huang2018neural}, block-NAF \citep{DeCao:2019:bnaf}, and Flow++ \citep{ho2019flow++}.

\paragraph{Integration-based transformers} Another way to define a non-affine transformer is by recognizing that the integral of some positive function is a monotonically increasing function. For example, \citet{Wehenkel:2019:umnn} define the transformer as:
\begin{equation}
    \tau(\ervz_i; \vh_i) = \integralab{g(\ervz; \valpha_i)}{\ervz}{0}{\ervz_i} + \beta_i
    \quad\text{where}\quad
    \vh_i = \set{\valpha_i, \beta_i},
\end{equation} where $g(\cdot; \valpha_i)$ can be any positive-valued neural network parameterized by $\valpha_i$. Typically $g(\cdot; \valpha_i)$ will have its own parameters in addition to $\valpha_i$. The derivative of the transformer required for the computation of the Jacobian determinant is simply equal to $g(\ervz_i; \valpha_i)$. This approach results in arbitrarily flexible transformers, but the integral lacks analytical tractability. One possibility is to resort to a numerical approximation.  

An analytically tractable integration-based transformer can be obtained by taking the integrand $g(\cdot; \valpha_i)$ to be a positive polynomial of degree $2L$.
The integral will be a polynomial of degree $2L+1$ in $\ervz_i$, and thus can be computed analytically.  Since every positive polynomial of degree $2L$ can be written as a sum of $2$ (or more) squares of polynomials of degree $L$ \citep[Proposition 1.1.2]{marshall2008positive}, this fact can be exploited to define a \textit{sum-of-squares polynomial transformer} \citep{jaini2019sum}:
\begin{equation}
    \tau(\ervz_i; \vh_i) = \integralab{\sum_{k=1}^{K} \left(\sum_{\ell=0}^{L} \alpha_{ik\ell} \,\ervz^{\ell} \right)^{2}}{\ervz}{0}{\ervz_{i}}  + \beta_i,
\end{equation}
where $\vh_i$ comprises $\beta_i$ and all polynomial coefficients $\alpha_{ik\ell}$, and $K \ge 2$. A nice property of the sum-of-squares polynomial transformer is that the coefficients $\alpha_{ik\ell}$ are unconstrained. Moreover, the affine transformer can be derived as the special case of $L=0$: \begin{equation}
    \integralab{\sum_{k=1}^{K} \left(\alpha_{ik0} \,\ervz^{0} \right)^{2}}{\ervz}{0}{\ervz_{i}} + \beta_i= \left(\sum_{k=1}^{K} \alpha_{ik0}^{2} \right) \ervz \Big|_{0}^{\ervz_{i}}  + \beta_i = \alpha_i \ervz_{i}  + \beta_i,
\end{equation}
where $\alpha_i = \sum_{k=1}^{K} \alpha_{ik0}^{2}$. It can be shown that, for large enough $L$, the sum-of-squares polynomial transformer can approximate arbitrarily well any monotonically increasing function \citep[Theorem 3]{jaini2019sum}. Nonetheless, since only polynomials of degree up to $4$ can be solved analytically, the sum-of-squares polynomial transformer is not analytically invertible for $L \ge 2$, and can only be inverted iteratively using e.g.~bisection search \citep{rburden81:numerical}.

\paragraph{Spline-based transformers}

\begin{figure}[t]
    \centering
    \begin{subfigure}[b]{0.42\textwidth}
        \includegraphics[width=\textwidth]{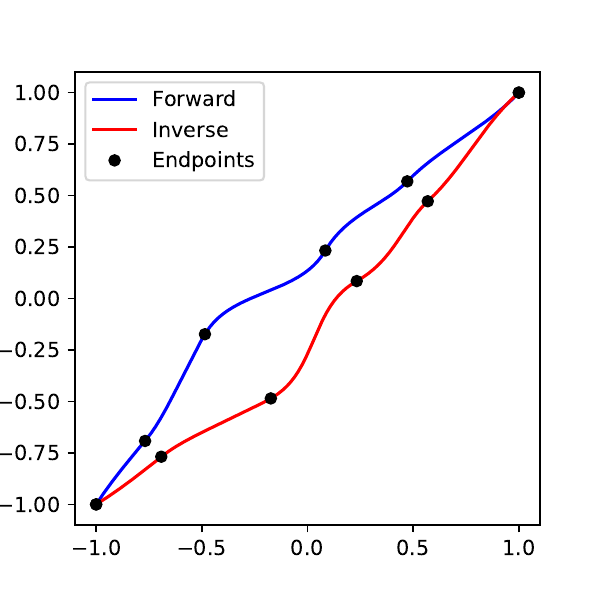}
        \caption{Forward and inverse transformer\label{fig:spline:forward_and_inverse}}
    \end{subfigure}~~
    \begin{subfigure}[b]{0.42\textwidth}
        \includegraphics[width=\textwidth]{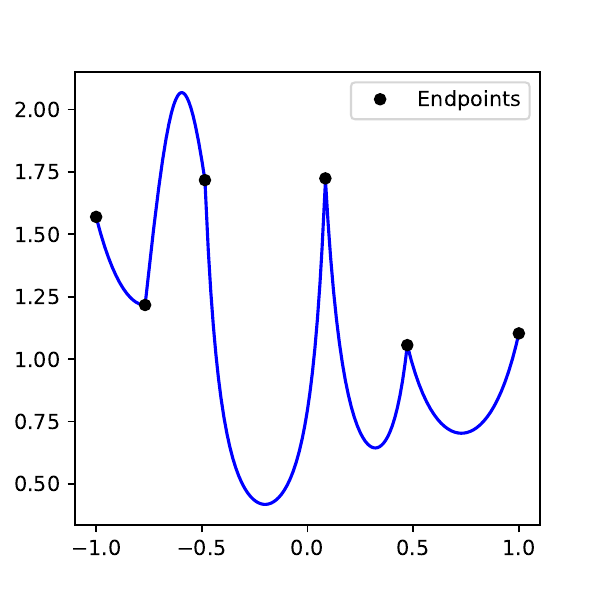}
        \caption{Transformer derivative\label{fig:spline:derivative}}
    \end{subfigure}
    \caption{Example of a spline-based transformer with $5$ segments. Each segment is a monotonic rational-quadratic function, which can be easily inverted \citep{durkan2019neural}. The locations of the endpoints (black dots) parameterize the spline. \label{fig:spline}} 
\end{figure}

So far, we have discussed non-affine transformers that can be made arbitrarily flexible, but don't have an analytic inverse.
One way to overcome this limitation is by implementing the transformer as a monotonic \textit{spline}, i.e.~a piecewise function consisting of $K$ segments, where each segment is a simple function that is easy to invert. Specifically, given a set of $K + 1$ input locations $\ervz_{i0}, \ldots, \ervz_{iK}$, the transformer $\tau(\ervz_i;\vh_i)$ is taken to be a simple monotonic function (e.g.~a low-degree polynomial) in each interval $[\ervz_{i(k-1)}, \ervz_{ik}]$, under the constraint that the $K$ segments meet at the endpoints $\ervz_{i1}, \ldots, \ervz_{i(K-1)}$. Outside the interval $[\ervz_{i0}, \ervz_{iK}]$, the transformer can default to a simple function such as the identity. Typically, the parameters $\vh_i$ of the transformer are the input locations $\ervz_{i0}, \ldots, \ervz_{iK}$, the corresponding output locations $\ervz_{i0}', \ldots, \ervz_{iK}'$,  and (depending on the type of spline) the derivatives (i.e.~slopes) at $\ervz_{i0}, \ldots, \ervz_{iK}$. See \Cref{fig:spline} for an illustration.

Spline-based transformers are distinguished by the type of spline they use, i.e.~by the functional form of the segments. The following options have been explored thus far, in order of increasing flexibility: linear and quadratic splines \citep{muller2019neural}, cubic splines \citep{durkan2019cubic}, linear-rational splines \citep{dolatabadi2020splines}, and rational-quadratic splines \citep{durkan2019neural}. 
Spline-based transformers are as fast to invert as to evaluate, while maintaining exact analytical invertibility. Evaluating or inverting a spline-based transformer is done by first locating the right segment---which can be done in $\bigo{\log K}$ time using binary search---and then evaluating or inverting that segment, which is assumed to be analytically tractable.  By increasing the number of segments $K$, a spline-based transformer can be made arbitrarily flexible.

\subsubsection{Implementing the Conditioner}
\label{sec:ar_flows:conditioner}

The conditioner $c_i(\rvz_{<i})$ can be any function of $\rvz_{<i}$, meaning that each conditioner can, in principle, be implemented as an arbitrary model with input $\rvz_{<i}$ and output $\vh_i$. However, a na\"{i}ve implementation in which each $c_i(\rvz_{<i})$ is a separate model would scale poorly with the dimensionality $D$, requiring $D$ model evaluations, each with a vector of average size $D/2$.  This is in addition to the cost of storing and estimating the parameters of $D$ independent models. In fact, early work on flow precursors \citep{chen2001gaussianization} dismissed the autoregressive approach as prohibitively expensive.

Nonetheless, this problem can be effectively addressed in practice by sharing parameters across the conditioners $c_i(\rvz_{<i})$, or even by combining the conditioners into a single model. In the following paragraphs, we will discuss some practical implementations of the conditioner that allow it to scale to high dimensions.

\paragraph{Recurrent conditioners}

One way to share parameters across conditioners is by implementing them jointly using a \textit{recurrent neural network} (RNN)\@. The $i$-th conditioner is implemented as:
\begin{equation}
    \vh_i = c(\vs_i)
    \quad{\text{where}}\quad
    \begin{array}{l}
          \vs_1 = \text{initial state} \\
          \vs_{i} = \mathrm{RNN}(\ervz_{i-1}, \vs_{i-1}) \text{ for }i>1.
    \end{array}
\end{equation}
The RNN processes $\rvz_{<D} = (\ervz_1, \ldots, \ervz_{D-1})$ one element at a time, and at each step it updates a fixed-size internal state $\vs_i$ that summarizes the subsequence $\rvz_{<i} = (\ervz_1, \ldots, \ervz_{i-1})$. The network $c$ that computes $\vh_i$ from $\vs_i$ can be the same for each step. The initial state $\vs_1$ can be fixed or it can be a learned parameter of the RNN\@. Any RNN architecture can be used, such as LSTM \citep{hochreiter1997long} or GRU \citep{cho2014gru}.

RNNs have been used extensively to share parameters across the conditional distributions of autoregressive models. Examples of RNN-based autoregressive models include distribution estimators \citep{larochelle2011neural, uria2013rnade, uria2014deep}, sequence models \citep{mikolov2010recurrent, graves2013generating, sutskever2014seq2seq}, and image/video models \citep{theis2015generative, van2016pixel, kalchbrenner2016video}. \Cref{sec:ar_models_and_ar_flows}  discusses  the relationship between autoregressive models and autoregressive flows in  detail.

In the autoregressive-flows literature, RNN-based conditioners have been proposed by e.g.\ \citet{oliva2018tan} and \citet{iaf}, but are relatively uncommon compared to alternatives. The main downside of RNN-based conditioners is that they turn an inherently parallel computation into a sequential one: the states $\vs_1, \ldots, \vs_D$ must be computed sequentially, even though each $\vh_i$ can in principle be computed independently and in parallel from $\rvz_{<i}$. Since this recurrent computation involves $\bigo{D}$ sequential steps, it can be slow for high-dimensional data such as images or videos.

\paragraph{Masked conditioners}

Another approach that shares parameters across conditioners but avoids the sequential computation of an RNN is based on \textit{masking}. This approach uses a single, typically feedforward neural network that takes in $\rvz$ and outputs the entire sequence $(\vh_1, \ldots, \vh_D)$ in one pass. The only requirement is that this network must obey the autoregressive structure of the conditioner: an output $\vh_i$ cannot depend on inputs $\rvz_{\ge i}$.

To construct such a network, one takes an arbitrary neural network and removes connections until there is no path from input $\ervz_i$ to outputs $(\vh_1, \ldots, \vh_i)$. A simple way to remove connections is by multiplying each weight matrix elementwise with a binary matrix of the same size.  This has the effect of removing the connections corresponding to weights that are multiplied by zero, while leaving all other connections unmodified. These binary matrices can be thought of as `masking out' connections, hence the term \textit{masking}. The masked network will have the same architecture and size as the original network.  In turn, it retains the computational properties of the original network, such as parallelism or ability to evaluate efficiently on a GPU\@.

A general procedure for constructing masks for multilayer perceptrons with arbitrarily many hidden layers or hidden units was proposed by \citet{germain2015made}. The key idea is to assign a `degree' between $1$ and $D$ to each input, hidden, and output unit, and mask-out the weights between subsequent layers such that no unit feeds into a unit with lower or equal degree. In convolutional networks, masking can be done by multiplying the filter with a binary matrix of the same size, which leads to a type of convolution often referred to as \textit{autoregressive} or \textit{causal} convolution \citep{oord2016conditional, oord2016wavenet, hoogeboom2019emerging}. In architectures that use self-attention, masking can be done by zeroing out the softmax probabilities \citep{vaswani2017attention}.

Masked autoregressive flows have two main advantages. First, they are efficient to evaluate. Given $\rvz$, the parameters $(\vh_1, \ldots, \vh_D)$ can be obtained in one neural-network pass, and then each dimension of $\rvz'$ can be computed in parallel via $\ervz_i' = \tau(\ervz_i; \vh_i)$. Second, masked autoregressive flows are universal approximators. Given a large enough conditioner and a flexible enough transformer, they can represent any autoregressive transformation with monotonic transformers and thus transform between any two distributions (as discussed in \Cref{sec:flows:expressivity}).

On the other hand, the main disadvantage of masked autoregressive flows is that they are not as efficient to invert as to evaluate. This is because the parameters $\vh_i$ that are needed to obtain $\ervz_i = \tau^{-1}(\ervz_i'; \vh_i)$ cannot be computed until all $(\ervz_i, \ldots, \ervz_{i-1})$ have been obtained. Following this logic, we must first compute $\vh_1$ by which we obtain $\ervz_1$, then compute $\vh_2$ by which we obtain $\ervz_2$, and so on until $\ervz_D$ has been obtained. Using a masked conditioner $c$, the above procedure can be implemented in pseudocode as follows:
\begin{equation}\begin{split}
    &\text{Initialize }\rvz\text{ to an arbitrary value} \\
    &\text{for }i=1,\ldots,D\\
    &\quad(\vh_1, \ldots, \vh_D) = c(\rvz)\\
    &\quad \ervz_i = \tau^{-1}(\ervz_i'; \vh_i).
\end{split}\end{equation}
To see why this procedure is correct, observe that if $\rvz_{\le i-1}$ is correct before the $i$-th iteration, then $\vh_i$ will be computed correctly (due to the autoregressive structure of $c$) and thus $\rvz_{\le i}$ will be correct before the $(i+1)$-th iteration. Since $\rvz_{\le 0} = \varnothing$ is correct before the first iteration (in a degenerate sense, but still), by induction it follows that $\rvz_{\le D} = \rvz$ will be correct at the end of the loop. Even though the above procedure can invert the flow exactly (provided the transformer is easy to invert), it requires calling the conditioner $D$ times. This means that inverting a masked autoregressive flow using the above method is about $D$ times more expensive than evaluating the forward transformation. For high-dimensional data such as images or video, this can be prohibitively expensive. 

An alternative way to invert the flow, proposed by \citet{song2019mintnet}, is to solve the equation $\rvz' = f_{\vphi}(\rvz)$ approximately, by iterating the following Newton-style fixed-point update:
\begin{equation}
    \rvz_{k+1} = \rvz_{k} - \alpha\, \mathrm{diag}\br{J_{f_{\vphi}}(\rvz_{k})}^{-1}\br{f_{\vphi}(\rvz_k) - \rvz'},
\end{equation}
where $\alpha$ is a step-size hyperparameter, and $\mathrm{diag}(\cdot)$ returns a diagonal matrix whose diagonal is the same as that of its input. A convenient initialization is $\rvz_0 = \rvz'$. \citet{song2019mintnet} showed that the above procedure is locally convergent for $0<\alpha<2$, and since $f_{\vphi}^{-1}(\rvz')$ is the only fixed point, the procedure must either converge to it or diverge. With a masked autoregressive flow, computing both $f_{\vphi}(\rvz_k)$ and $\mathrm{diag}\br{J_{f_{\vphi}}(\rvz_{k})}$ can be done efficiently by calling the conditioner once. Hence the above Newton-like procedure can be more efficient than inverting the flow exactly when the number of iterations to convergence in practice is significantly less than $D$. On the other hand, the above Newton-like procedure is approximate and guaranteed to converge only locally.

Despite the computational difficulties associated with inversion, masking remains one of the most popular techniques for implementing autoregressive flows.  It is well suited to situations for which inverting the flow is not needed or the data dimensionality is not too large. Examples of flow-based models that use masking include IAF \citep{iaf}, MAF \citep{papamakarios2017masked}, NAF \citep{huang2018neural}, block-NAF \citep{DeCao:2019:bnaf}, MintNet \citep{song2019mintnet} and MaCow \citep{ma2019macow}. Masking can also be used and has been popular in implementing non-flow-based autoregressive models such as MADE \citep{germain2015made}, PixelCNN \citep{oord2016conditional, salimans2017pixelcnn++} and WaveNet \citep{oord2016wavenet, oord2017parallel}.

\paragraph{Coupling layers}

As we have seen, masked autoregressive flows have computational asymmetry that impacts their application and usability. Either sampling or density evaluation will be $D$ times slower than the other. If both of these operations are required to be fast, a different implementation of the conditioner is needed.  One such implementation that is computationally symmetric, i.e.~equally fast to evaluate or invert, is the \textit{coupling layer} \citep{dinh2014nice, dinh2016density}. The idea is to choose an index $d$ (a common choice is $D/2$ rounded to an integer) and design the conditioner such that:
\begin{itemize}
    \item Parameters $(\vh_1, \ldots, \vh_{d})$ are constants, i.e.~not a function of $\rvz$.
    \item Parameters $(\vh_{d+1}, \ldots, \vh_{D})$ are functions of $\rvz_{\le d}$ only, i.e.~they don't depend on $\rvz_{> d}$.
\end{itemize}
This can be easily implemented using an arbitrary function approximator $F$ (such as a neural network) as follows:
\begin{equation}\begin{split}
    (\vh_1, \ldots, \vh_{d}) &= \text{constants, either fixed or estimated}\\
    (\vh_{d+1}, \ldots, \vh_{D}) &= F(\rvz_{\le d}).
\end{split}\end{equation}
In other words, the coupling layer splits $\rvz$ into two parts such that $\rvz = [ \rvz_{\le d}, \rvz_{> d} ]$. The first part is transformed elementwise independently of other dimensions. The second part is transformed elementwise in a way that depends on the first part.  We can also think of the coupling layer as implementing an aggressive masking strategy that allows only $(\vh_{d+1}, \ldots, \vh_{D})$ to depend only on $\rvz_{\le d}$. 

\begin{figure}[t]
    \centering
    \begin{subfigure}[b]{0.49\textwidth}
        \includegraphics[width=\textwidth]{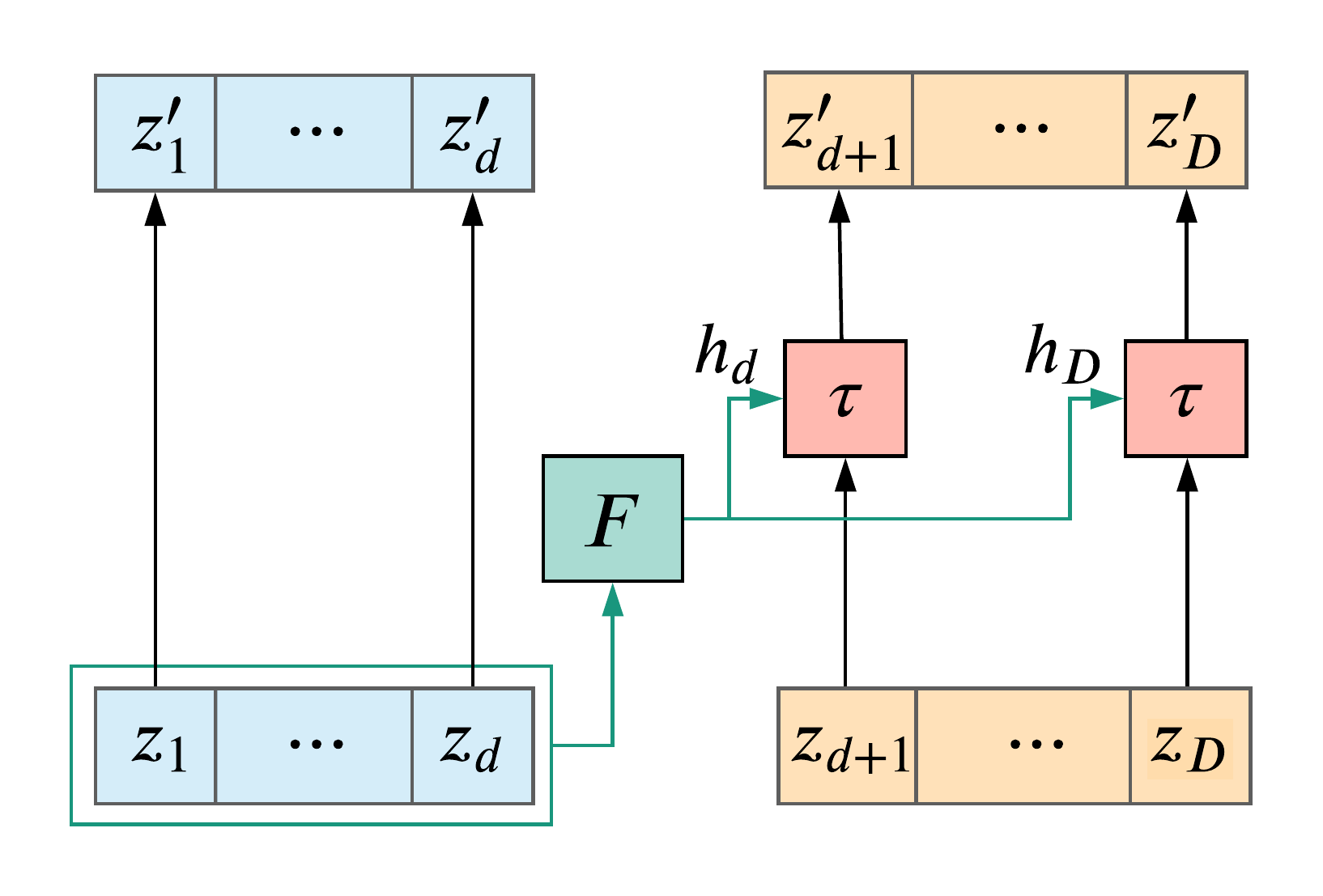}
        \caption{Forward\label{fig:coupling_layer:forward}}
    \end{subfigure}~~
    \begin{subfigure}[b]{0.49\textwidth}
        \includegraphics[width=\textwidth]{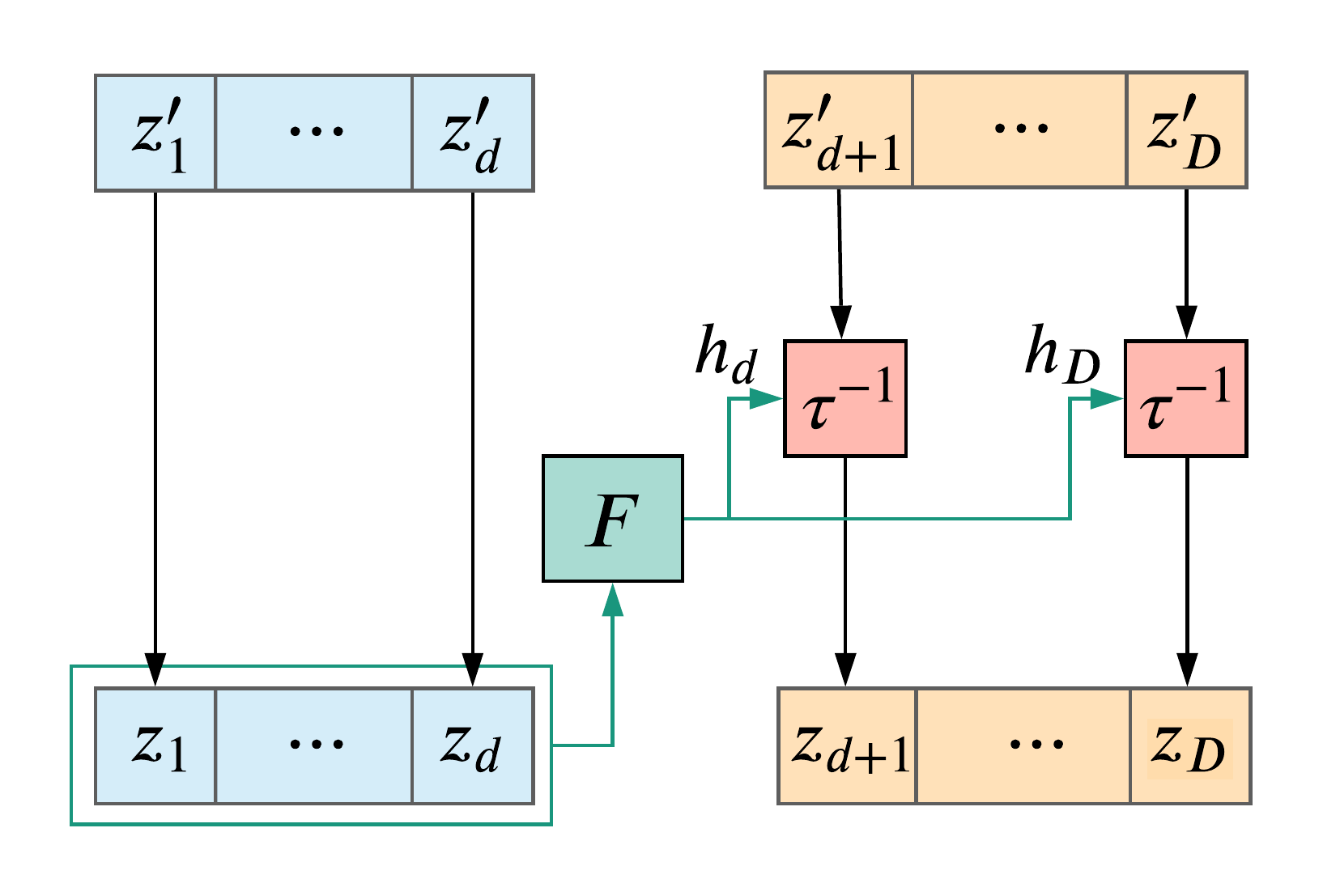}
        \caption{Inverse\label{fig:coupling_layer:inverse}}
    \end{subfigure}
    \caption{Illustration of a coupling layer. \label{fig:coupling_layer}} 
\end{figure}

Common implementations of coupling layers fix the transformers $\tau(\cdot; \vh_1), \ldots, \tau(\cdot; \vh_D)$ to the identity function. In this case, the transformation can be written as follows:
\begin{equation}\begin{split}
\rvz'_{\le d} &= \rvz_{\le d}\\
(\vh_{d+1}, \ldots, \vh_{D}) &= F(\rvz_{\le d})\\
\rvz'_{i} &= \tau(\rvz_{i}; \vh_i) \text{ for }i>d.
\end{split}\end{equation}
In turn, the inverse transformation is straightforward,  given by:
\begin{equation}\begin{split}
\rvz_{\le d} &= \rvz'_{\le d}\\
(\vh_{d+1}, \ldots, \vh_{D}) &= F(\rvz_{\le d})\\
\rvz_{i} &= \tau^{-1}(\rvz'_{i}; \vh_i) \text{ for }i>d.
\end{split}\end{equation}
These are illustrated in \Cref{fig:coupling_layer}.
Like all autoregressive flows, the Jacobian of the transformation is lower triangular, but in addition it has the following special structure:
\begin{equation}
 J_{f_{\vphi}} = \begin{bmatrix}
    \rmI & \mathbf{0}  \\
    \rmA & \rmD \\
\end{bmatrix},
\end{equation}
where $\rmI$ is the $d\times d$ identity matrix, $\mathbf{0}$ is the $d\times (D-d)$ zero matrix, $\rmA$ is a $(D-d)\times d$ full matrix, and $\rmD$ is a $(D-d)\times (D-d)$ diagonal matrix. The Jacobian determinant is simply the product of the diagonal elements of $\rmD$, which are equal to the derivatives of the transformers $\tau(\cdot; \vh_{d+1}), \ldots, \tau(\cdot; \vh_{D})$.

Coupling layers and fully autoregressive flows are two extremes on a spectrum of possible implementations. A coupling layer splits $\rvz$ into two parts and transforms the second part elementwise as a function of the first part, whereas a fully autoregressive flow splits the input into $D$ parts (each with one element in it) and transforms each part as a function of all previous parts. Clearly, there are intermediate choices: one can split the input into $K$ parts and transform the $k$-th part elementwise as a function of parts $1$ to $k-1$, with $K=2$ corresponding to a coupling layer and $K=D$ to a fully autoregressive flow. Using masking, inverting the transformation will be $\bigo{K}$ times more expensive than evaluating it, hence $K$ could be chosen based on the computational requirements of the task. 

The efficiency of coupling layers comes at the cost of reduced expressive power. Unlike a recurrent or masked autoregressive flow, a single coupling layer can no longer represent any autoregressive transformation, regardless of how expressive the function $F$ is.  As a result, an autoregressive flow with a single coupling layer is no longer a universal approximator. Nonetheless, the expressivity of the flow can be increased by composing multiple coupling layers. When composing coupling layers, the elements of $\rvz$ need to be permuted between layers so that all dimensions have a chance to be transformed as well as interact with one another. Previous work across various domains \citep[see e.g.][]{kingma2018glow, prenger2018waveglow, durkan2019neural} has shown that composing coupling layers can indeed create flexible flows.

Theoretically, it is easy to show that a composition of $D$ coupling layers is indeed a universal approximator, as long as the index $d$ of the $i$-th coupling layer is equal to $i-1$. Observe that the $i$-th coupling layer can express any transformation of the form $\ervz'_i = \tau(\ervz_i; c_i(\rvz_{<i}))$, hence a composition of $D$ such layers will have transformed each dimension fully autoregressively. However, this construction involves $D$ sequential computations (one for each layer) in both the forward and inverse directions, so it doesn't provide an improvement over recurrent or masked autoregressive flows. It is an open problem whether it's possible to obtain a universal approximator by composing strictly fewer than $\bigo{D}$ coupling layers.

Coupling layers are one of the most popular methods for implementing flow-based models because they allow both density evaluation and sampling to be fast. A flow based on coupling layers can be tractably fitted with maximum likelihood and then be sampled from efficiently.  Thus coupling layers are often found in generative models of high-dimensional data such as images, audio and video. Examples of flow-based models with coupling layers include NICE \citep{dinh2014nice}, Real NVP \citep{dinh2016density}, Glow \citep{kingma2018glow}, WaveGlow \citep{prenger2018waveglow}, FloWaveNet \citep{kim2018flowavenet} and Flow++ \citep{ho2019flow++}.

\subsubsection{Relationship with Autoregressive Models}
\label{sec:ar_models_and_ar_flows}

Alongside normalizing flows, another popular class of models for high-dimensional distributions is the class of \textit{autoregressive models}. Autoregressive models have a long history, from the general framework of Bayesian networks \citep{pearl1988probabilistic, frey1998graphical} to more recent neural-network-based implementations \citep{bengio2000modeling, uria2016nadejmlr}.

To construct an autoregressive model of $\px(\rvx)$, we first decompose $\px(\rvx)$ into a product of $1$-dimensional conditionals using the chain rule of probability:
\begin{equation}\label{eq:chanRuleProb}
    \px(\rvx) = \prod_{i=1}^{D} \px(\ervx_{i} \g \rvx_{<i}).
\end{equation}
We then model each conditional by some parametric distribution with parameters $\vh_i$:
\begin{equation}
    \px(\ervx_{i} \g \rvx_{<i}) = \px(\ervx_{i}; \vh_i),
    \quad\text{where}\quad
    \vh_i = c_i(\rvx_{<i}).
\end{equation}
For example, $\px(\ervx_{i}; \vh_i)$ can be a Gaussian parameterized by its mean and variance, or a mixture of Gaussians parameterized by the mean, variance and mixture coefficient of each component. The functions $c_i$ are analogous to the conditioners of an autoregressive flow, and are often implemented with neural networks using either RNNs or masking as discussed in previous sections. Apart from continuous data, autoregressive models can be readily used for discrete or even mixed data. If $\ervx_i$ is discrete for some $i$, then $\px(\ervx_{i}; \vh_i)$ can be a parametric probability mass function such as a categorical or a mixture of Poissons.

We now show that \emph{all} autoregressive models of continuous variables are in fact autoregressive flows with a single autoregressive layer. Let $\tau(\ervx_i; \vh_i)$ be the cumulative distribution function of $\px(\ervx_{i}; \vh_i)$, defined as follows:
\begin{equation}
    \tau(\ervx_i; \vh_i) = \integralab{\px(\ervx'_{i}; \vh_i)}{\ervx'_{i}}{-\infty}{\ervx_i}.
\end{equation}
The function $\tau(\ervx_i; \vh_i)$ is differentiable if $\px(\ervx_{i}; \vh_i)$ is continuous, and strictly increasing if $\px(\ervx_{i}; \vh_i) > 0$, both of which are the case in standard implementations of autoregressive models. As shown in \Cref{sec:flows:expressivity}, the vector $\rvu = (\ervu_1, \ldots, \ervu_D)$, obtained by
\begin{equation}
    \ervu_i = \tau(\ervx_i; \vh_i)\quad\text{where}\quad
    \vh_i = c_i(\rvx_{<i}),
\end{equation}
is always distributed uniformly in $(0, 1)^D$. The above expression has exactly the same form as the definition of an autoregressive flow in \Cref{eq:ar_flow_forward}, with $\rvz=\rvx$ and $\rvz'=\rvu$. Therefore, an autoregressive model is in fact an autoregressive flow with a single autoregressive layer.  Moreover, the layer's transformers are the cumulative distribution functions of the conditionals of the autoregressive model, and the layer's base distribution is a uniform in $(0, 1)^D$. We can make the connection explicit by writing the density under the change of variables: \begin{equation}
    \log \px(\rvx) = \log \prod_{i=1}^{D} \text{Uniform}(\tau(\ervx_i; \vh_i); 0, 1) + \log \prod_{i=1}^{D} \px(\ervx_i; \vh_i) = \sum_{i=1}^{D} \log  \px(\ervx_{i} \g \rvx_{<i}).
\end{equation} The term involving the uniform base density drops from the expression, leaving just the Jacobian determinant.\footnote{\citet{inouye18a} termed flows of this form---whereby the density is fully determined by the Jacobian determinant---\textit{density destructors}.} Following \Cref{eq:ar_flow_inverse}, the inverse autoregressive flow that maps $\rvu$ to $\rvx$ is obtained by iterating the following for $i\in\set{1, \ldots, D}$:
\begin{equation}
    \ervz_i = \tau^{-1}(\ervu_i; \vh_i)\quad\text{where}\quad
    \vh_i = c_i(\rvx_{<i}).
\end{equation}
The above corresponds exactly to sampling from the autoregressive model one element at a time, where at each step the corresponding conditional is sampled from using \textit{inverse transform sampling}.

Yet the transformer is not necessarily limited to being the inverse CDF\@.  We can make further connections between specific types of autoregressive models and the transformers discussed in \Cref{sec:ar_flows:transformer}. For example, consider an autoregressive model with Gaussian conditionals of the form:
\begin{equation}
    \px(\ervx_{i}; \vh_i) = \mathcal{N}\br{\ervx_{i} ; \mu_{i}, \sigma_{i}^2}
    \quad\text{where}\quad
    \vh_i = \set{\mu_i, \sigma_i}.
\end{equation}
The above conditional can be reparameterized as follows:
\begin{equation}
    \ervx_i = \sigma_i\ervu_i + \mu_i
    \quad\text{where}\quad
    \ervu_i \sim \mathcal{N}(0, 1).
\end{equation}
Hence, the entire autoregressive model can be reparameterized as an \textit{affine autoregressive flow} as shown in \Cref{eq:affine_ar_flow}, where $\alpha_i = \sigma_i$, $\beta_i = \mu_i$, and the base distribution is a standard Gaussian \citep{iaf, papamakarios2017masked}. In the same way, we can relate other types of autoregressive models to non-affine transformers. For example, an autoregressive model whose conditionals are mixtures of Gaussians can be reparameterized as an autoregressive flow with combination-based transformers such as those in \Cref{eq:non_affine_neural_transformer}. Similarly, an autoregressive model whose conditionals are histograms can be reparameterized as an autoregressive flow with transformers given by linear splines.

In consequence, we can think of autoregressive flows as subsuming and further extending autoregressive models for continuous variables. There are several benefits of viewing autoregressive models as flows. First, this view decouples the model architecture from the source of randomness, which gives us freedom in specifying the base distribution. Thus, we can enhance the flexibility of an autoregressive model by choosing a more flexible base distribution; for example, the base distribution can be another autoregressive model with its own learnable parameters. This provides a framework for composing autoregressive models, like layers in a flow \citep{papamakarios2017masked}. Also, it allows us to compose autoregressive models with other types of flows, potentially non-autoregressive ones.

\subsection{Linear Flows}

As discussed in the previous section, autoregressive flows restrict an output variable $\ervz'_i$ to depend only on inputs $\rvz_{\le i}$, making the flow dependent on the order of the input variables. 
As we showed, in the limit of infinite capacity, this restriction doesn't limit the flexibility of the flow-based model. 
However, in practice we don't operate at infinite capacity. The order of the input variables \emph{will} determine the set of distributions the model can represent. Moreover, the target transformation may be easy to learn for some input orderings and hard to learn for others. The problem is further exacerbated when using coupling layers since only part of the input variables is transformed. 

To cope with these limitations in practice, it often helps to permute the input variables between successive autoregressive layers. For coupling layers it is in fact necessary: if we don't permute the input variables between successive layers, part of the input will never be transformed. A permutation of the input variables is itself an easily invertible transformation, and its absolute Jacobian determinant is always $1$ (i.e.~it is volume-preserving).  Hence, permutations can seamlessly be composed with other invertible and differentiable transformations the usual way.

An approach that generalizes the idea of a permutation of input variables is that of a \textit{linear flow}. A linear flow is essentially an invertible linear transformation of the form:
\begin{equation}
    \rvz' = \rmW \rvz,
\end{equation}
where $\rmW$ is a $D\times D$ invertible matrix that parameterizes the transformation. The Jacobian of the above transformation is simply $\rmW$, making the Jacobian determinant equal to $\det\rmW$. A permutation is a special case of a linear flow, where $\rmW$ is a permutation matrix (i.e.~a binary matrix with exactly one entry of $1$ in each row and column and $0$s everywhere else). Alternating invertible linear transformations with autoregressive/coupling layers is often used in practice \citep[see e.g.][]{kingma2018glow, durkan2019neural}.

A straightforward implementation of a linear flow is to directly parameterize and learn the matrix $\rmW$. However, this approach can be problematic. First, $\rmW$ is not guaranteed to be invertible. Second, inverting the flow, which amounts to solving the linear system $\rmW \rvz = \rvz'$ for $\rvz$, takes time $\bigo{D^3}$ in general, which is prohibitively expensive for high-dimensional data. Third, computing $\det\rmW$ also takes time $\bigo{D^3}$ in general.

To address the above challenges, many approaches restrict $\rmW$ to a structured matrix, or a product of structured matrices. For example, if we restrict $\rmW$ to be triangular, we can guarantee its invertibility by e.g.\ making the diagonal elements positive. Moreover, inversion then costs $\bigo{D^2}$ (i.e.~about the same as a matrix multiplication) and computing the determinant costs $\bigo{D}$. In Appendix \ref{app:linFlow}, we discuss in more detail this and a few more parameterizations that restrict the form of $\rmW$ in various ways.

In any case, it is important to note that it is \emph{impossible} to parameterize all invertible matrices of size $D\times D$ in a continuous way,
so any continuous parameterization of $\rmW$ that guarantees its invertibility
will unavoidably leave out some invertible matrices.
That's because there is no continuous surjective function from $\R^{D^2}$ to the set of $D\times D$ invertible matrices. To see why, consider two invertible matrices $\rmW_A$ and $\rmW_B$ such that $\det\rmW_A > 0$ and $\det\rmW_B < 0$. If there exists a continuous parameterization of all invertible matrices, then there exists a continuous path that connects $\rmW_A$ and $\rmW_B$. However, since the determinant is a continuous function of the matrix entries, any such path must include a matrix with zero determinant, i.e.~a non-invertible matrix, which is a contradiction. This argument shows that the set of $D\times D$ invertible matrices contains two disconnected `islands'---one containing matrices with positive determinant, the other with negative determinant---that are fully separated by the set of non-invertible matrices. In practice, this means that we can only hope to continuously parameterize one of these two islands, fixing the sign of the determinant from the outset.

\subsection{Residual Flows}\label{sec:residual_flows}

\begin{wrapfigure}{r}{0.3\textwidth}
  \centering
  \includegraphics[width=0.2\textwidth]{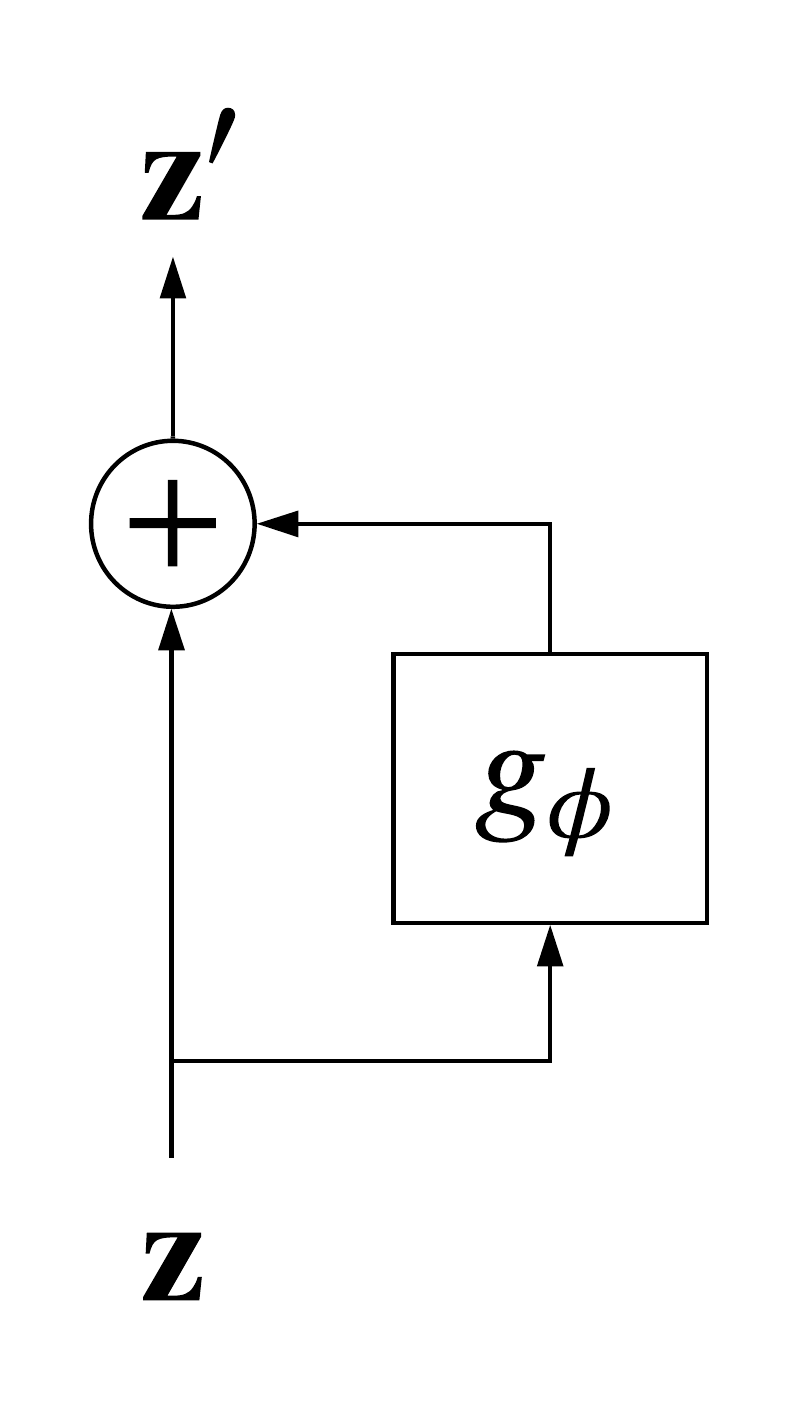}
  \caption{Residual flow \label{fig:residual_flow}}
\end{wrapfigure}

In this section, we consider a class of invertible transformations of the general form:
\begin{equation}\label{eq:residual_flow}
    \rvz' = \rvz + g_{\vphi}(\rvz),
\end{equation}
where $g_{\vphi}$ is a function that outputs a $D$-dimensional translation vector, parameterized by $\vphi$ (\Cref{fig:residual_flow}).
This structure bears a strong similarity to \textit{residual networks} \citep{he2016deep}, and thus we use the term \textit{residual flow} to refer to a normalizing flow composed of such transformations. Residual transformations are not always invertible, but can be made invertible if $g_{\vphi}$ is constrained appropriately. In what follows, we discuss two general approaches to designing invertible residual transformations: the first is based on \textit{contractive maps}, and the second is based on the \textit{matrix determinant lemma}.

\subsubsection{Contractive Residual Flows}\label{sec:contractive_residual_flows}

A residual transformation is guaranteed to be invertible if $g_{\vphi}$ can be made \textit{contractive} with respect to some distance function \citep{behrmann2019invertible, chen2019residual}.
In general, a map $F:\R^D \rightarrow\R^D$ is said to be contractive with respect to a distance function $\delta$ if there exists a constant $L<1$ such that for any two inputs $\rvz_A$ and $\rvz_B$ we have:
\begin{equation}
    \delta \br{F(\rvz_A), F(\rvz_B)} \le L \,\delta(\rvz_A, \rvz_B).
\end{equation}
In other words, a contractive map brings any two inputs closer together (as measured by $\delta$) by at least a factor $L$. It directly follows that $F$ is Lipschitz continuous with a Lipschitz constant equal to $L$. The \textit{Banach fixed-point theorem} \citep[Theorem 9.23]{rudin1976principles} states that any such contractive map has exactly one fixed point $\rvz_* = F(\rvz_*)$. Furthermore, this fixed point is the limit of any sequence $(\rvz_0, \rvz_1, \ldots)$ that is formed by an arbitrary starting point $\rvz_0$ and repeated application of $F$, i.e.~$\rvz_{k+1} = F(\rvz_k)$ for all $k\ge 0$.

Invertibility of the residual transformation $\rvz' = f_{\vphi}(\rvz) = \rvz + g_\vphi(\rvz)$ follows directly from $g_\vphi$ being contractive. Given $\rvz'$, consider the map:
\begin{equation}
    F(\hat{\rvz}) = \rvz' - g_\vphi(\hat{\rvz}).
\end{equation}
If $g_\vphi$ is contractive with Lipschitz constant $L$, then $F$ is also contractive with the same Lipschitz constant. Hence, from the Banach fixed-point theorem, there exists a unique $\rvz_*$ such that $\rvz_* = \rvz' - g_\vphi(\rvz_*)$. By rearranging, we see that $\rvz' = f_\vphi(\rvz_*)$, and since $\rvz_*$ is unique, it follows that $f_\vphi$ is invertible.

In addition to a proof of the invertibility of $f_\vphi$, the above argument also gives us an algorithm for inversion. Starting from an arbitrary input $\rvz_0$ (a convenient choice is $\rvz_0 = \rvz'$), we can iteratively apply $F$ as follows:
\begin{equation}
    \rvz_{k+1} = \rvz' - g_\vphi(\rvz_k)
    \quad\text{for }k\ge 0.
\end{equation}
The Banach fixed-point theorem guarantees that the above procedure converges to $\rvz_* = f_\vphi^{-1}(\rvz')$ for any choice of starting point $\rvz_0$. Moreover, it can be shown that the rate of convergence (with respect to $\delta$) is exponential in the number of iterations $k$, and can be quantified as follows:
\begin{equation}
    \delta(\rvz_k, \rvz_*) \le \frac{L^k}{1 - L}\,\delta(\rvz_0, \rvz_1).
\end{equation} The smaller the Lipschitz constant is, the faster $\rvz_k$ converges to $\rvz_*$. We can think of $L$ as trading off flexibility for efficiency: as $L$ gets smaller, the fewer iterations it takes to approximately invert the flow, but the residual transformation becomes more constrained, i.e.~less flexible. In the extreme case of $L=0$, the inversion procedure converges after one iteration, but the transformation reduces to adding a constant.

A challenge in building contractive residual flows is designing the function $g_\vphi$ to be contractive without impinging upon its flexibility. It is easy to see that the composition of $K$ Lipschitz-continuous functions $F_1, \ldots, F_K$ is also Lipschitz continuous with a Lipschitz constant equal to $\prod_{k=1}^K L_k$, where $L_k$ is the Lipschitz constant of $F_k$. Hence, if $g_\vphi$ is a composition of neural-network layers (as is common in deep learning), it is sufficient to make each layer Lipschitz continuous with $L_k \le 1$, with at least one layer having $L_k < 1$, for the entire network to be contractive. Many elementwise nonlinearities used in deep learning---including the logistic sigmoid, hyperbolic tangent (tanh), and rectified linear (ReLU)---are in fact already Lipschitz continuous with a constant no greater than $1$. Furthermore, linear layers (including dense layers and convolutional layers) can be made contractive with respect to a norm by dividing them with a constant strictly greater than their induced operator norm. One such implementation was proposed by \citet{behrmann2019invertible}: spectral normalization \citep{miyato2018spectral} was used to make linear layers contractive with respect to the Euclidean norm.

One drawback of contractive residual flows is that there is no known general, efficient procedure for computing their Jacobian determinant. Rather, one would have to revert to automatic differentiation to obtain the Jacobian and an explicit determinant computation to obtain the Jacobian determinant, which costs $\bigo{D^3}$ as discussed earlier. Without an efficient way to compute the Jacobian determinant, exactly evaluating the density of the flow model is costly and potentially infeasible for high-dimensional data such as images.

Nonetheless, it is possible to obtain an unbiased estimate of the log absolute Jacobian determinant, and hence of the log density, which 
is enough to train the flow model e.g.~with maximum likelihood using stochastic gradients. We begin by writing the log absolute Jacobian determinant as a power series:\footnote{This power series is essentially the Maclaurin series $\log(1+x) = x - \frac{x^2}{2} + \frac{x^3}{3} - \ldots$ extended to matrices.}
\begin{equation}\label{eq:residual_flow_logdet_taylor}
    \log \abs{\det J_{f_\vphi}(\rvz)} = \log \abs{\det\br{\rmI + J_{g_\vphi}(\rvz)}} = \sum_{k=1}^\infty \frac{(-1)^{k+1}}{k}\trace{J^k_{g_\vphi}(\rvz)},
\end{equation}
where $J^k_{g_\vphi}(\rvz)$ is the $k$-th power of the Jacobian of $g_\vphi$ evaluated at $\rvz$. The above series converges if $\norm{J_{g_\vphi}(\rvz)}<1$ for some submultiplicative matrix norm $\norm{\cdot}$, which in our case holds due to $g_\vphi$ being contractive.  The trace of $J^k_{g_\vphi}(\rvz)$ can be efficiently estimated using the \textit{Hutchinson trace estimator} \citep{Hutchinson:1990:estimator}:
\begin{equation}
    \trace{J^k_{g_\vphi}(\rvz)} \approx \rvv^\top J^k_{g_\vphi}(\rvz) \,\rvv,
\end{equation}
where $\rvv$ can be any $D$-dimensional random vector with zero mean and unit covariance. The Jacobian-vector product $\rvv^\top J^k_{g_\vphi}(\rvz)$ can then be computed with $k$ backpropagation passes. Finally, the infinite sum can be estimated by a finite sum of appropriately re-weighted terms using the \textit{Russian-roulette estimator} \citep{chen2019residual}.

Unlike autoregressive flows, which are based on constraining the Jacobian to be sparse, contractive residual flows have a dense Jacobian in general, which allows all input variables to affect all output variables. As a result, contractive residual flows can be very flexible and have demonstrated good results in practice. On the other hand, unlike the one-pass density evaluation and sampling offered by flows based on coupling layers, exact density evaluation is computationally expensive and sampling is done iteratively, which limits the applicability of contractive residual flows in certain tasks.

\subsubsection{Residual Flows Based on the Matrix Determinant Lemma}

Suppose $\rmA$ is an invertible matrix of size $D\times D$ and $\rmV$, $\rmW$ are matrices of size $D\times M$. The \textit{matrix determinant lemma} states:
\begin{equation}
    \det\br{\rmA + \rmV\rmW^\top} = \det\br{\rmI + \rmW^\top\rmA^{-1}\rmV}\det\rmA.
\end{equation}
If the determinant and inverse of $\rmA$ are tractable and $M$ is less than $D$, the matrix determinant lemma can provide a computationally efficient way to compute the determinant of $\rmA + \rmV\rmW^\top$. For example, if $\rmA$ is diagonal, computing the left-hand side costs $\bigo{D^3 + D^2M}$, whereas computing the right-hand side costs $\bigo{M^3 + DM^2}$, which is preferable if $M<D$. In the context of flows, the matrix determinant lemma can be used to efficiently compute the Jacobian determinant. In this section, we will discuss examples of residual flows that are specifically designed such that application of the matrix determinant lemma leads to efficient Jacobian-determinant computation.

\paragraph{Planar flow} One early example is the \textit{planar flow} \citep{rezende2015variational}, where the function $g_\vphi$ is a one-layer neural network with a single hidden unit:
\begin{equation}
    \rvz' = \rvz + \rvv \sigma(\rvw^{\top} \rvz + b).
\end{equation}
The parameters of the planar flow are $\rvv \in \mathbb{R}^{D}$, $\rvw \in \mathbb{R}^{D}$ and $b \in \mathbb{R}$, and $\sigma$ is a differentiable activation function such as the hyperbolic tangent. This flow can be interpreted as expanding/contracting the space in the direction perpendicular to the hyperplane $\rvw^\top\rvz + b = 0$. The Jacobian of the transformation is given by:
\begin{equation}
    J_{f_\vphi}(\rvz) = \rmI + \sigma'(\rvw^{\top} \rvz + b)\,\rvv\rvw^\top,
\end{equation}
where $\sigma'$ is the derivative of the activation function. The Jacobian has the form of a diagonal matrix plus a rank-$1$ update. Using the matrix determinant lemma, the Jacobian determinant can be computed in time $\bigo{D}$ as follows:
\begin{equation}
    \det J_{f_\vphi}(\rvz) = 1 + \sigma'(\rvw^{\top} \rvz + b)\,\rvw^\top\rvv.
\end{equation}
In general, the planar flow is not invertible for all values of $\rvv$ and $\rvw$. However, assuming that $\sigma'$ is positive everywhere and bounded from above (which is the case if $\sigma$ is the hyperbolic tangent, for example), a sufficient condition for invertibility is $\rvw^\top\rvv > -\frac{1}{\sup_x \sigma'(x)}$.

\paragraph{Sylvester flow} Planar flows can be extended to $M$ hidden units, in which case they are known as \textit{Sylvester flows} \citep{van2018sylvester} and can be written as:
\begin{equation}
    \rvz' = \rvz + \rmV \sigma(\rmW^\top \rvz + \rvb).
\end{equation}
The parameters of the flow are now $\rmV \in \mathbb{R}^{D \times M}$, $\rmW \in \mathbb{R}^{D \times M}$ and $\rvb \in \mathbb{R}^{M}$, and the activation function $\sigma$ is understood elementwise. The Jacobian can be written as:
\begin{equation}
    J_{f_\vphi}(\rvz) = \rmI + \rmV\rmS(\rvz)\rmW^\top,
\end{equation}
where $\rmS(\rvz)$ is an $M\times M$ diagonal matrix whose diagonal is equal to $\sigma'(\rmW^\top \rvz + \rvb)$. Applying the matrix determinant lemma we get:
\begin{equation}
    \det J_{f_\vphi}(\rvz) = \det\br{\rmI + \rmS(\rvz)\rmW^\top\rmV},
\end{equation}
which can be computed in $\bigo{M^3 + DM^2}$. To further reduce the computational cost, \citet{van2018sylvester} proposed the parameterization $\rmV=\rmQ\rmU$ and $\rmW=\rmQ\rmL$, where $\rmQ$ is a $D\times M$ matrix whose columns are an orthonormal set of vectors (this requires $M\le D$), $\rmU$ is $M\times M$ upper triangular, and $\rmL$ is $M\times M$ lower triangular. Since $\rmQ^\top\rmQ=\rmI$ and the product of upper-triangular matrices is also upper triangular, the Jacobian determinant becomes:
\begin{equation}
     \det J_{f_\vphi}(\rvz) = \det\br{\rmI + \rmS(\rvz)\rmL^\top\rmU} = \prod_{i=1}^D\br{1 + S_{ii}(\rvz)L_{ii}U_{ii}}.
\end{equation}
Similar to planar flows, Sylvester flows are not invertible for all values of their parameters. Assuming $\sigma'$ is positive everywhere and bounded from above, a sufficient condition for invertibility is $L_{ii}U_{ii} > -\frac{1}{\sup_x \sigma'(x)}$ for all $i\in\set{1, \ldots, D}$.

\paragraph{Radial flow}
\textit{Radial flows} \citep{tabak2013family, rezende2015variational} take the following form:
\begin{equation}
    \rvz' = \rvz + \frac{\beta}{\alpha + r(\rvz)}(\rvz-\rvz_0)
    \quad\text{where}\quad
    r(\rvz) = \norm{\rvz - \rvz_0}.
\end{equation}
The parameters of the flow are $\alpha\in(0, +\infty)$, $\beta\in\R$ and $\rvz_0\in\R^D$, and $\norm{\cdot}$ is the Euclidean norm. The above transformation can be thought of as a contraction/expansion radially with center $\rvz_0$. The Jacobian can be written as follows:
\begin{equation}
    J_{f_\vphi}(\rvz) = \br{1 + \frac{\beta}{\alpha + r(\rvz)}}\rmI - \frac{\beta}{r(\rvz)\br{\alpha + r(\rvz)}^2}(\rvz - \rvz_0)(\rvz-\rvz_0)^\top,
\end{equation}
which is a diagonal matrix plus a rank-$1$ update. Applying the matrix determinant lemma and rearranging, we get the following expression for the Jacobian determinant, which can be computed in $\bigo{D}$:
\begin{equation}
    \det J_{f_\vphi}(\rvz) = \br{1 + \frac{\alpha\beta}{\br{\alpha + r(\rvz)}^2}}\!\left(1 + \frac{\beta}{\alpha + r(\rvz)}\right)^{D-1}.
\end{equation}
The radial flow is not invertible for all values of $\beta$. A sufficient condition for invertibility is $\beta > -\alpha$.

In summary, planar, Sylvester and radial flows have Jacobian determinants that cost $\bigo{D}$ to compute, and can be made invertible by suitably restricting their parameters. However, there is no analytical way of computing their inverse, which is why these flows have mostly been used to approximate posteriors for variational autoencoders. Moreover, each individual transformation is fairly simple, and it's not clear how the flexibility of the flow can be increased other than by increasing the number of transformations.

\subsection{Practical Considerations when Combining Transformations}
Implementing a flow often amounts to composing as many transformations as computation and memory will allow. For instance, \citet{kingma2018glow}'s Glow architecture employs as many as $320$ sub-transformations distributed across $40$ GPUs to achieve state-of-the-art image generation. Working with such deep flows introduces additional challenges of a practical nature. In this section, we summarize two techniques that, respectively, stabilize the optimization and ease the computational demands of deep flows.  

\paragraph{Normalization} Like with deep neural networks trained with gradient-based methods, normalizing the intermediate representations $\rvz_{k}$ is crucial for maintaining stable gradients throughout the flow.  \textit{Batch normalization} or \textit{batch norm} \citep{ioffe2015batch} has been widely demonstrated to be effective in stabilizing and improving neural-network training, thus making it attractive for use in deep flows as well.  Viewing the batch statistics as fixed, batch norm is essentially a composition of two affine transformations.  The first has scale and translation parameters set by the batch statistics, and the second has free parameters $\valpha$ (scale) and $\vbeta$ (translation):
\begin{equation}    
 \mathrm{BN}(\rvz) = \valpha \odot \frac{\rvz - \hat{\vmu}}{\sqrt{\hat{\vsigma}^{2} + \vepsilon}} + \vbeta,  \ \  \ \ \mathrm{BN}^{-1}(\rvz') = \hat{\vmu} + \frac{\rvz' - \vbeta}{\valpha} \odot \sqrt{\hat{\vsigma}^{2} + \vepsilon}.
 \end{equation}  Moreover, batch norm has an easy-to-compute Jacobian determinant due to it acting elementwise (and thus having a diagonal Jacobian):
\begin{equation}
\det J_{\mathrm{BN}}(\rvz) =  \prod_{i=1}^{D} \frac{\alpha_{i}}{\sqrt{\hat{\sigma}_{i}^{2} + \epsilon_{i}}}.
\end{equation} 
In consequence, batch norm can be inserted between consecutive sub-transformations and treated simply as another member of the composition: $ T_{k} \circ  \mathrm{BN}  \circ T_{k-1} $.

The above formulas assume that batch statistics are fixed, which is true for a trained model. During training, however, batch statistics are not fixed, but are functions of all examples in the batch. This makes batch norm not invertible, unless the batch statistics have been cached during a forward pass. Also, the Jacobian determinant as written above makes little sense mathematically, since batch norm is now a function of the whole batch. Yet, using this Jacobian-determinant formula as an approximation often suffices for training, at least if the batch is large enough \citep{dinh2016density, papamakarios2017masked}.

Glow employs a variant termed \textit{activation normalization} or \textit{act norm} \citep{kingma2018glow} that doesn't use batch statistics $\hat{\vmu}$ and $\hat{\vsigma}$. Instead, before training begins, a batch is passed through the flow, and $\valpha$ and $\vbeta$ are set such that the transformed batch has zero mean and unit variance.  After this data-dependent initialization, $\valpha$ and $\vbeta$ are optimized as model parameters.  Act norm is preferable when training with small mini-batches since batch norm's statistics become noisy and can destabilize training.      

\paragraph{Multi-scale architectures} As mentioned in \Cref{sec:flows:definition}, $\rvx$ and $\rvu$ must have the same dimensionality and every sub-transformation $T_{k}$ must preserve dimensionality. This means that evaluating $T$ incurs an increasing computational cost as dimensionality grows. This constraint is at direct odds with our desire to use as many steps in the flow as possible. \citet{dinh2016density} proposed side-stepping this issue by way of a \textit{multi-scale architecture}.  At regular intervals in the steps of the flow when going from $\rvx$ to $\rvu$, some number of sub-dimensions of $\rvz_{k}$ are clamped and no additional transformation is applied.  One can think of this as implementing a skip-connection, mapping those dimensions directly to the corresponding dimensions in the final representation $\rvu$: $(\ervu_{j},\ldots, \ervu_{i}) = (\ervz_{k,j},\ldots, \ervz_{k,i}) $ where $k$ is the step at which the clamping is applied.  All $K$ steps are applied to only a small subset of dimensions, which is less costly than applying all steps to all dimensions. \citet{dinh2016density} also argue that these skip-connections help with optimization, distributing the objective throughout the full depth of the flow.

Besides having this practical benefit, multi-scale architectures are a natural modeling choice for granular data types such as pixels \citep{dinh2016density, kingma2018glow} and waveforms \citep{prenger2018waveglow, kim2018flowavenet}.  The macro-structures that we often care about---such as shapes and textures, in the case of images---typically do not need all $D$ dimensions to be described.  \citet{dinh2016density} showed that multi-scale architectures do indeed encode more global, semantic information in the dimensions that undergo all transformations.  On the other hand, dimensions that are factored out earlier in the flow represent lower-level information; see \citet{dinh2016density}'s Appendix D for demonstrations.

\section{Constructing Flows Part II: Continuous-Time Transformations}
\label{sec:constructing_continuous}

In the preceding section, we considered constructing flows by parameterizing a one-step transformation $\rvz_{k} = T_{k}(\rvz_{k-1})$, several of which are then composed to create a flow of $K$ discrete steps.
An alternative strategy is to construct flows in \emph{continuous time} by parameterizing the flow's infinitesimal dynamics, and then \emph{integrating} to find the corresponding transformation.  In other words, we construct the flow by defining an \textit{ordinary differential equation} (ODE) that describes the flow's evolution in time.  We call these `continuous-time' flows as they evolve according to a real-valued scalar variable analogous to the number of steps.  We call this scalar `time' as it determines how long the dynamics are run.  In this section, we will describe this class of continuous-time flows and summarize numerical tools necessary for their implementation. 

\subsection{Definition}

Let $\rvz_{t}$ denote the flow's state at time $t$ (or `step' $t$, thinking in the discrete setting). Time $t$ is assumed to run continuously from $t_{0}$ to $t_{1}$, such that $\rvz_{t_{0}} = \rvu$ and $\rvz_{t_{1}} = \rvx$.  A \textit{continuous-time flow} is constructed by parameterizing the time derivative of $\rvz_t$ with a function $g_\vphi$ with parameters $\vphi$, yielding the following \textit{ordinary differential equation} (ODE): \begin{equation}\label{eq:resnet-ivp}
    \frac{d \rvz_t }{d t} = g_{\vphi}(t, \rvz_t).
\end{equation}
The function $g_{\vphi}$ takes as inputs both the time $t$ and the flow's state $\rvz_t$, and outputs the time derivative of $\rvz_t$ at time $t$.  The only requirements for $g_{\vphi}$ are that it be uniformly Lipschitz continuous in $\rvz_t$ (meaning that there is a single Lipschitz constant that works for all $t$) and continuous in $t$ \citep{chen2018neural}. From \textit{Picard's existence theorem}, it follows  that satisfying these requirements ensures that the above ODE has a unique solution \citep{coddington1955theory}. Many neural-network layers meet these requirements \citep{Gouk:2018:Lipschitz}, and unlike the architectures described in \Cref{sec:constructing_discrete} that require careful structural assumptions to ensure invertibility and tractability of their Jacobian determinant, $g_{\vphi}$ has no such requirements.  

To compute the transformation $\rvx = T(\rvu)$, we need to run the dynamics forward in time by integrating: \begin{equation}\label{eq:resnet-T}
    \rvx = \rvz_{t_{1}} = \rvu + \integralab{g_{\vphi}(t, \rvz_t)}{t}{t=t_{0}}{t_{1}}.
\end{equation}  The inverse transform $T^{-1}$ is then: \begin{equation}\label{eq:resnet-T-inverse}
    \rvu = \rvz_{t_{0}} = \rvx + \integralab{g_{\vphi}(t, \rvz_t)}{t}{t=t_{1}}{t_{0}}  = \rvx - \integralab{g_{\vphi}(t, \rvz_t)}{t}{t=t_{0}}{t_{1}},
\end{equation} where in the right-most expression we used the fact that switching the limits of integration is equivalent to negating the integral.  We write the inverse in this last form to show that, unlike many flows comprised of discrete compositions (\Cref{sec:constructing_discrete}), continuous-time flows \emph{have the same computational complexity in each direction}.  In consequence, choosing which direction is the forward and which is the inverse is not a crucial implementation choice as it is for e.g.~autoregressive flows.

The change in log density for continuous-time flows can be characterized directly as \citep{chen2018neural}: \begin{equation}\label{eq:continuous-time-flows-logp}
    \frac{d \log p(\rvz_{t})}{d t} =  -\trace{ J_{g_\vphi(t, \cdot)} (\rvz_{t}) }
\end{equation} where $\trace{\cdot}$ denotes the trace operator and $J_{g_{\vphi}(t, \cdot)}(\rvz_{t})$ is the Jacobian of $g_\vphi(t, \cdot)$ evaluated at $\rvz_{t}$. The above equation can be obtained as a special case of the \textit{Fokker--Planck equation} for zero diffusion \citep{risken1996fokkerplanck}. While the trace operator at first glance seems more computationally tractable than a determinant, in practice it still requires $\mathcal{O}(D)$ backpropagation passes to obtain the diagonal elements of $J_{g_{\vphi}(t, \cdot)}(\rvz_{t})$. Similarly to contractive residual flows (\Cref{sec:contractive_residual_flows}), the \textit{Hutchinson's trace estimator} \citep{Hutchinson:1990:estimator} can be used to obtain an approximation in high-dimensional settings \citep{grathwohl2018ffjord}:
\begin{equation}
    \trace{J_{g_\vphi(t, \cdot)}(\rvz_t)} \approx \rvv^\top J_{g_\vphi(t, \cdot)}(\rvz_t) \,\rvv,
\end{equation}
where $\rvv$ can be any $D$-dimensional random vector with zero mean and unit covariance. The Jacobian-vector product $\rvv^\top J_{g_\vphi(t, \cdot)}(\rvz_t)$ can be computed in a single backpropagation pass, which makes the Hutchinson trace estimator about $D$ times more efficient than calculating the trace exactly. \citet{chen2019diffop} propose an alternative solution in which the architecture of $g_\vphi$ is carefully constrained so that the exact Jacobian trace can be computed in a single backpropagation pass.  

By integrating the derivative of $\log p(\rvz_t)$ over time, we obtain an expression for the log density of $\rvx$ under the continuous-time flow: \begin{equation}
    \log \px(\rvx) = \log \pu(\rvu) - \integralab{\trace{ J_{g_{\vphi}(t, \cdot)} (\rvz_{t}) }}{t}{t=t_{0}}{t_{1}}.
\end{equation}  Evaluating the forward transform and the log density can be done simultaneously by computing the following combined integral: \begin{equation}
\begin{bmatrix} 
\rvx \\
  \log \px(\rvx)
\end{bmatrix} = 
\begin{bmatrix} 
\rvu  \\
  \log \pu(\rvu)
\end{bmatrix} + \integralab{
\begin{bmatrix} 
g_{\vphi}(t, \rvz_t)  \\
 -\trace{ J_{g_{\vphi}(t, \cdot)} (\rvz_{t}) }
\end{bmatrix}}{t}{t=t_{0}}{t_{1}}.
\end{equation} Computation is not analytically feasible for a general $g_{\vphi}$. In practice, a numerical integrator is used, as we discuss next.

\subsection{Solving and Optimizing Continuous-Time Flows}
Due to continuous-time flows being defined by an ODE, the vast literature on numerical ODE solvers and the corresponding software can be leveraged to implement these flows.  See \citet{suli2010numerical} for an accessible introduction to numerical methods for ODEs.  While there are numerous numerical methods that could be employed, below we briefly describe \textit{Euler's method} and the \textit{adjoint method}.      

\subsubsection{Euler's Method and Equivalence to Residual Flows}  Perhaps the simplest numerical technique one can apply is \textit{Euler's method}. The idea is to first discretize the ODE using a small step-size $\epsilon > 0$ as follows: \begin{equation}
    \rvz_{t+\epsilon} \approx f_\vphi(\rvz_{t}) = \rvz_{t} + \epsilon\, g_{\vphi}(t, \rvz_{t})
    \end{equation} 
with the approximation becoming exact as $\epsilon \rightarrow 0$. The ODE can then be (approximately) solved by iterating the above computation starting from $\rvz_{t_0} = \rvu$ until obtaining $\rvz_{t_1} = \rvx$. This way, the parameters $\vphi$ can be optimized with gradients computed via backpropagation through the ODE solver. It is relatively straightforward to use other discrete solvers such as any in the \textit{Runge--Kutta} family.  The discretized forward solution would be backpropagated through just as with Euler's method. 

Interestingly, the Euler approximation implements the continuous-time flow as a discrete-time residual flow of the class described in  \Cref{eq:residual_flow}. Having assumed that $g_\vphi(t, \cdot)$ is uniformly Lipschitz continuous with a Lipschitz constant $L$ independent of $t$, it immediately follows that $\epsilon\,g_\vphi(t, \cdot)$ is contractive for any $\epsilon < 1/L$. Hence, for small enough $\epsilon$ we can think of the above Euler discretization as a particular instance of a contractive residual flow (\Cref{sec:contractive_residual_flows}).

Approximating continuous-time flows using discrete-time residual flows gives us insight on \Cref{eq:continuous-time-flows-logp}'s description of the time evolution of $\log p(\rvz_t)$. Using the Taylor-series expansion of \Cref{eq:residual_flow_logdet_taylor}, we can write the log absolute Jacobian determinant of $f_\vphi$ as follows:
\begin{equation}
    \log \abs{\det J_{f_\vphi}(\rvz_t)} = \sum_{k=1}^\infty \frac{(-1)^{k+1}\epsilon^k}{k}\trace{J^k_{g_\vphi(t, \cdot)}(\rvz_t)} 
    = \epsilon\,\trace{J_{g_\vphi(t, \cdot)}(\rvz_t)} + \bigo{\epsilon^2}.
\end{equation}
Substituting the above into the change-of-variables formula and rearranging we get:
\begin{align}
    &\log p(\rvz_{t+\epsilon}) =  \log p(\rvz_{t}) - \epsilon\,\trace{J_{g_\vphi(t, \cdot)}(\rvz_t)} + \bigo{\epsilon^2} \Rightarrow\\
    &\frac{\log p(\rvz_{t+\epsilon}) -  \log p(\rvz_{t})}{\epsilon} = -\trace{J_{g_\vphi(t, \cdot)}(\rvz_t)} + \bigo{\epsilon},
\end{align}
from which we directly obtain \Cref{eq:continuous-time-flows-logp} by letting $\epsilon\rightarrow0$.

\subsubsection{The Adjoint Method}  
\citet{chen2018neural} proposed an elegant alternative to the discrete, fixed-step methods mentioned above.  For a general optimization target $\mathcal{L}(\rvx; \vphi)$ (such as log likelihood), they show that the gradient $\partial \mathcal{L} / \partial \rvz_{t}$ with respect to the flow's intermediate state $\rvz_{t}$ can be characterized by the following ODE:
\begin{equation}
    \frac{d}{dt}\br{\frac{\partial\mathcal{L}}{\partial\rvz_t}} = -\left(\frac{\partial\mathcal{L}}{\partial\rvz_t}\right)^\top \frac{\partial g_{\vphi}(t, \rvz_t)}{\partial\rvz_t},
\end{equation}
a result known widely as the \textit{adjoint sensitivity method} \citep{pontryagin1962mathematical}. In neural-network terminology, the adjoint method can be thought of as the continuous-time analog of backpropagation.  The gradient with respect to $\vphi$ can be computed by: \begin{equation}
    \frac{\partial \mathcal{L}(\rvx; \vphi)}{\partial \vphi} = \integralab{\frac{\partial \mathcal{L}}{\partial \rvz_{t}} \ \frac{\partial g_{\vphi}(t, \rvz_{t})}{\partial \vphi}}{t}{t=t_{1}}{t_{0}}.
\end{equation}  Optimization of $\vphi$ can then be done with stochastic gradients.

Formulating gradient computation as a separate ODE means that both forward evaluation and gradient computation can be treated by black-box ODE solvers, without the need to backpropagate through the solver's computational graph. This results in significant practical benefits since backpropagating through a solver is costly both in terms of computation and memory requirements. Another benefit is that a more sophisticated solver can allocate instance-dependent computation based on a user-specified tolerance, a common hyperparameter in most off-the-shelf solvers. At test time, this tolerance level can be tuned based on runtime or other constraints on computation.

\section{Generalizations}
\label{sec:generalizations}

So far, we have discussed normalizing flows as invertible transformations in Euclidean space. In this section, we go beyond the standard definition and explore more general classes of probability transformations. We show how more general types of flow can be derived from a unifying framework, point to specific implementations that extend the standard definition, and explore the frontiers of normalizing-flow research.

\subsection{General Probability-Transformation Formula}\label{sec:generalizations:formula}

\begin{figure}[t]
    \centering
    \includegraphics[height=4cm]{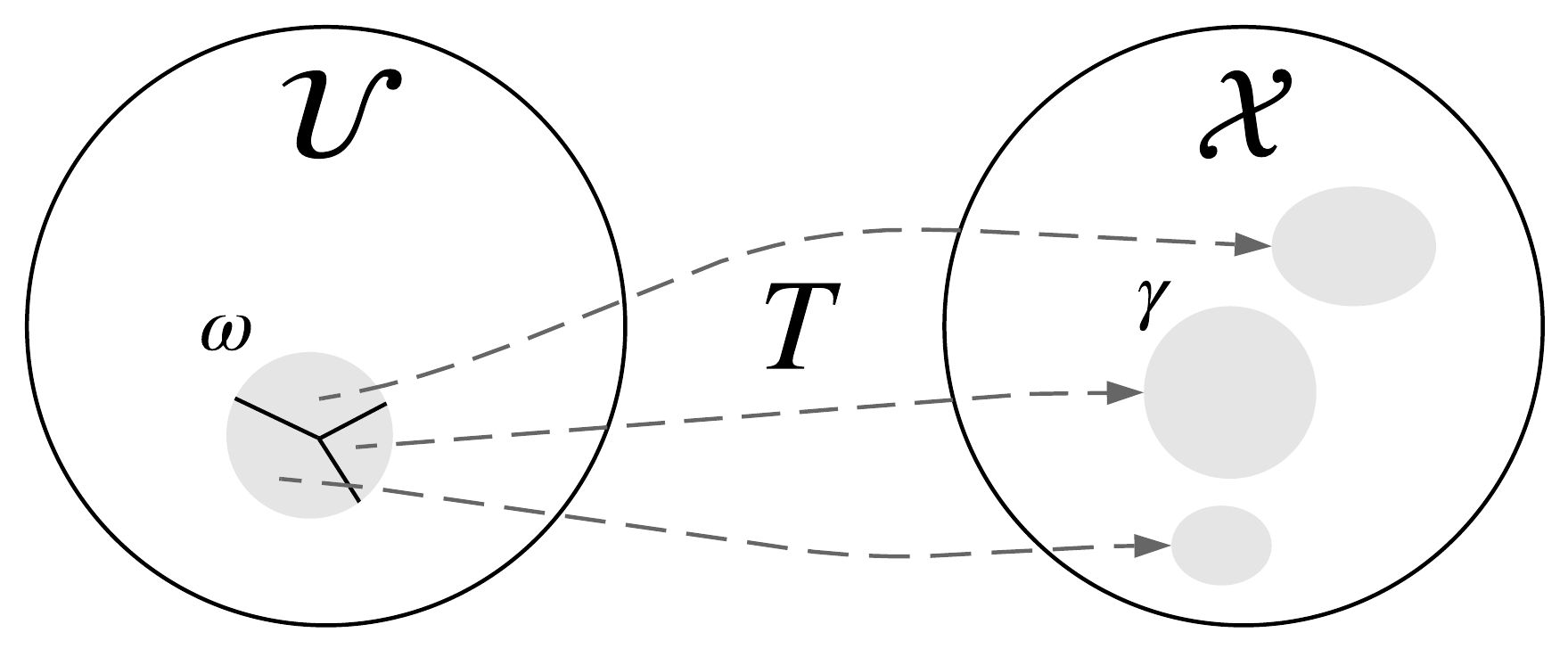}
    \caption{Illustration of the general probability-transformation formula. The probability $\prob(\rvu \in \omega)$ must be equal to the probability $\prob(\rvx \in \gamma)$ for any $\omega \subseteq \setU$, and this relation will remain true even if the image of $\omega$ through $T$ has disconnected components.
    \label{fig:general_flow}}
\end{figure}

Assume a probability density $\pu: \setU \rightarrow [0, +\infty)$ defined on a set $\setU$, and a transformation $T: \setU \rightarrow \setX$ that maps the set $\setU$ to the set $\setX$. The density $\pu(\rvu)$ together with the transformation $T$ induce a probability density $\px: \setX \rightarrow [0, +\infty)$ on $\setX$ via the general relation:
\begin{align}
     \integralx{\pu(\rvu)}{\mu(\rvu)}{ \rvu \in \omega \subseteq \setU } &= \integralx{\px(\rvx)}{\nu(\rvx)}{\rvx \in \gamma \subseteq \setX}
    \quad\forall \omega \subseteq \setU. \label{eq.general.flow}
\end{align}
This is illustrated in \Cref{fig:general_flow}. In the above formula, $\gamma = \set{T(\rvu)\,|\,\rvu \in \omega}$ is the image of $\omega$ under $T$, and $d\mu(\rvu)$ and $d\nu(\rvx)$ are the integration measures in $\setU$ and $\setX$ with respect to which the densities $\pu(\rvu)$ and $\px(\rvx)$ are defined. The formula has a simple interpretation as \emph{conservation of probability measure}: its LHS is the probability that a sample from $\pu(\rvu)$ falls in $\omega \subseteq \setU$, whereas the RHS is the probability that a sample from $\px(\rvx)$ falls in $\gamma \subseteq \setX$. Since $\gamma$ is the image of $\omega$ under $T$, these two probabilities must be the same for any $\omega$.

Standard flow-based models are a special case of this formula, where $T$ is a diffeomorphism (i.e.~an differentiable invertible transformation with differentiable inverse), the sets $\setU$ and $\setX$ are equal to the $D$-dimensional Euclidean space $\R^D$, and $d\mu(\rvu)$ and $d\nu(\rvx)$ are equal to the Lebesgue measure in $\R^D$. In that case, the conservation of probability measure can be written as:
\begin{align}
     \integralx{\pu(\rvu)}{\rvu}{ \rvu \in \omega } &= \integralx{\px(\rvx)}{\rvx}{\rvx \in \gamma}.
\end{align}
Since $T$ is a diffeomorphism, we can express the LHS integral via the change of variable $\rvu = T^{-1}(\rvx)$ as follows \citep{rudin2006real}:
\begin{align}
     \integralx{\pu\br{T^{-1}(\rvx)}\abs{\det J_{T^{-1}}(\rvx)}}{\rvx}{ \rvx \in \gamma } &= \integralx{\px(\rvx)}{\rvx}{\rvx \in \gamma}.
\end{align}
Since the above must be true for any $\gamma\subseteq\setX$, it follows that:
\begin{equation}
    \px(\rvx) = \pu\br{T^{-1}(\rvx)}\abs{\det J_{T^{-1}}(\rvx)},
\end{equation}
which is the familiar formula for the density of a flow-based model.

Using the conservation of probability measure as a general framework, we study different transformations $T$, sets $\setU$ and $\setX$, and integration measures $d\mu(\rvu)$ and $d\nu(\rvx)$. In the following sections, we explore further possibilities and potential implementations.

\subsection{Piecewise-Invertible Transformations and Mixtures of Flows}
\label{sec:generalizations_piecewise_invertible}

\begin{figure}[t]
    \centering
    \begin{subfigure}[b]{0.49\textwidth}
        \includegraphics[width=\textwidth]{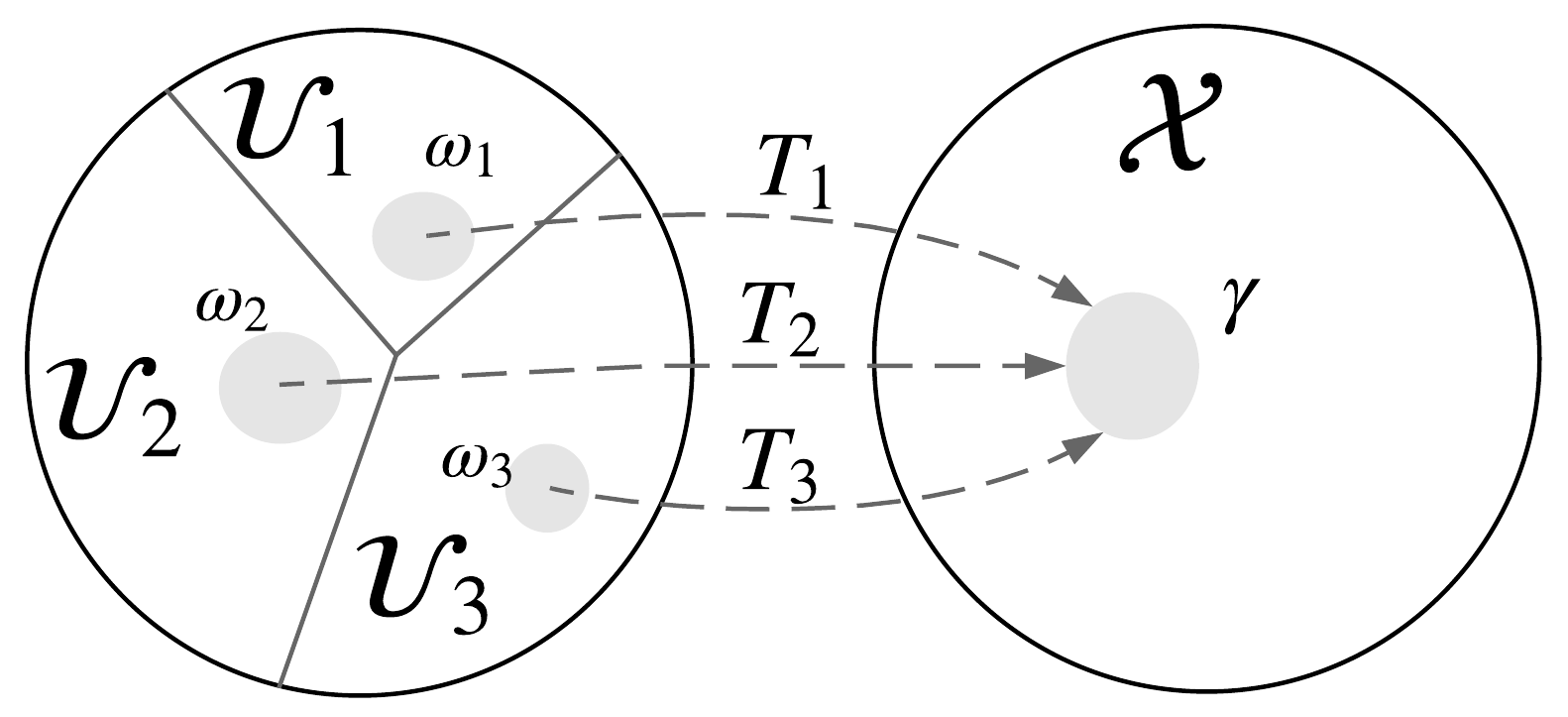}
        \caption{Many-to-one\label{fig:piecewise_inverible:many_to_one}}
    \end{subfigure}
    \begin{subfigure}[b]{0.49\textwidth}
        \includegraphics[width=\textwidth]{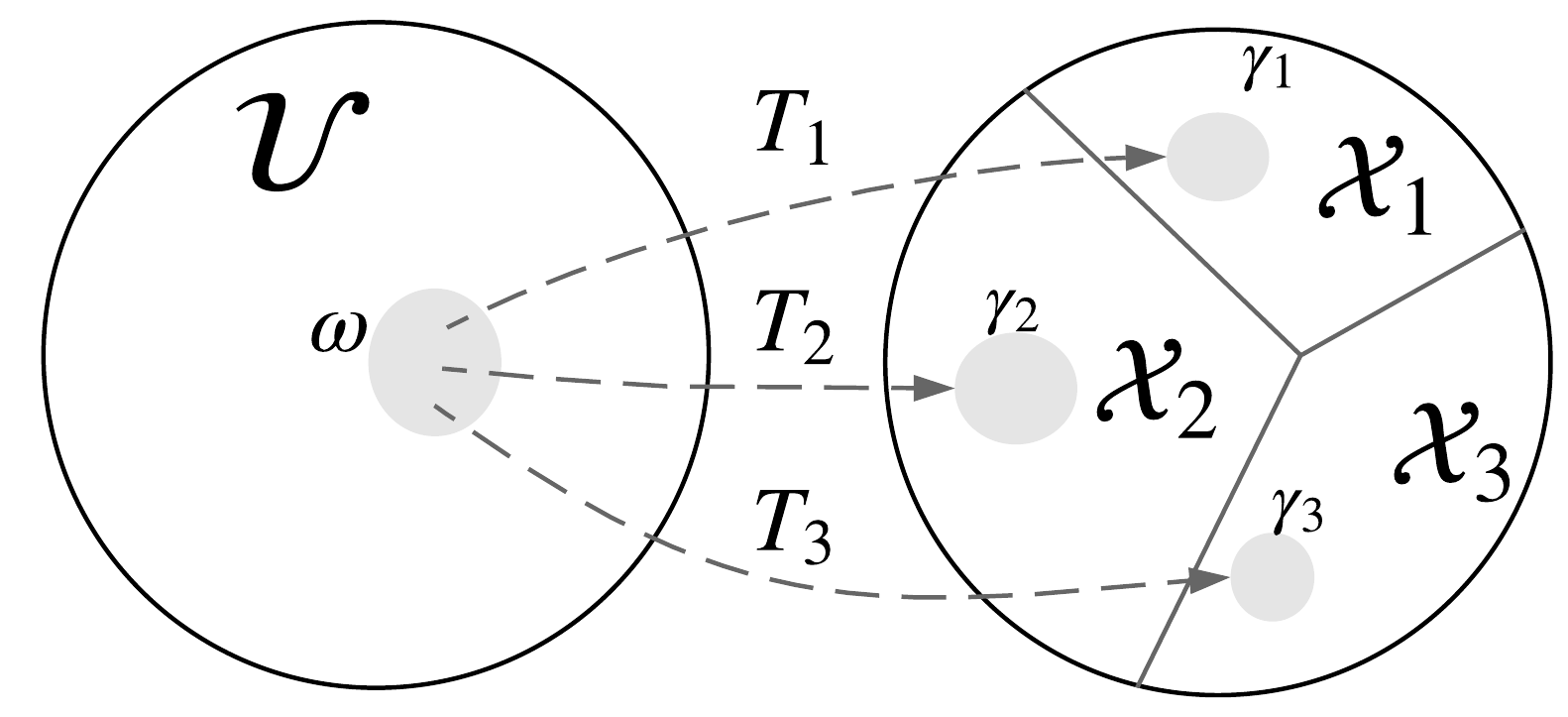}
        \caption{One-to-many\label{fig:piecewise_inverible:one_to_many}}
    \end{subfigure}
    \caption{Illustrations of a piecewise-invertible transformations.\label{fig:piecewise_inverible}}
\end{figure}

We can extend the transformation $T$ to be \textit{many-to-one}, whereby multiple $\rvu$'s are mapped to the same $\rvx$. One possibility is for $T$ to be \textit{piecewise invertible}, in which case we can partition $\setU$ into a countable collection of non-overlapping subsets $\set{\setU_i}_{i\in\setI}$ such that the restriction of $T$ to $\setU_i$ is an invertible transformation $T_i :\setU_i \rightarrow\setX$ (\Cref{fig:piecewise_inverible:many_to_one}). Then, the conservation of probability measure can be written as:
\begin{align}
     \sum_{i\in\setI}\integralx{\pu(\rvu)}{\rvu}{ \rvu \in \omega_i } &= \integralx{\px(\rvx)}{\rvx}{\rvx \in \gamma},
\end{align}
where each $\omega_i \subseteq \setU_i$ is the image of $\gamma$ under $T_i^{-1}$. Using the change of variables $\rvx = T_i(\rvu)$ for each corresponding integral on the LHS, we obtain:
\begin{align}
     \integralx{\sum_{i\in\setI}\pu\br{T_i^{-1}(\rvx)}\abs{\det J_{T_i^{-1}}(\rvx)}} {\rvx}{ \rvx \in \gamma }
     = \integralx{\px(\rvx)}{\rvx}{\rvx \in \gamma}.
\end{align}
Since the above must be true for all $\gamma\subseteq\setX$, it follows that:
\begin{equation}
    \px(\rvx) = \sum_{i\in\setI}\pu\br{T_i^{-1}(\rvx)}\abs{\det J_{T_i^{-1}}(\rvx)}.
\end{equation}
The above can be interpreted as a \textit{mixture of flows}, where the $i$-th flow has transformation $T_i$, base density $\pu(\rvu)$ restricted to $\setU_i$ and mixture weight equal to $\prob(\rvu\in\setU_i)$.

An alternative would be to take the transformation $T$ to be \textit{one-to-many}, where a single $\rvu$ is mapped to multiple $\rvx$'s. One possibility, proposed by \citet{dinh2019rad} and known as \textit{real and discrete} (RAD), is to think of $T$ as the `inverse' of a piecewise-invertible transformation $R:\setX\rightarrow\setU$. We partition $\setX$ into a countable collection of non-overlapping subsets $\set{\setX_i}_{i\in\setI}$ such that the restriction of $R$ to $\setX_i$ is an invertible transformation whose inverse is $T_i:\setU\rightarrow\setX_i$ (\Cref{fig:piecewise_inverible:one_to_many}). Then, $T$ simply maps $\rvu$ to $\set{T_i(\rvu)}_{i\in\setI}$\@. To make sense of the above as a generative model of $\rvx$, we can select one of $\set{T_i(\rvu)}_{i\in\setI}$ at random from some distribution $p(i\g\rvu)$ over $\setI$ that can depend on $\rvu$. This is equivalent to extending the input space to $\setU' = \setU\times\setI$, defining the input density $p(\rvu, i) = \pu(\rvu)\,p(i\g\rvu)$ with respect to the Lebesgue measure on $\setU$ and the counting measure on $\setI$, and defining the extended transformation $T':\setU'\rightarrow\setX$ to be $T'(\rvu, i) = T_i(\rvu)$. Then, the conservation of measure in the extended space can be written as:
\begin{align}
     \sum_{i\in\xi}\integralx{\pu(\rvu)\,p(i\g\rvu)}{\rvu}{ \rvu \in \omega } &= \sum_{i\in\xi}\integralx{\px(\rvx)}{\rvx}{\rvx \in \gamma_i}
     \quad\forall\xi\subseteq\setI,\,\,\forall\omega\subseteq\setU,
\end{align}
where $\gamma_i\subseteq\setX_i$ is the image of $\omega$ under $T_i$. Using the change of variables $\rvx = T_i(\rvu)$ for each corresponding integral on the LHS, we obtain:
\begin{align}
     \sum_{i\in\xi}\integralx{\pu\br{T_i^{-1}(\rvx)}\,p\br{i\g T_i^{-1}(\rvx)}\abs{\det{J_{T_i^{-1}}(\rvx)}}}{\rvx}{ \rvx \in \gamma_i } &= \sum_{i\in\xi}\integralx{\px(\rvx)}{\rvx}{\rvx \in \gamma_i}.
\end{align}
The above must be true for all $\xi\subseteq\setI$ and $\gamma_i\subseteq\setX$, and by definition $T_i^{-1}(\rvx) = R(\rvx)$ for all $\rvx\in\setX_i$, therefore:
\begin{equation}
    \px(\rvx) = \pu\br{R(\rvx)}\,p\br{i(\rvx)\g R(\rvx)}\abs{\det{J_{R}(\rvx)}},
\end{equation}
where $i(\rvx)$ indexes the subset of $\setX$ in which $\rvx$ belongs. The above can be interpreted as a mixture of flows with non-overlapping components, where the $i$-th component uses transformation $T_i:\setU\rightarrow\setX_i$, base distribution $\pu(\rvu\g i) \propto \pu(\rvu)\,p(i\g\rvu)$ and mixture weight $p(i) = \integral{\pu(\rvu)\,p(i\g\rvu)}{\rvu}$.

\subsection{Flows for Discrete Random Variables}
\label{sec:discrete_random_variables}

Using the general probability-transformation formula, we can extend normalizing flows to discrete sets $\setU$ and $\setX$. Taking the integration measures $d\mu(\rvu)$ and $d\nu(\rvx)$ to be the counting measure, we can write the conservation of measure as follows:
\begin{equation}
    \sum_{\rvu\in\omega}\pu(\rvu) = \sum_{\rvx\in\gamma}\px(\rvx).
\end{equation}
Letting $\gamma = \set{\rvx}$ for an arbitrary $\rvx$, we obtain:
\begin{equation}
    \px(\rvx) = \sum_{
         \rvu\in\omega
    }\pu(\rvu)
    \quad\text{where}\quad
    \omega=\set{\rvu\,|\,T(\rvu) = \rvx}.
\end{equation}
Finally, by restricting $T$ to be bijective (assuming $\setU$ and $\setX$ have the same cardinality), we obtain a normalizing flow on discrete random variables, whose density\footnote{The probability mass function $\px(\rvx)$ can be though of as a density with respect to the counting measure.} is given by:
\begin{equation}\label{eq:discrete_cov}
    \px(\rvx) = \pu\br{T^{-1}(\rvx)}.
\end{equation}
We will refer to the above type of normalizing flow as a \textit{discrete flow}.
Unlike standard flows, discrete flows don't involve a Jacobian term in their density calculation.

\citet{hoogeboom2019integer} proposed a discrete flow for $\setU = \setX = \sZ^D$ based on affine autoregressive flows (\Cref{sec:autoregressive_flows}). Specifically, they implement the transformer $\tau:\sZ\rightarrow\sZ$ as follows:
\begin{equation}
    \tau(\ervz_i; \beta_i) = \ervz_i + \mathrm{round}(\beta_i),
\end{equation}
where $\beta_i$ is given by the conditioner and is a function of $\rvz_{<i}$, and $\mathrm{round}(\cdot)$ maps its input to the nearest integer. Similarly, \citet{tran2019discrete} proposed a discrete flow for $\setU=\setX=\set{0, \ldots, K-1}^{D}$ also based on affine autoregressive flows and whose transformer $\tau: \set{0, \ldots, K-1}\rightarrow\set{0, \ldots, K-1}$ is given by:
\begin{equation}
    \tau(\ervz_i; \alpha_i, \beta_i) = \br{\alpha_i\ervz_i + \beta_i}\!\!\!\!\mod K,
\end{equation}
where $\alpha_i$ and $\beta_i$ are each given by the argmax of a $K$-dimensional vector outputted by the conditioner. The above transformer can be shown to be bijective whenever $\alpha_i$ and $K$ are coprime. To backpropagate through the discrete-valued functions $\mathrm{round}(\cdot)$ and argmax, both \citet{hoogeboom2019integer} and \citet{tran2019discrete} use the straight-through gradient estimator \citep{bengio2013estimating}.

Compared to flows on $\R^D$, discrete flows have notable theoretical limitations. As shown in \Cref{sec:flows:expressivity}, under mild conditions, flows on $\R^D$ can transform any base density $\pu(\rvu)$ to any target density $\px(\rvx)$. However, this is not true for discrete flows---at least not as long as they are defined using a bijective transformation. For example, if the base distribution $\pu(\rvu)$ of a discrete flow is uniform, $\px(\rvx)$ will necessarily be uniform too. More generally, due to bijectivity, for every $\rvx$ there must be a $\rvu$ such that $\px(\rvx) = \pu(\rvu)$. In other words, a discrete flow can never change the values of $\pu(\rvu)$, only permute them. Recalling the change-of-variables formula for discrete flows (\Cref{eq:discrete_cov}), the absence of a Jacobian term may seem to be a computational boon that makes discrete flows preferable to continuous ones.  However, this absence is also a restriction. 

Another useful property of standard flows is that they can model any target density using a fully factorized base distribution $\pu(\rvu) = \prod_{i=1}^D \pu(\ervu_i)$. Such a base density can be evaluated and sampled from in parallel, which is important for scalability to high dimensions. Nonetheless, a discrete flow with a fully factorized base distribution may not be able to model all target densities, even if the base distribution is learned. To see this, consider the target density $\px(\ervx_1, \ervx_2)$ with $\ervx_1\in\set{0, 1}$, $\ervx_2 \in\set{0, 1}$ given by:
\begin{equation}
    \begin{array}{ccc}
        \px(0, 0) = 0.1 && \px(0, 1) = 0.3\hphantom{.}\\
        \px(1, 0) = 0.2 && \px(1, 1) = 0.4.
    \end{array}
\end{equation}
Assuming $\setU=\setX=\set{0, 1}^2$, all a discrete flow can do to $\px(\ervx_1, \ervx_2)$ is permute the $4$ values of the probability table:
$$\begin{bmatrix}0.1 & 0.3\\0.2 & 0.4\end{bmatrix}$$
into a new $2\times 2$ probability table. Therefore, to model $\px(\ervx_1, \ervx_2)$ with a factorized base $\pu(\ervu_1, \ervu_2) = \pu(\ervu_1)\pu(\ervu_2)$, there must be a permutation of the above table such that the permuted table is of rank $1$ (so it can be factorized as the outer product of two vectors). However, after checking all $4!=24$  permutations, we find that all permuted tables are of rank $2$, which shows that the above target can't be modeled using a factorized base. This limitation means that, in practice, we may need to explicitly incorporate dependencies  in the base distribution to increase the model's capacity. Both \citet{hoogeboom2019integer} and \citet{tran2019discrete} take this approach, modeling the base distribution autoregressively.

A possible way to overcome this limitation, pointed out by \citet{vandenberg2020idfpp}, is to embed $\setU$ and $\setX$ into extended spaces $\setU'$ and $\setX'$, such that the base distribution factorizes in the extended space. In the above example, we could take $\setU'=\setX'=\set{0, 1, 2, 3}^2$, so that the probability table becomes:
$$\begin{bmatrix}0.1 & 0.3 & 0 & 0\\0.2 & 0.4 & 0 & 0\\0 & 0 & 0 & 0\\0 & 0 & 0 & 0\end{bmatrix}.$$
Then, a discrete flow on the extended space can rearrange the probability table into:
$$\begin{bmatrix}0.1 & 0 & 0 & 0\\0.2 & 0 & 0 & 0\\0.3 & 0 & 0 & 0\\0.4 & 0 & 0 & 0\end{bmatrix},$$
which is rank $1$ and thus can be factorized. 

\subsection{Flows on Riemannian Manifolds}

Next, we can extend normalizing flows to Riemannian manifolds embedded in a higher-dimensional space \citep{gemici2016normalizing, wang2019riemannian}.
Assume that $\setX$ is a $D$-dimensional manifold embedded in $\R^{M}$ via an injective map $T:\R^D \rightarrow \R^{M}$ with $M \geq D$.
The embedding map $T$ can be thought of as a curvilinear coordinate system that traces points $\rvx\in\R^M$ on the manifold using coordinates $\rvu \in \R^D$. The map $T$ induces a metric $G(\rvu)$ on the tangent space of $\setX$ at $\rvx=T(\rvu)$, given by \citep{kobayashi1963foundations}:
\begin{equation}
    G(\rvu) = J_T(\rvu)^\top J_T(\rvu).
\end{equation}
As a result, an infinitesimal volume on $\setX$ is given by $d\nu(\rvx) = \sqrt{\det{G(\rvu)}}\,d\rvu$.
A formula relating the density on $\setX$ to that on the Euclidean space $\R^D$ can be derived from the conservation of measure (\Cref{eq.general.flow})
by setting $\setU=\R^D$, taking $d\mu(\rvu)$ to be the Lebesgue measure on $\R^D$, and reparameterizing $d\nu(\rvx) = \sqrt{\det{G(\rvu)}}\,d\rvu$, which yields:
\begin{align}
\integralx{\pu(\rvu)}{\rvu}{\omega} = \integralx{\px(T(\rvu)) \sqrt{\det G(\rvu)}}{\rvu}{\omega}.
\end{align}
Since the above must be true for any $\omega\subseteq\R^D$, it follows that:
\begin{align}
\pu(\rvu) = \px(T(\rvu)) \sqrt{\det G(\rvu)},
\end{align}
which gives the density on $\R^D$ as a function of the density on the manifold. If we restrict the range of $T$ to $\setX$, we can define the inverse mapping $T^{-1}:\setX\rightarrow \R^D$ and then use it to obtain the density on the manifold:
\begin{align}
\px(\rvx) = \pu\br{T^{-1}(\rvx)} \left[\det G\br{T^{-1}(\rvx)}\right]^{-1/2}.
\label{eq:riemman_flow}
\end{align}
The usual density-transformation formula for flows on $\R^D$ is a particular case of \Cref{eq:riemman_flow}. Taking $\setX = \R^D$ and $M=D$, the Jacobian $J_T(\rvu)$ becomes $D\times D$, and the infinitesimal volume on $\setX$ simplifies to:
\begin{equation}
    d\nu(\rvx) = \sqrt{\br{\det{J_T(\rvu)}}^2}\,d\rvu = \abs{\det{J_T(\rvu)}}\,d\rvu,
\end{equation}
which retrieves the standard flow on $\R^D$.

\begin{figure}[t]
    \centering
    \includegraphics[height=4cm]{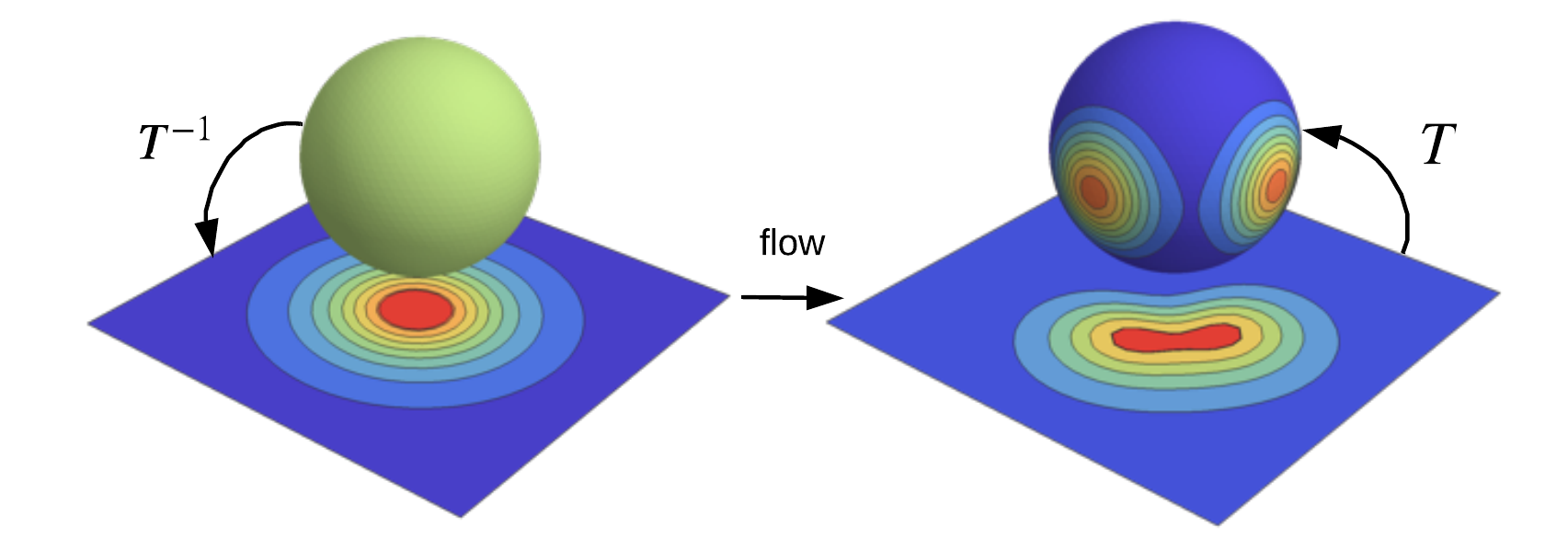}
    \caption{Density on a $2$-dimensional sphere formed by mapping the sphere to $\R^2$, transforming the density there, and mapping $\R^2$ back to the sphere.
    \label{fig:generalisations_reimann}}
\end{figure}

Using the above approach, we can define flows for which both $\setU$ and $\setX$ are $D$-dimensional Riemannian manifolds. We start from a base density defined on a manifold $\setU$, transform it to $\R^D$ using the inverse embedding map for $\setU$, perform any number of standard flow steps on $\R^D$, and finally transform the resulting density on the target manifold $\setX$ using the embedding map for $\setX$. We illustrate this approach in \Cref{fig:generalisations_reimann}, where $\setU$ and $\setX$ are \mbox{$2$-dimensional} spheres embedded in $\R^{3}$.

An important limitation of the above approach is that it assumes the existence of a differentiable invertible map $T:\R^D\rightarrow \setX$, whose inverse maps the manifold onto $\R^D$. However, such a map can only exist for manifolds that are homeomorphic, i.e.~topologically equivalent, to $\R^D$ \citep{kobayashi1963foundations}. There are several manifolds that are not homeomorphic to $\R^D$, including spheres, tori, and arbitrary products thereof, for which the above approach will be problematic. In the sphere example in \Cref{fig:generalisations_reimann}, any embedding map will create at least one coordinate singularity (akin to how all longitude lines meet at Earth's poles). Such coordinate singularities can manifest as points of infinite density, and thus create numerical instabilities in practice.

An alternative approach was proposed by \citet{falorsi2019reparameterizing} for the special case where the manifold is also a group, in which case it is known as a \textit{Lie group}. A Lie group can be reparameterized with respect to its Lie algebra (i.e.~the tangent space at the identity element) using the exponential map, which maps the Lie algebra to the Lie group. \citet{falorsi2019reparameterizing} show that this parameterization can be done analytically in certain cases; in general however, computing the exponential map has a cubic cost with respect to the dimensionality $D$.

\subsection{Bypassing Topological Constraints}

Due to $T$ being a diffeomorphism, a flow must preserve topological properties, which means that $\mathcal{U}$ and $\mathcal{X}$ must be homeomorphic, i.e.~topologically equivalent \citep{kobayashi1963foundations}. For example, a flow cannot map $\R^D$ to the sphere $\sS^D$, or $\R^D$ to $\R^{D'}$ for $D\neq D'$, as these spaces have different topologies. For the same reason, every intermediate space $\setZ_k$ along the flow's trajectory must be topologically equivalent to both $\mathcal{U}$ and $\mathcal{X}$. In the previous section, we discussed how this constraint can create problems for flows that map between Riemannian manifolds and Euclidean space.

This constraint's effect on continuous-time flows is prominent. In this case there is a continuum of intermediate spaces $\setZ_t$ for $t\in[t_0, t_1]$, all of which must be topologically equivalent. For example, a continuous-time flow cannot implement a transformation $T:\R\rightarrow\R$ with $T(1)=-1$ and $T(-1)=1$, due to any such transformation requiring an intersection at some intermediate stage \citep{dupont2019augmented}. One way to bypass this restriction is to instantiate the flow in a lifted space, as was proposed by \citet{dupont2019augmented}. Introducing $\rho$ auxiliary variables, the flow's transformation is now a map between $D + \rho$ dimensional spaces, $T:\R^{D+\rho} \rightarrow \R^{D+\rho}$.  Although the lifted spaces must  still be topologically equivalent, $T$ can now represent a wider class of functions when projected to the original $D$-dimensional space. However, density evaluation in the original $D$-dimensional space can no longer be done analytically, which removes a benefit of flows that makes them an attractive modeling choice in the first place.  Rather, the $\rho$ auxiliary variables must be numerically integrated in practice. \citet{dupont2019augmented}'s augmentation method can be thought of as defining a mixture of flows whereby the auxiliary variables serve as an index; integrating out the auxiliary variables is then akin to summing over the (infinitely many) mixture components.

For another example, if the target density is comprised of disconnected modes, the base density must have the same number of disconnected modes. This is true for all flows, not just continuous-time ones.  If the base density has fewer modes than the target (which will likely be the case in practice), the flow will be forced to assign a non-zero amount of probability mass to the `empty' space between the disconnected modes. To alleviate this issue, \citet{cornish2019localised} propose a latent-variable flow defined as follows:
\begin{equation}
    \rvu \sim \pu(\rvu), \quad\rvz \sim p(\rvz \g \rvu),\quad \rvx = T(\rvu; \rvz),
\end{equation}
where $\rvz$ is a latent variable. Just as with the auxiliary-variable formulation of \citet{dupont2019augmented}, we can think of the above as a mixture of flows, with $\rvz$ indexing each mixture component. Additionally, the above formulation is the same as RAD \citep{dinh2019rad} discussed in \Cref{sec:generalizations_piecewise_invertible}, except that here $\rvz$ can be continuous and the mixture components can overlap. However, $\px(\rvx)$ can no longer be calculated analytically in general. \citet{cornish2019localised} make do by performing variational inference, introducing an approximation $q(\rvz \g \rvx)$ that then allows for optimization via a lower bound on $\log \px(\rvx)$.

\subsection{Symmetric Densities and Equivariant Flows}

In many real-world applications, we have domain knowledge of the symmetries of the target density. 
For example, if we want to model a physical system composed of many interacting particles, the physical interactions between particles (e.g.~gravitational or electrostatic forces) are invariant to translations. Another example is modeling the configuration of a molecular structure, where there may be rotational symmetries in the molecule. Such symmetries result in a probability density (e.g.~of molecular configurations) that is invariant to specific transformations (e.g.~rotations). 

In such situations, it is desirable to build the known symmetries directly into the model, not only so that the model has the right properties, but also so that estimating the model is more data efficient. However, in general it is not trivial to build expressive probability densities that are also invariant to a prescribed set of symmetries.
Flows are a good candidate class of models to combine with such domain knowledge because we have a lot of control of how they deform an initial density. In this section, we explore ways to build flow-based models that respect a prescribed set of symmetries.

We say that $g$ is a \textit{symmetry} of the density $p:\R^D\rightarrow[0, +\infty)$ with representation an invertible matrix $\mathbf{R}_g\in\R^{D\times D}$ if the density remains invariant after transforming its input by $\mathbf{R}_g$, that is, if $p(\mathbf{R}_g\rvx) = p(\rvx)$ for all $\rvx\in\R^D$. For example, a standard-normal density is symmetric with respect to rotations, reflections, and axes permutations. It is easy to show that $\abs{\det\mathbf{R}_g} = 1$ for any symmetry $g$. Using the change of variables $\rvx'=\mathbf{R}_g^{-1}\rvx$, we have the following:
\begin{equation}
    1 = \integral{p(\rvx)}{\rvx} =  \integral{p(\mathbf{R}_g\rvx')\abs{\det\mathbf{R}_g}}{\rvx'} = \abs{\det\mathbf{R}_g}\integral{p(\rvx')}{\rvx'} = \abs{\det\mathbf{R}_g}.
\end{equation}
The set of all symmetries of a target density is closed under composition, is associative, has an identity element, and each element has an inverse---therefore forming a group $G$\@.

An important concept for dealing with symmetries in group theory is that of {\it equivariance}. We say that a transformation $T: \R^D \rightarrow \R^D$ is equivariant with respect to the group $G$ if $T(\mathbf{R}_g\rvu) = \mathbf{R}_{g} T(\rvu)$ for all $g\in G$ and $\rvu\in\R^D$\@---that is, we get the same result regardless of whether we transform the input or the output of $T$ by $\mathbf{R}_g$. It is straightforward to see that the composition of two equivariant transformations is also equivariant, and if $T$ is bijective its inverse is also equivariant.

These observations lead to the result shown in Lemma \ref{lemma.invariant_flow}, which provides a general mechanism for constructing flow-based models whose density is invariant with respect to a prescribed symmetry group $G$.
\begin{lemma}[\textbf{Equivariant flows}]\label{lemma.invariant_flow} Let $\px(\rvx)$ be the density function of a flow-based model with transformation $T: \R^D \rightarrow \R^D$ and base density $\pu(\rvu)$. If $T$ is equivariant with respect to $G$ and $\pu(\rvu)$ is invariant with respect to $G$, then $\px(\rvx)$ is invariant with respect to $G$.
\end{lemma}
\begin{proof}
The flow density evaluated at $\mathbf{R}_g\rvx$ for some $g\in G$ is equal to:
\begin{equation}
    \px(\mathbf{R}_g\rvx) = \pu\br{T^{-1}\br{\mathbf{R}_g\rvx}}\,\abs{\det J_{T^{-1}}(\mathbf{R}_g\rvx)}.
\end{equation}
From the equivariance of $T$ and hence of $T^{-1}$, we have that $T^{-1}\br{\mathbf{R}_g\rvx} = \mathbf{R}_g T^{-1}(\rvx)$. Taking the Jacobian of both sides, we obtain:
\begin{equation*}\begin{split}
    J_{T^{-1}}(\mathbf{R}_g\rvx)\, \mathbf{R}_g = \mathbf{R}_g J_{T^{-1}}(\rvx)
    \,&\Rightarrow\,
    \abs{\det J_{T^{-1}}(\mathbf{R}_g\rvx)} \abs{\det \mathbf{R}_g} = \abs{\det \mathbf{R}_g} \abs{\det{J_{T^{-1}}(\rvx)}} \\
    \,&\Rightarrow\,\abs{\det J_{T^{-1}}(\mathbf{R}_g\rvx)}=\abs{\det{J_{T^{-1}}(\rvx)}}.
\end{split}\end{equation*}
Finally, from the invariance of $\pu(\rvu)$ we have that $\pu\br{\mathbf{R}_g T^{-1}(\rvx)} = \pu\br{T^{-1}(\rvx)}$. Therefore, $\px(\mathbf{R}_g\rvx) = \px(\rvx)$ for all $g\in G$ and $\rvx\in\R^D$.
\end{proof}
The concept of equivariant flows was introduced concurrently by \citet{kohler2019equivariant} and \citet{rezende2019equivariant}. Both of these works explored equivariance in the context of  continuous-time flows. \citet{kohler2019equivariant} provided a concrete example of a continuous-time flow that is equivariant with respect to rigid translations and rotations, whereas
\citet{rezende2019equivariant} proposed a general method for enforcing equivariance using known symmetry generators.

In practice, taking the base density to be invariant with respect to a prescribed symmetry group is usually easy. What it more challenging is constructing the transformation $T$ to be equivariant. One approach is based on invariant functions with respect to $G$, that is, functions $f$ for which $f(\mathbf{R}_g\rvu) = f(\rvu)$ for all $g\in G$ and $\rvu\in\R^D$. This approach is outlined in Lemma \ref{lemma.equivariance_from_invariance} below.
\begin{lemma}[\textbf{Equivariance from invariance}]\label{lemma.equivariance_from_invariance} Let $f: \R^D \rightarrow \R$ be invariant with respect to $G$, and assume that $\mathbf{R}_g$ is orthogonal for all $g\in G$. Then $\nabla_\rvu f(\rvu)$ is equivariant with respect to $G$.
\end{lemma}
\begin{proof}
From the invariance of $f$ we have that $f(\mathbf{R}_g \rvu) = f(\rvu)$. Taking the gradient on both sides, we obtain:
\begin{equation*}\begin{split}
     \mathbf{R}_g^\top \nabla_{\mathbf{R}_g \rvu} f(\mathbf{R}_g \rvu) = \nabla_\rvu f(\rvu)\quad&\Rightarrow\quad
     \nabla_{\mathbf{R}_g \rvu} f(\mathbf{R}_g \rvu)  =  \mathbf{R}_g^{-\top} \nabla_\rvu f(\rvu)\\
     &\Rightarrow\quad
     \nabla_{\mathbf{R}_g \rvu} f(\mathbf{R}_g \rvu)  =  \mathbf{R}_g \nabla_\rvu f(\rvu).
\end{split}\end{equation*}
\end{proof}
Lemma \ref{lemma.equivariance_from_invariance} gives a general mechanism for constructing equivariant transformations with respect to symmetries with orthogonal representations. Several symmetries fall into this category, including rotations, reflections and axes permutations. The practical significance of Lemma \ref{lemma.equivariance_from_invariance} is that it is often easier to construct an invariant function than an equivariant one. For instance, any function of the $\ell_2$ norm $\norm{\rvu}$ is invariant with respect to symmetries with orthogonal representations, since $\norm{\mathbf{R}\rvu} = \norm{\rvu}$ for any orthogonal matrix $\mathbf{R}$.

\section{Applications}
\label{sec:applications}

Normalizing flows have two primitive operations: density calculation and sampling.  In turn, flows are effective in any application requiring a probabilistic model with either of those capabilities.  In this section, we summarize applications to probabilistic modeling, inference, supervised learning, and reinforcement learning.  

\subsection{Probabilistic Modeling}
\label{sec:applications:probabilistic_modeling}

Normalizing flows, due to their ability to be expressive while still allowing for exact likelihood calculations, are often used for probabilistic modeling of data.  For this application, we assume access to a finite number of draws $\rvx$ from some unknown generative process $\px^{*}(\rvx)$.  These draws constitute a size $N$ data set $\rmX = \set{\rvx_{n}}_{n=1}^{N}$.  Our goal then is to fit a flow-based model $\px(\rvx; \vtheta)$ to $\rmX$ such that the model serves as a good approximation for $\px^{*}(\rvx)$.

Often, the data $\set{\rvx_{n}}_{n=1}^{N}$ are discrete; for example, they could be images with pixel values in $\set{0, 1,\ldots, 255}$. Flow-based models are defined over continuous random variables (with the exception of discrete flows, \Cref{sec:discrete_random_variables}), so they are not directly applicable to discrete data. To use flows with discrete data, we often dequantize $\set{\rvx_{n}}_{n=1}^{N}$ by adding continuous noise. The noise distribution can be fixed (e.g.\ uniform in $[0, 1]$ for the image example above), or learned, as for example in \textit{variational dequantization} \citep{ho2019flow++}.

One of the most popular method for fitting $\px(\rvx; \vtheta)$ is \textit{maximum likelihood estimation}, which exploits the forward KL divergence first introduced in \Cref{sec:mle}: \begin{equation}\begin{split}\label{eq:kld}
    \kl{\px^*(\rvx)}{\px(\rvx; \vtheta)} &= -\avg{\log \px(\rvx; \vtheta)}{\px^{*}(\rvx)} + \mathrm{const} \\ &\approx -\frac{1}{N} \sum_{n=1}^{N} \log \px(\rvx_{n}; \vtheta) + \mathrm{const} \\ &= -\frac{1}{N} \sum_{n=1}^{N} \log \pu(T^{-1}(\rvx_{n}; \vphi); \vpsi)  +\log  \left| J_{T^{-1}}(\rvx_{n}; \vphi)  \right| + \mathrm{const}.
\end{split}\end{equation}  However, in principle any valid divergence or integral probability metric can be used as an optimization target, as discussed in \Cref{sec:alternative_divergences}. Typically, there are two downstream uses for the resulting model $\px(\rvx; \vtheta)$: \textit{density estimation} and \textit{generation}.

\subsubsection{Density Estimation}  The first task is primarily quantitative: we can use the model to estimate densities, expectations, marginals, or other quantities of interest on never-before-seen data.  Early work \citep{chen2001gaussianization, tabak2013family} considered only synthetic low-dimensional cases, showing that normalizing flows could indeed represent skewed, multi-modal densities as well as kernel density estimators could.  It was \citet{laparra2011iterative} that first applied Gaussianization to real data, using the density function to perform one-class classification to detect urban areas in satellite images.  \citet{rippel2013high} next showed that their deep flow model's density could detect rotations and corruptions of images.  \citet{papamakarios2017masked} performed a systematic comparison of their masked autoregressive flow on unconditional and conditional density estimation tasks, showing that the composition enabled by their framework allows for better density estimation than other variants (namely MADE and Real NVP)\@.  \citet{grathwohl2018ffjord} performed similar experiments to validate the effectiveness of continuous-time flows.

\subsubsection{Generation}
The second task is generation: sampling from the model novel instances that could have plausibly been sampled from $\px^{*}(\rvx)$.  In this latter case, the availability of exact likelihood values is not the end goal so much as a principled training target that one would expect to result in good generative performance.  Generation has been a popular application of flows in machine learning, and below we summarize their use for various categories of data. 

\paragraph{Images \& video}
Image generation has been given serious effort since the earliest work on flows.  \citet{laparra2011iterative}, in the same work mentioned above, used Gaussianization to generate gray-scale images of faces.  \citet{rippel2013high} also demonstrated early success in generative performance as MNIST samples from their model looked rather compelling.  \citet{dinh2014nice}, through the use of their coupling parameterization, showed further improvements including density estimation competitive with other high-capacity models (such as deep mixtures of factor analysers) and respectable generation of SVHN digits.  In follow-up work, \citet{dinh2016density} increased the capacity of their model by including scale transformations (instead of just translations), being the first to demonstrate that flows could produce sharp, visually compelling full-color images.  Specifically, \citet{dinh2016density} showed compelling samples from models trained on CelebA, ImageNet ($64 \times 64$), CIFAR-10, and LSUN\@.  \citet{kingma2018glow}, using a similar model but with additional convolutional layers, further improved upon \citet{dinh2016density}'s results in density estimation and generation of high-dimensional images.  Continuous \citep{grathwohl2018ffjord} and residual \citep{behrmann2019invertible, chen2019residual} flows have been demonstrated to produce sharp, high-dimensional images as well. Finally, \citet{kumar2019videoflow} propose a normalizing flow for modeling video data, adapting the Glow architecture 
\citep{kingma2018glow} to synthesize raw RGB frames.

\paragraph{Audio} 
The autoregressive model \textit{WaveNet} \citep{oord2016wavenet} demonstrated impressive performance in audio synthesis.  While WaveNet is not a normalizing flow, in follow-up work \citet{oord2017parallel} defined a proper flow for audio synthesis by distilling WaveNet into an inverse autoregressive flow so as to make test-time sampling more efficient.  \citet{prenger2018waveglow} and \citet{kim2018flowavenet} have since  formulated WaveNet variants built from coupling layers to enable fast likelihood and sampling, in turn obviating the need for \citet{oord2017parallel}'s post-training distillation step.  

\paragraph{Text}  
The most direct way to apply normalizing flows to text data is to define a discrete flow over characters or a vocabulary.  \citet{tran2019discrete} take this approach, showing performance in character-level language modeling competitive to RNNs while having superior generation runtime.  An alternative approach that has found wider use is to define a latent variable model with a discrete likelihood but a continuous latent space.  A normalizing flow can then be defined on the latent space as usual. \citet{pmlr-v97-ziegler19a} use such an approach for character-level language modeling.  \citet{zhou2019density}, \citet{he2018unsupervised}, and \citet{jin-etal-2019-unsupervised} define normalizing flows on the continuous space of word embeddings as a subcomponent of models for translation, syntactic structure, and parsing respectively.

\paragraph{Other structured objects}
Extending flows to operate on other structured objects is a burgeoning area of work. So far, flows have been applied to graphs \citep{geng2019continuous}, molecules \citep{madhawa2019graphnvp, honda2019graph}, point clouds \citep{pointCloudFlow}, and part models for motion synthesis \citep{henter2019moglow}.

\subsection{Inference}
In the previous section our focus was on modeling data and recovering its underlying distribution.  We now turn to inference: estimating unknown quantities within a model.  The most common setting is the computation of high-dimensional, analytically intractable integrals of the form:
\begin{equation}
    \integral{\pi(\veta)}{\veta}.    
\end{equation}
Bayesian inference usually runs into such an obstacle when computing the posterior's normalizing constant or when computing expectations under the posterior.  Below we summarize the use of flows for sampling, variational inference, and likelihood-free inference.    

\subsubsection{Importance and Rejection Sampling}
\textit{Importance sampling} (IS) computes intractable integrals by converting them to an expectation under an auxiliary distribution $q(\veta)$: \begin{equation}
    \integral{\pi(\veta)}{\veta} = \integral{q(\veta) \frac{\pi(\veta)}{q(\veta)}}{{\veta}} = \avg{\frac{\pi(\veta)}{q(\veta)} }{q(\veta)} \approx \frac{1}{S} \sum_{s=1}^{S} \frac{\pi(\hat{\veta}_{s})}{q(\hat{\veta}_{s})},
\end{equation} where $q(\veta)$ is a user-specified density function and $\hat{\veta}_{s}$ is a sample from $q(\veta)$. Clearly, IS requires both sampling and density evaluation.  Since both operations are tractable for many flows, they make for an attractive model from which to construct the proposal.

\citet{muller2019neural} do just this: they implement $q(\veta)$ using  normalizing flows.  The practicality of IS crucially depends on the choice of proposal and thus the flow's parameters must be optimized.  \citet{muller2019neural} discuss two strategies for fitting $q(\veta)$.  When $\pi(\veta)$ can be interpreted as an unnormalized density, the first is to minimize the KL divergence between the normalized target and the flow: $\kl{p(\veta)}{q(\veta)}$ where $p(\veta) = \pi(\veta) / Z$ with $Z = \integral{\pi(\veta)}{\veta}$ being an intractable normalizing constant (equal to the very quantity that we wish to compute).  While $\kl{p(\veta)}{q(\veta)}$ cannot be computed, IS can be used to compute the divergence's gradient with respect to the flow's parameters.  The second is to minimize the variance of the IS estimator directly.  When $\pi(\veta)$ is again an unnormalized density, this is equivalent to minimizing a $\chi^{2}$-divergence between the proposal and $p(\veta) = \pi(\veta) / Z$. Flows have also been used for the proposal distribution in similar ways by \citet{noe2019boltzmann} and \citet{wirnsberger2020lbar}. 

The related technique of \textit{rejection sampling} (RS) aims to draw samples from $p(\veta) = \pi(\veta) / Z$, where $\pi(\veta)$ is again an unnormalized density.  Both density evaluation and sampling are required from the proposal in RS, again making normalizing flows well-suited.  \citet{bauer2019resampled} use the Real NVP architecture \citep{dinh2016density} to parameterize a proposal distribution for RS since the coupling layers allow for fast density evaluation and sampling. 

\subsubsection{Markov Chain Monte Carlo}

The application of flows in \textit{Markov chain Monte Carlo} (MCMC)  precedes the appearance of flows in deep learning by at least a few decades. One prominent example is \textit{Hamiltonian Monte Carlo} (HMC), also known as \textit{Hybrid Monte Carlo} \citep{duane1987hmc, neal2010hmc}. HMC operates on the `phase space' $(\veta, \rvv)$, where $\veta$ are the variables of interest and $\rvv$ are additional `momentum' variables of the same dimensionality as $\veta$. HMC generates samples from a joint distribution $p(\veta, \rvv)$ constructed so that its marginal over $\veta$ is the distribution of interest. Central to HMC is the \textit{Hamiltonian} defined by $H(\veta, \rvv)= -\log p(\veta, \rvv)$. Given a state $(\veta, \rvv)$, HMC proposes a new state $(\veta', \rvv') = T(\veta, \rvv)$ deterministically, where $T$ is a \textit{Hamiltonian flow} followed by negation of the momentum variables. The proposed state is then accepted/rejected using the usual Metropolis--Hastings step. The Hamiltonian flow is the continuous-time flow generated by the following ODE:
\begin{equation}
    \frac{d(\veta, \rvv)}{dt} = \br{\frac{\partial H}{\partial\rvv}, -\frac{\partial H}{\partial\veta}}.
\end{equation}
This flow is volume-preserving, meaning that its absolute Jacobian determinant is $1$ everywhere, which, in combination with the negation of the momentum variables, ensures that the proposal is symmetric and thus cancels in the Metropolis--Hastings ratio.

It is also possible to construct MCMC algorithms with flows other than the Hamiltonian flow described above. One example is A-NICE-MC \citep{song2017anicemc}, which is similar to HMC but constructs the proposal using an arbitrary volume-preserving flow $T(\cdot; \vphi)$ parameterized by $\vphi$. \citet{song2017anicemc} use the NICE model of \citet{dinh2014nice}, but their method applies to any volume-preserving flow more generally. Given a state $(\veta, \rvv)$, a new state is proposed which is equal to either $T(\veta, \rvv; \vphi)$ or $T^{-1}(\veta, \rvv; \vphi)$ with equal probability. This proposal is symmetric and so it cancels in the Metripolis--Hastings ratio. The parameters $\vphi$ are tuned to the distribution of interest using adversarial training.

Another way of applying flows to MCMC is to use the flow to reparameterize the target distribution.  It is well understood that the efficiency of MCMC drastically depends on the target distribution being easy to explore.  If the target is highly skewed and/or multi-modal, the performance of MCMC suffers, resulting in slow mixing and convergence.  Normalizing flows can effectively `smooth away' these pathologies in the target's geometry by allowing MCMC to be run on the simpler and better-behaved base density.  Given the unnormalized target $\pi(\veta)$, we can reparameterize the model in terms of a base density $\pu(\rvu)$ such that $\veta = T(\rvu; \vphi)$. Assuming a symmetric proposal distribution for simplicity, applying the Metropolis--Hastings ratio to the reparameterized model yields: 
\begin{equation}
    r(\hat{\rvu}_{*} ; \hat{\rvu}_{t}) = \frac{\pu(\hat{\rvu}_{*})}{\pu(\hat{\rvu}_{t})} = \frac{\pi(T(\hat{\rvu}_{*}; \vphi)) \,  \abs{\det J_{T}\br{\hat{\rvu}_{*}; \vphi}}}{\pi(T(\hat{\rvu}_{t}; \vphi)) \, \abs{\det J_{T}\br{\hat{\rvu}_{t};  \vphi}}},
\end{equation} where $\hat{\rvu}_{*}$ denotes the proposed value and $\hat{\rvu}_{t}$ the current value.  Assuming $T$ is sufficiently powerful such that $T^{-1}(\veta; \vphi)$ is truly distributed according to the simpler base distribution, exploring the target should become considerably easier.  In practice, it is still useful to generate proposals via Hamiltonian dynamics rather than from a simple isotropic proposal, even if $\pu(\rvu)$ is isotropic \citep{hoffman2019neutra}.

While the reparameterization above is relatively straightforward, there is still the crucial issue of how to set or optimize the parameters of $T$. \citet{titsias2017learning} interleaves runs of the chain with updates to the flow's parameters, performing optimization by maximizing the unnormalized reparameterized target under the last sample from a given run: 
\begin{equation}
    \argmax_{\vphi} \log \pi(T(\hat{\rvu}_{t_{\text{final}}} ; \vphi)) + \log\abs{ \det J_{T}\br{\hat{\rvu}_{t_{\text{final}}};  \vphi}}.
\end{equation}
However, this choice is a heuristic, and is not guaranteed to encourage the chain's mixing.  As \citet{hoffman2019neutra} point out, such a choice may emphasize mode finding.  As an alternative, \citet{hoffman2019neutra} propose fitting the flow model to $p(\veta) = \pi(\veta)/Z$ first via variational inference and then running Hamiltonian Monte Carlo on the reparameterized model, using a sample from the flow to initialize the chain.

\subsubsection{Variational Inference}\label{sec:applications:variational_inference}

We can also use normalizing flows to fit distributions over latent variables or model parameters.  Specifically, flows can usefully serve as posterior approximations for local \citep{rezende2015variational,van2018sylvester,iaf,tomczak2016householder} and global \citep{louizos2017multiplicative} variables. 

For example, suppose we wish to infer variables $\veta$ given some observation $\rvx$. In variational inference with normalizing flows, we use a (trained) flow-based model $q(\veta; \vphi)$ to approximate the posterior as follows:
\begin{equation}
p(\veta\g\rvx) \approx q(\veta; \vphi) = \qu(\rvu) \,\left| \det J_{T}(\rvu; \vphi)  \right|^{-1},
\end{equation}
where $\qu(\rvu)$ is the base distribution (which here is typically fixed) and $ T(\cdot; \vphi)$ is the transformation (parameterized by $\vphi$). If we want to approximate the posterior for multiple values of $\rvx$, we can make the flow model conditional on $\rvx$ and amortize the cost of inference across values of $\rvx$.
The flow is trained by maximizing the \textit{evidence lower bound} (ELBO), which can be written as:  \begin{equation}\begin{split}\label{eq:elbo}
\log p(\rvx) &\ge \avg{\log p\br{\rvx, \veta}}{q(\veta; \vphi)} - \avg{ \log q( \veta; \vphi) }{q(\veta; \vphi)}    \\
&= \avg{\log p\br{\rvx, T(\rvu; \vphi)}}{\qu(\rvu)} - \avg{\log \qu( \rvu)}{\qu(\rvu)} + \avg{\log \left| \det J_{T}(\rvu; \vphi)  \right|}{\qu(\rvu)} \\ &= \avg{\log p\br{\rvx, T(\rvu; \vphi)}}{\qu(\rvu)} + \mathbb{H}\left[\,\qu(\rvu)\,\right] + \avg{\log \left| \det J_{T}(\rvu; \vphi)  \right|}{\qu(\rvu)},
\end{split}\end{equation} where $\mathbb{H}\left[\,\qu(\rvu)\,\right]$ is the differential entropy of the base distribution, which is a constant with respect to $\vphi$.
The expectation terms can be estimated by Monte Carlo, using samples from the base distribution as follows: \begin{equation}\begin{split}
    \avg{\log p(\rvx, T(\rvu; \vphi))}{\qu(\rvu)} &\approx \frac{1}{S} \sum_{s=1}^{S} \log p(\rvx, T(\hat{\vu}_{s}; \vphi)), \\ \avg{\log \left| \det J_{T}(\rvu; \vphi)  \right|}{\qu(\rvu)} &\approx \frac{1}{S} \sum_{s=1}^{S} \log \left| \det J_{T}(\hat{\vu}_{s}; \vphi)  \right|.
\end{split}
\end{equation}
Normalizing flows can be thought of as implementing a `generalized reparameterization trick' \citep{kingma2013auto, rezende2014stochastic, kingma2014efficient}, as they leverage a transformation of a fixed distribution to draw samples from a distribution of interest. Flows therefore define flexible approximate posteriors that are readily reparameterizable by design.

\subsubsection{Likelihood-Free Inference}

Models are often implicit, meaning they are not defined in terms of a likelihood function $p(\rvx\g\veta)$ that describes how observable variables $\rvx$ depend on model parameters $\veta$. Rather, they come in the form of a simulator that takes in parameters $\veta$ and simulates variables $\rvx$ \citep{diggle1984monte}. Such simulator-based models are common in scientific fields such as cosmology \citep{Alsing:2018:lfi_cosmology}, high-energy physics \citep{Brehmer:2018:eft}, and computational neuroscience \citep{goncalves2020training}. Inferring the parameters $\veta$ of a simulator-based model given observed data $\rvx$ is often referred to as \textit{likelihood-free inference} \citep{papamakarios2019phd_thesis}, \textit{simulation-based inference} \citep{cranmer2020frontier}, or \textit{approximate Bayesian computation} \citep{beaumont2002abc, beaumont2010abc}. The typical assumption in likelihood-free inference is that it is easy to simulate variables $\rvx$ from the model given  $\veta$, but it is intractable to evaluate the likelihood $p(\rvx\g\veta)$. 

Normalizing flows are a natural fit for likelihood-free inference, especially flows conditioned on side information \citep[e.g.][]{winkler2019conditional, ardizzone2019guided}. Assuming a tractable prior distribution $p(\veta)$ over the parameters of interest, we can generate a data set $\set{(\veta_n, \rvx_n)}_{n=1}^N$ where $\veta_n\sim p(\veta)$ and $\rvx_n$ is simulated from the model with parameters $\veta_n$. In other words, $(\veta_n, \rvx_n)$ is a joint sample from $p(\veta, \rvx)=p(\veta)\,p(\rvx\g\veta)$. Then, using the techniques described in \Cref{sec:applications:probabilistic_modeling}, we can fit a flow-based model $q(\veta \g \rvx)$ conditioned on $\rvx$ to the generated data set $\set{(\veta_n, \rvx_n)}_{n=1}^N$ in order to approximate the posterior $p(\veta\g\rvx)$ \citep{greenberg2019automatic, goncalves2020training}. Alternatively, we can fit a flow-based model $q(\rvx \g \veta)$ conditioned on $\veta$ in order to approximate the intractable likelihood $p(\rvx\g\veta)$ \citep{papamakarios2019sequential}. In either case, the trained flow-based model is useful for various downstream tasks that require density evaluation or sampling.

\subsection{Using Flows for Representation Learning}

Flows also have applications as building blocks for downstream tasks. We discuss two such cases, namely supervised learning and reinforcement learning.
\subsubsection{Classification and Hybrid Modeling}  
  Invertible ResNets---in addition to implementing residual flows, as discussed in \Cref{sec:residual_flows}---have been explored for classification \citep{jacobsen2018irevnet, behrmann2019invertible}.  This line of work exploits invertibility for other end-purposes than density computation.  The first is for engineering improvements: to reduce the model's memory footprint by obviating the need to store activations for backpropagation \citep{gomez2017reversible}.  The second is for improved model interpretability and understanding of the mechanics of deep learning.  \citet{jacobsen2018irevnet} showed that an invertible ResNet could be trained to nearly the same ImageNet accuracy as a non-invertible ResNet.  This achievement may help us understand to what degree discarding information is crucial to deep learning's success \citep{tishby2015deep}.  \citet{jacobsen2018excessive} used invertible architectures to study the relationship between invariance and vulnerability to adversarial attacks.

Flows can be used for joint generative and predictive modeling by using them as the core component of a \textit{hybrid model} \citep{nalisnick2019hybrid}.  Like in invertible ResNets, the flow is used as a deep, neural feature extractor, but unlike in ResNets, the architecture is chosen such that the Jacobian determinant is tractable.  Leveraging the generative model $\rvx = T(\rvu)$ to reparameterize the joint density, we can write: \begin{equation}\label{eq:hybrid}\begin{split}
    \log p(\rvy, \rvx) &= \log p(\rvy \g \rvx) + \log p(\rvx) \\ &= \log p\br{\rvy \g T^{-1}(\rvx)} + \log \pu\br{T^{-1}(\rvx)} + \log \abs{\det J_{T^{-1}}(\rvx)}. 
\end{split}
\end{equation}  
We can think of the flow as defining the first $L-1$ layers of the architecture, and of the last layer as a (generalized) linear model operating on the features $\rvu = T^{-1}(\rvx)$.  The second term then makes these features distribute according to $\pu(\rvu)$, which could be viewed as a regularizer on the feature space.  For instance, if $\pu(\rvu)$ is standard Normal, then $\log \pu\br{T^{-1}(\rvx)}$ effectively acts as an $\ell_2$ penalty.  The Jacobian determinant serves its usual role in ensuring the density is properly normalized.  Hence, a hybrid model in the form of \Cref{eq:hybrid} can compute the joint density of labels and features at little additional cost to standard forward propagation.  The extra computation is introduced by the right-most two terms and depends on their particular forms.  If $\pu(\rvu)$ is standard Normal and $T$ defined via coupling layers, then the additional computation is $\mathcal{O}(DL)$ with $D$ being the number of input dimensions and $L$ the number of layers.

\subsubsection{Reinforcement Learning}  
Finally, in this section we give two examples of how normalizing flows have been used thus far for reinforcement learning (RL)\@.  

\paragraph{Reparameterized policies} The most popular use for flows in RL has been to model (continuous) policies. An action $\rva_{t} \in \mathbb{R}^{D}$ taken in state $\rvs_{t}$ at time $t$ is sampled according to $\hat{\rva}_{t} = T(\hat{\rvu}_{t}; \rvs_{t}, \vphi)$, $\hat{\rvu}_{t} \sim \pu(\rvu_{t})$ where $\pu$ denotes the base density.  The corresponding conditional density is written as: 
\begin{equation}
pi(\rva_{t} \g \rvs_{t}) = \pu(\rvu_{t}) \abs{\det J_{T}(\rvu_{t}; \rvs_{t}, \vphi)}^{-1}.
\end{equation}
\citet{haarnoja2018latent} and \citet{ward2019improving} use such a policy within the maximum entropy and soft actor-critic frameworks respectively. 

\paragraph{Imitation learning}  \citet{schroecker2019imitation} use normalizing flows within the imitation-learning paradigm for control problems.  Given the observed expert (continuous) state-action pairs $\br{\bar{\rvs}, \bar{\rva}}$, a core challenge to imitation learning is accounting for unobserved intermediate states. \citet{schroecker2019imitation} use conditional flows to simulate these intermediate states and their corresponding actions.  Specifically, for a state $\bar{\rvs}_{t+j}$ ($j \ge 1$), predecessor state-action pairs are sampled from the model:
\begin{equation}
p(\rvs_{t}, \rva_{t} \g \bar{\rvs}_{t+j}) = p(\rva_{t} \g \rvs_{t}, \bar{\rvs}_{t+j})\, p(\rvs_{t} \g \bar{\rvs}_{t+j}).
\end{equation}
Both terms on the right-hand side are defined by MAFs \citep{papamakarios2017masked}.

\section{Conclusions}
\label{sec:conclusions}

We have described normalizing flows and their use for probabilistic modeling and inference.  We addressed key issues such as their expressive power (\Cref{sec:flows:expressivity}) and the fundamentals underlying their construction (both in discrete and continuous time).  We also described the general principle of probability transformations (\Cref{sec:generalizations:formula}) and its implications for defining flows beyond Euclidean space. In particular, we showed that discrete domains, mixtures of flows, and extensions to Riemannian manifolds all follow from this generalized perspective.  Lastly, we summarized the primary applications of flows (\Cref{sec:applications}): tasks ranging from density estimation to likelihood-free inference to classification.

While many flow designs and specific implementations will inevitably become out-of-date as work on normalizing flows continues, we have attempted to isolate foundational ideas that will continue to guide the field well into the future.  One of these keystone principles is the chain rule of probability and its relationship to transformations with a triangular Jacobian. Autoregressive flows stand on these two pillars, with the former underlying their expressive power and the latter providing their efficient implementation.  Similarly, the Banach fixed-point theorem provides the mathematical foundation for contractive residual flows.  While alternative parameterizations of the translation function $g_\vphi$ or normalization strategies may be developed, the underlying Lipschitz constraints cannot be deserted without violating the fixed-point theorem.

Throughout the text we emphasized crucial implementation notes that guide a successful application of flows.  Perhaps of foremost importance is determining the computational constraints on evaluating the forward and inverse transformations.  As we showed in \Cref{sec:flows}, sampling and density evaluation place distinct demands on the transformation. Since flows have no inherent requirement as to whether $T$ should implement $\mathcal{U} \rightarrow \mathcal{X}$ or vice versa, we are free to choose which direction better suits our application.  If either sampling or density estimation is the primary objective---but not both---then autoregressive flows present an attractive, flexible class of model.  Yet if both sampling and density evaluation must be done often or quickly, then implementing the autoregressive flow with a coupling-based conditioner will make both operations efficient at the cost of expressive power.  On the other hand, non-autoregressive flows such as linear and residual flows allow for interaction between all dimensions at each step in the flow.  While these interactions can be useful at times---making linear flows good for permuting variables between successive autoregressive flows and residual flows highly expressive---other limitations arise.  For instance, contractive residual flows typically require iterative algorithms for sampling \emph{and} density evaluation.  As all flow constructions present trade-offs of some form, we hope this article provides a coherent and accessible summary to guide practitioners through these choice points.   

Looking forward, the obstacles preventing wider application of normalizing flows are similar in spirit to those faced by any probabilistic model.  However, unlike other probabilistic models that require approximate inference as they scale, flows usually admit analytical calculations and exact sampling even in high dimensions.  Rather, the difficulty is transferred to the construction of the flow's transformation: how can we define ever more flexible transformations while keeping exact density evaluation and sampling computationally tractable?  This is currently the focus of much work and will likely remain a core issue for some time.  More study of the theoretical properties of flows is also needed.  Understanding their approximation capabilities for finite sample and finite depth settings would help practitioners select which flow classes are best for a given application.  Our discussion of generalizations in \Cref{sec:generalizations} will hopefully provide grounding as well as inspiration for this next wave of developments in the theory and application of normalizing flows.

\acks{We would like to thank Ivo Danihelka for his invaluable feedback on the manuscript, and the anonymous reviewers for their many improvement suggestions. We also thank Hyunjik Kim and S\'{e}bastien Racani\`{e}re for useful discussions on a wide variety of flow-related topics.}

\appendix

\section{Proof of KL Dualities}\label{app:kl_duality}

Let $\px(\rvx)$ be the distribution induced by a flow with transformation $T$ and base distribution $\pu(\rvu)$. Also, let $\pu^*(\rvu)$ be the distribution induced by the inverse flow with transformation $T^{-1}$ and base distribution $\px^*(\rvx)$. Using the formula for the density of a flow-based model and a change of variables, we have the following:
\begin{equation}\begin{split}
    \kl{\px^*(\rvx)}{\px(\rvx)} &= \avg{\log\px^*(\rvx) - \log\px(\rvx)}{\px^*(\rvx)}\\
    &= \avg{\log\px^*(\rvx) - \log\abs{\det J_{T^{-1}}(\rvx)} - \log\pu\br{T^{-1}(\rvx)}}{\px^*(\rvx)}\\
    &= \avg{\log\px^*(T(\rvu)) + \log\abs{\det J_{T}(\rvu)} - \log\pu\br{\rvu}}{\pu^*(\rvu)}\\
    &= \avg{\log\pu^*(\rvu) - \log\pu\br{\rvu}}{\pu^*(\rvu)}\\
    &= \kl{\pu^*(\rvu)}{\pu(\rvu)}.
\end{split}\end{equation}
Similarly, we have the following:
\begin{equation}\begin{split}
    \kl{\px(\rvx)}{\px^*(\rvx)} &= \avg{\log\px(\rvx) - \log\px^*(\rvx)}{\px(\rvx)}\\
    &= \avg{\log\pu\br{T^{-1}(\rvx)} + \log\abs{\det J_{T^{-1}}(\rvx)} - \log\px^*\br{\rvx}}{\px(\rvx)}\\
    &= \avg{\log\pu(\rvu) - \log\abs{\det J_{T}(\rvu)} - \log\px^*\br{T(\rvu)}}{\pu(\rvu)}\\
    &= \avg{\log\pu(\rvu) - \log\pu^*\br{\rvu}}{\pu(\rvu)}\\
    &= \kl{\pu(\rvu)}{\pu^*(\rvu)}.
\end{split}\end{equation}

\section{Constructing Linear Flows}\label{app:linFlow}
A common approach to efficiently parameterizing invertible matrices is via \textit{matrix decompositions}. The idea is to decompose the matrix $\rmW$ into a product of structured matrices, each of which can be easily constrained to be invertible and has a $\bigo{D^2}$ inverse and $\bigo{D}$ determinant. Below we discuss a few such parameterizations.

\paragraph{PLU flows} Every $D\times D$ matrix $\rmW$ can be written as follows:
\begin{equation}
    \rmW = \rmP\rmL\rmU,
\end{equation}
where $\rmP$ is a permutation matrix, $\rmL$ is lower triangular, $\rmU$ is upper triangular, and all three are of size $D\times D$. We can easily constrain $\rmW$ to be invertible by restricting $\rmL$ and $\rmU$ to have positive diagonal entries. In that case, the absolute determinant of $\rmW$ can be computed in $\bigo{D}$ time by:
\begin{equation}
    \abs{\det\rmW} = \prod_{i=1}^D L_{ii} U_{ii}.
\end{equation}
Moreover, the linear system $\rmP\rmL\rmU \rvz = \rvz'$ can be solved by (a) undoing the permutation, (b) solving a lower-triangular system, and (c) solving an upper-triangular system. Solving triangular systems can be done in time $\bigo{D^2}$ by forward/backward substitution, hence inverting the PLU flow can also be done in time $\bigo{D^2}$. In practice, $\rmP$ is typically fixed to a chosen or random permutation, and $\rmL$ and $\rmU$ are learned.

Multiplication with a triangular matrix can be viewed as a special case of an autoregressive flow with affine transformer and linear conditioner. Hence, the PLU flow is simply a composition of two such autoregressive flows with opposite order, followed by a permutation. Linear flows based on the PLU decomposition have been proposed by e.g.~\citet{oliva2018tan, kingma2018glow}. A similar decomposition in the case where $\rmW$ is a convolution operator was proposed by \citet{hoogeboom2019emerging} and was termed \textit{emerging convolution}.

\paragraph{QR flows} Another way to decompose a $D\times D$ matrix is using the QR decomposition as follows:
\begin{equation}
    \rmW = \rmQ\rmR,
\end{equation}
where $\rmW$ is an orthogonal matrix and $\rmR$ is upper triangular, both of size $D \times D$. In fact, we can parameterize any invertible matrix using the QR decomposition, even if we restrict the diagonal entries of $\rmR$ to be positive. In that case, the absolute determinant of  $\rmW$ is given in $\bigo{D}$ time by:
\begin{equation}
    \abs{\det\rmW} = \prod_{i=1}^D R_{ii}.
\end{equation}
Inverting the linear system $\rmQ\rmR \rvz = \rvz'$ can be done in $\bigo{D^2}$ time by first multiplying by $\rmQ^\top$ and then solving a triangular system by forward/backward substitution. We can also think of the QR flow as a linear autoregressive flow followed by an orthogonal flow; in the next paragraph we will discuss in more detail how to parameterize an orthogonal flow. The QR flow was proposed by \citet{hoogeboom2019emerging}.

\paragraph{Orthogonal flows} This is a special case of a linear flow where $\rmW = \rmQ$ is an orthogonal matrix, i.e.~a matrix whose columns form an orthonormal basis in $\R^D$. Orthogonal flows are volume preserving since $\abs{\det\rmQ} = 1$ and trivially invertible since $\rmQ^{-1} = \rmQ^\top$. They generalize permutations (every permutation matrix is orthogonal), and can be used on their own or in composition with other structured matrices as in the QR flow above.

Nonetheless, an effective parameterization of orthogonal matrices can be challenging \citep[see e.g.][for reviews on the subject]{shepard2015orthogonal, lezcanocasado2019orthogonal}. As we discussed earlier, either the orthogonal matrices with determinant $1$ or those with determinant $-1$ can be continuously parameterized, but not both. In the normalizing-flows literature, the following parameterizations have been explored, each with their own strengths an weaknesses.
\begin{itemize}[leftmargin=*]
    \item \textit{The exponential map} \citep{golinsky2019orthogonal}. Given a skew-symmetric matrix $\rmA$, i.e.\ a \mbox{$D\times D$} matrix such that $\rmA^\top = -\rmA$, the matrix exponential $\rmQ = \exp\rmA$ is always an orthogonal matrix with determinant $1$. Moreover, any orthogonal matrix with determinant~$1$ can be written this way. However, computing the matrix exponential takes in general $\bigo{D^3}$ time, so this parameterization is only suitable for small-dimensional data.
    \item \textit{The Cayley map} \citep{golinsky2019orthogonal}. Again given a skew-symmetric matrix $\rmA$ as above, the matrix $\rmQ = (\rmI + \rmA)(\rmI - \rmA)^{-1}$ is always orthogonal with determinant $1$. However, this parameterization also takes $\bigo{D^3}$ time to compute, and can only parameterize those orthogonal matrices that don't have $-1$ as an eigenvalue.
    \item \textit{Householder transformations} \citep{tomczak2016householder}. Any orthogonal transformation in $\R^D$ can be expressed as a composition of at most $D$ reflections. A single reflection about a hyperplane perpendicular to a non-zero vector $\rvv_k$, known as a \textit{Householder transformation}, is given by the following matrix:
    \begin{equation}
        \rmH_k = \rmI - 2\frac{\rvv_k\rvv_k^\top}{\norm{\rvv_k}^2}.
    \end{equation}
    Hence, an orthogonal matrix can be parameterized by a product of $K$ such matrices, i.e.~$\rmQ = \prod_{k=1}^K \rmH_k$, where $K$ doesn't need to be greater than $D$. Each Householder matrix has determinant $-1$, so the determinant of $\rmQ$ is $(-1)^K$. Each Householder transformation can be computed in time $\bigo{D}$, so an orthogonal transformation parameterized this way can be computed in time $\bigo{KD}$. The Householder parameterization is not unique and contains saddle points (for example, any permutation of the vectors $\set{\rvv_1, \ldots, \rvv_K}$ gives the same orthogonal matrix) which can make optimization harder.
\end{itemize}

\bibliography{references}

\end{document}